\documentclass{article} 
\usepackage[eandd, final]{neurips_2026}
\setcitestyle{numbers,square,sort&compress}

\usepackage{amsmath,amsfonts,bm}

\def\eqref#1{equation~\ref{#1}}

\def\1{\bm{1}}

\DeclareMathAlphabet{\mathsfit}{\encodingdefault}{\sfdefault}{m}{sl}
\SetMathAlphabet{\mathsfit}{bold}{\encodingdefault}{\sfdefault}{bx}{n}

\usepackage{hyperref}
\usepackage{multirow}
\usepackage{subcaption}
\usepackage{placeins}
\usepackage{adjustbox}
\usepackage{tcolorbox}
\usepackage{algorithm}      
\usepackage{algpseudocode}  
\tcbuselibrary{skins,breakable}
\usepackage{enumitem}

\usepackage[table]{xcolor}
\definecolor{ProcBlue}{RGB}{0, 100, 200}
\definecolor{ProcOrange}{RGB}{240, 130, 40}
\definecolor{ProcPurple}{RGB}{120, 100, 175}
\newcommand{\best}[1]{\cellcolor[HTML]{D9D9D9}\textbf{#1}}
\usepackage{url}

\makeatletter
\newenvironment{breakablealgorithm}
  {%
   \begin{center}
     \refstepcounter{algorithm}
     \hrule height.8pt depth0pt \kern2pt
     \renewcommand{\caption}[2][\relax]{%
       {\raggedright\textbf{\ALG@name~\thealgorithm} ##2\par}%
       \ifx\relax##1\relax
         \addcontentsline{loa}{algorithm}{\protect\numberline{\thealgorithm}##2}%
       \else
         \addcontentsline{loa}{algorithm}{\protect\numberline{\thealgorithm}##1}%
       \fi
       \kern2pt\hrule\kern2pt
     }
  }
  {%
     \kern2pt\hrule\relax
   \end{center}
  }
\makeatother

\title{DualWorldBench: Can Agents Plan Deliveries across Symbolic and Grounded Worlds?}

\author{%
\begin{tabular}{c}
Yang Chen\textsuperscript{1,2}\thanks{Equal contribution.}\hspace{1.0em}
Bin Wen\textsuperscript{1,2,4}\footnotemark[1]\hspace{1.0em}
Danyang Peng\textsuperscript{1,2}\\[3pt]
Hong-Jie You\textsuperscript{1,3}\hspace{1.0em}
Zi-Qiao Shang\textsuperscript{1,2}\hspace{1.0em}
Lan-Zhe Guo\textsuperscript{1,2}\thanks{Corresponding author.}\\[5pt]
\normalfont\textsuperscript{1}\,National Key Laboratory for Novel Software Technology, Nanjing University\\[2pt]
\normalfont\textsuperscript{2}\,School of Intelligence Science and Technology, Nanjing University\\[2pt]
\normalfont\textsuperscript{3}\,School of Artificial Intelligence, Nanjing University\\[2pt]
\normalfont\textsuperscript{4}\,Shanghai Innovation Institute
\end{tabular}
}

\usepackage{booktabs}
\usepackage{pifont}
\usepackage{graphicx}
\usepackage{wrapfig}

\begin{document}

\maketitle

\begin{abstract}
Recent multimodal agents have shown promising progress in embodied planning and web interaction. However, existing evaluations often rely on simplified tasks with limited combinatorial structure, leaving it unclear whether current agents can robustly integrate perception and reasoning for multi-goal planning under temporal, physical, and multimodal constraints. 
We introduce \textbf{DualWorldBench}, a unified benchmark for evaluating agent planning under controlled \emph{symbolic--temporal} and \emph{grounded--topological} complexity. Built around a shared pickup--delivery interface, DualWorldBench consists of three complementary components: \textbf{TextWorld}, which isolates symbolic planning under varying task compositions and release dynamics; \textbf{SimWorld}, which evaluates grounded path planning under controlled topological variation with text, image, or multimodal inputs; and \textbf{DualWorld-VQA}, a diagnostic suite spanning perception, scene understanding, and planning. 
Experiments reveal a common gap between local understanding and globally consistent planning. In TextWorld, agents struggle to derive optimal task sequences under interacting temporal constraints; in SimWorld, grounded observations are not reliably converted into feasible action sequences. DualWorld-VQA further confirms that perception and topology understanding do not necessarily translate into constrained planning. Together, these results identify the central challenge as composing symbolic or grounded evidence into globally feasible and efficient multi-goal plans.
These findings establish DualWorldBench as a challenging testbed for agents with tighter perception--reasoning--planning integration. All datasets and code are publicly available at our project website \url{https://dualworld.netlify.app/}.

\end{abstract}

\section{Introduction}
\label{sec:intro}

Recent years have witnessed rapid progress in multimodal agents for embodied planning~\citep{eai,embodiedbench,eb1,deliverybench}, web interaction~\citep{webembodied,webshop,web1}, and visual reasoning~\citep{cvbench,embspatial,spatial1,spatial2}. Yet, despite their strong performance on benchmark tasks, current agent systems are largely evaluated in simplified settings: they reason over text and images under limited constraints, with little pressure to jointly handle temporal dependencies, physical feasibility, and multi-source evidence. In realistic deployments, however, agents must make decisions under coupled symbolic and grounded constraints. A web-style workflow may involve temporally ordered actions and delayed task release; an embodied agent may need to infer physically feasible routes from visual observations rather than rely on coordinate proximity alone. The central question is not only whether agents can \emph{perceive} or \emph{reason}, but whether they can integrate both capabilities into effective planning under complex constraints.

\begin{figure*}[t]
  \centering
  \includegraphics[width=\textwidth]{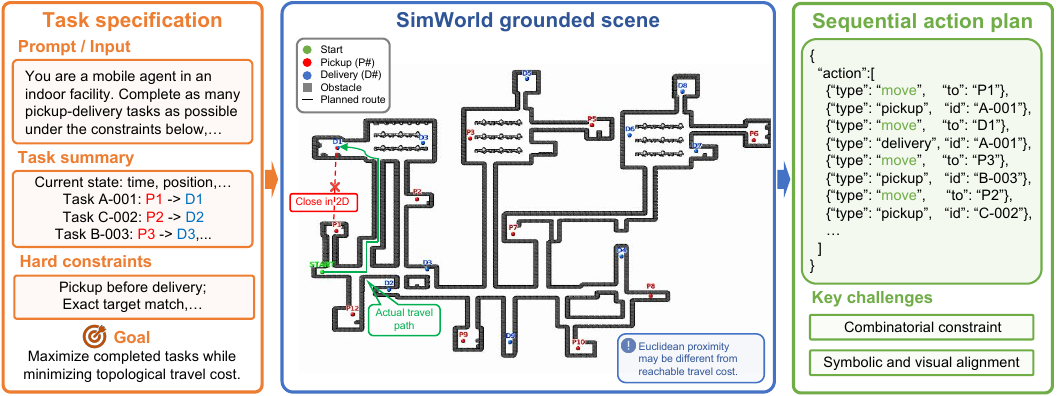}
  \caption{\textbf{DualWorldBench task overview.} The agent maps task specifications and a grounded scene into a topology-aware pickup--delivery action plan under combinatorial constraints.}
  \label{fig:intro}
\end{figure*}

We study this problem through a unified pickup--delivery setting, shown in Fig.~\ref{fig:intro}. An agent is given user demands together with textual and/or visual information about task locations and environment structure, and must execute a sequence of pickup and delivery actions. Success requires jointly reasoning over order timing, task reward, pickup--delivery precedence, capacity limits, and travel cost. This setting exposes two core difficulties. First, planning difficulty grows combinatorially with the number of goals and constraints: the agent must balance reward, deadline, and execution cost at the sequence level rather than solve isolated subtasks. Second, symbolic descriptions and visual environments are not trivially aligned: locations that are close in 2D coordinates may be far apart in actual navigable distance due to scene topology. Effective planning therefore requires both grounded scene understanding and structured constraint reasoning.

Delivery-style tasks have recently emerged as a practical testbed for embodied decision making. Existing delivery-oriented benchmarks study city-scale courier scenarios or human-centered indoor service delivery, showing that delivery naturally involves task ordering, temporal constraints, physical feasibility, and cost--reward trade-offs~\citep{deliverybench,delivery2,delivery3}. These efforts highlight the relevance of delivery as a realistic long-horizon planning problem beyond isolated navigation or perception.

However, despite the importance of this setting, existing benchmarks do not cleanly characterize the joint challenge we aim to study. Some emphasize realistic simulation or end-to-end embodied interaction, where perception, control, and planning are tightly entangled~\citep{behavior}; others stress symbolic routing or scheduling while abstracting away visual grounding and topology-aware feasibility~\citep{planbench}. As a result, it remains difficult to diagnose whether failures arise from weak perception, incomplete spatial grounding, or flawed combinatorial planning. We therefore use pickup--delivery not merely as an application domain, but as a unified task interface for evaluating agent planning that is both \emph{integrated}, in the sense of assessing perception and planning together, and \emph{decomposable}, in the sense of isolating symbolic--temporal difficulty from grounded--topological difficulty.

To address this gap, we introduce \textbf{DualWorldBench}, a benchmark for studying agent planning under controlled combinatorial constraints and modality/topology variation. We cast tasks under a shared pickup--delivery interface and instantiate three complementary components.
\textbf{(1) TextWorld} isolates symbolic--temporal planning. It procedurally generates delivery tasks with different reward structures and temporal urgency, and evaluates global planning and online replanning under synchronous or time-triggered task release.
\textbf{(2) SimWorld} isolates grounded--topological planning. Built on Gazebo, it contains procedural scenes with controlled topology complexity as well as transfer environments with realistic indoor layouts and large-scale campus traversal, and supports text-only, image-only, and multimodal inputs.
\textbf{(3) DualWorld-VQA} serves as a diagnostic suite spanning perception, understanding, and planning, allowing us to probe where and why models fail.

We evaluate a broad range of state-of-the-art LLM-based agents and identify a consistent gap between local competence and globally consistent planning. In TextWorld, frontier agents approach strong online combinatorial baselines but remain below the theoretical optimum when heterogeneous task values, deadlines, and online arrivals interact, indicating that they still struggle to construct the optimal task sequence under temporal constraints. In SimWorld, weaker models often preserve task coverage yet produce infeasible action sequences, whereas stronger models remain reliable across modalities. DualWorld-VQA further shows that, despite improved perception--planning integration, constrained route optimization remain challenging.
Overall, DualWorldBench reveals that the central unresolved challenge is converting symbolic or grounded evidence into globally feasible, constraint-consistent multi-goal plans as temporal and topological complexity increases.

Our contributions are threefold:
\begin{itemize}
    \item We introduce \textbf{DualWorldBench}, a new benchmark for evaluating agent planning under controlled and decomposable symbolic--temporal constraints and grounded--topological constraints within a unified pickup--delivery interface.
    \item We design a decomposable evaluation framework with \textbf{TextWorld}, \textbf{SimWorld}, and \textbf{DualWorld-VQA}, enabling systematic analysis of scheduling, grounding, topology-aware planning, and their interactions.
    \item We provide extensive empirical evidence that current state-of-the-art LLM agents are still brittle under growing combinatorial constraints and multi-source inputs, establishing DualWorldBench as a challenging testbed for future integrated agents.
\end{itemize}

\section{Related Work}
\label{sec:relatedwork}

\paragraph{Grounded spatial reasoning and embodied agents.}
Web agent benchmarks~\citep{mind2web,webarena,visualwebarena} evaluate instruction following and visual grounding through browser interactions. Spatial reasoning benchmarks probe visual relation understanding, multimodal spatial inference, and spatial belief construction through active exploration~\citep{vsr,Spatialmm,spatial3}. Embodied benchmarks further extend evaluation to long-horizon household interaction, dialogue, manipulation, and vision-driven decision making~\citep{alfred,teach,embodiedarena}. Complementary methodological work explores training-free embodied navigation~\citep{harnessvln,navharness} and neuro-symbolic approaches to spatial reasoning, feasibility tracking, and skill induction~\citep{nesy1,nesy2,nesy3,nesy4}. These efforts advance grounded reasoning and long-horizon execution, but end-to-end task outcomes often couple perception, grounding, control, and planning, making individual sources of failure difficult to diagnose. Local spatial judgments do not necessarily translate into globally feasible multi-goal plans.

\paragraph{Constrained planning and pickup--delivery benchmarks.}
Pickup--delivery and dynamic vehicle routing are established problems in operations research, encompassing routing constraints, transportation requests, and changing operational conditions~\citep{survey,survey2}. Recent benchmarks examine spatial planning, path-plan assessment, and multi-criteria route planning for vision-language and multimodal models~\citep{pathplan1,pathplan2,maptab}. Delivery-oriented benchmarks introduce courier-style or service-oriented tasks with deadlines, travel costs, and resource constraints~\citep{deliverybench,travel1,travel2}. Here, route choices affect deadline feasibility, while task timing changes which routes and action sequences remain viable. Effective planning thus requires coordinating task selection, pickup--delivery precedence, and resource allocation across the full execution sequence. In grounded environments, these decisions further depend on navigable connectivity, which cannot be inferred from coordinate proximity alone. Building on these directions, DualWorldBench focuses on distinguishing symbolic--temporal reasoning from grounded--topological planning under a unified task interface. Its shared pickup--delivery formulation enables controlled variation along both axes, supporting targeted analysis of how temporal constraints and grounded topology affect planning beyond aggregate task completion. This design probes the gap between local task competence and globally constraint-consistent planning.

\section{Problem Formulation}
\label{sec:problem_formulation}

We study a unified pickup--delivery planning problem underlying all components of \textbf{DualWorldBench}. The benchmark varies decision complexity, input modality, and environment topology while keeping the task interface fixed. This design enables direct comparison of symbolic planning, grounded multimodal planning, and cross-scene generalization within a common formal framework.

Both worlds instantiate the same pickup-delivery planning problem. A benchmark instance is
\begin{equation}
    \mathcal{I} = \bigl(s_0, \mathcal{T}, \mathcal{E}\bigr),
\end{equation}
where $s_0 = (\mathbf{x}_0, \mathcal{L}_0, b_0, t_0)$ denotes the initial robot state, with initial location $\mathbf{x}_0$, initial carrying set $\mathcal{L}_0$, remaining payload capacity $b_0$, and initial time $t_0$. $\mathcal{T}=\{\tau_i\}_{i=1}^N$ is the task set, and $\mathcal{E}$ is the environment model used to evaluate motion cost. Each task is
\begin{equation}
    \tau_i = \bigl(c_i, \mathbf{p}_i, \mathbf{d}_i, t_i^{\mathrm{trig}}, t_i^{\mathrm{end}}, r_i \bigr),
\end{equation}
where $c_i \in \{A,B,C\}$ is the task type, $\mathbf{p}_i$ and $\mathbf{d}_i$ are pickup/delivery coordinates, $t_i^{\mathrm{trig}}$ is the release time, $t_i^{\mathrm{end}}$ is the deadline, and $r_i$ is the base reward. The executable plan is represented as an action sequence
\begin{equation}
    \pi = (a_1, a_2, \dots, a_M),
\end{equation}
with actions from $\{\texttt{move}, \texttt{pickup}, \texttt{deliver}\}$. Feasibility requires pickup-before-delivery precedence, capacity limits, and execution at correct coordinates. The realized reward is
\begin{equation}
    R_i = \max\bigl(0, r_i - \lambda(c_i)\max(0, t_i^{\mathrm{arr}} - t_i^{\mathrm{end}})\bigr),
\end{equation}
where $t_i^{\mathrm{arr}}$ is the delivery time and $\lambda(c_i)$ is a class-dependent penalty rate. In our implementation, task types $A$, $B$, and $C$ induce increasingly sharp trade-offs among travel distance, urgency, and value. The benchmark objective is to maximize cumulative revenue
\begin{equation}
    J(\pi) = \sum_i R_i,
\end{equation}
subject to the feasibility constraints above.

This formulation is the backbone of DualWorldBench. It keeps the objective fixed while varying the evaluation setting. TextWorld instantiates the full heterogeneous setting above, whereas the current SimWorld release uses a static, uniform-reward subset to foreground grounded spatial reasoning.

\begin{figure*}[t]
  \centering
  \includegraphics[width=\textwidth]{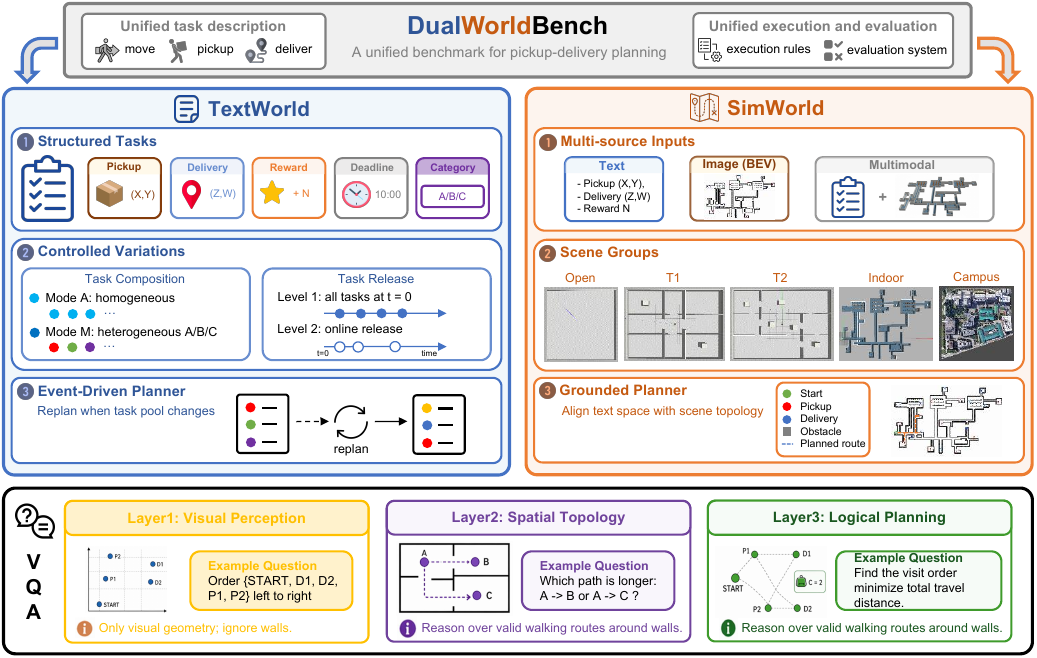}
  \caption{\textbf{DualWorld framework.}
A unified pickup--delivery benchmark with TextWorld for symbolic planning, SimWorld for grounded planning, and DualWorld-VQA for diagnostic evaluation.}
  \label{fig:framework}
\end{figure*}

\section{DualWorldBench}

DualWorldBench is a benchmark for evaluating the planning capabilities of LLMs under controlled variations in task dynamics, input modality, and environment topology. Built upon a shared pickup--delivery interface, the benchmark decomposes evaluation into \textbf{TextWorld}, \textbf{SimWorld}, and \textbf{DualWorld-VQA}. The overall framework is illustrated in Fig.~\ref{fig:framework}.

\subsection{TextWorld Benchmark}
\paragraph{Design Objective.}
TextWorld is designed to isolate the \emph{symbolic--temporal} component of planning. It focuses exclusively on the agent's ability to perform offline and online planning under combinatorial constraints. As such, TextWorld measures multi-task scheduling, temporal priority reasoning, and online replanning in a purely symbolic decision space.

\paragraph{Task Setup.}
We procedurally generate pickup--delivery instances, each specified by pickup/delivery coordinates, task category, reward, and deadline. Valid execution must satisfy both precedence constraints and capacity limits. The agent must reason over these constraints to produce a task sequence that maximizes return while minimizing completion time.

To disentangle different sources of symbolic difficulty, we vary TextWorld along two controlled dimensions. First, we vary the \emph{task composition}. In \textbf{Mode A}, all tasks are of Type-$A$, forming a homogeneous task pool with a simple value structure. In \textbf{Mode M}, tasks are sampled from Type-$A/B/C$, introducing heterogeneous reward scales, deadline urgency, and penalty sensitivity. This A/M split separates pure ordering complexity from value-aware scheduling: Mode A primarily tests combinatorial sequencing, whereas Mode M further requires the model to prioritize across task types with different constraint profiles.
Second, we vary the \emph{task release mechanism}. In \textbf{Level 1}, all tasks are available at $t=0$, yielding a static scheduling problem. In \textbf{Level 2}, tasks are released online according to predefined triggering rules, causing the active task pool to evolve over time and requiring the planner to repeatedly revise its decisions. The resulting factorial design, $\{A, M\}\times\{L1, L2\}$, allows TextWorld to separately evaluate the difficulty of combinatorial growth, heterogeneous value trade-offs, and interruption-driven replanning within a unified symbolic benchmark.

\paragraph{Planning and Evaluation Interface.}
TextWorld is evaluated through a unified event-driven simulator. The simulator maintains the current time, agent state, active task set, carried task set, and completed task set, and invokes the planner whenever the task pool changes or the current plan is exhausted. If a new task arrives while the agent is in motion, execution is interrupted at the interpolated stopping point, and planning restarts from the updated state. Because the input is fully symbolic and the executor is deterministic, this benchmark directly tests whether a planner can convert structured states into executable action sequences and remain consistent across repeated replanning calls.

\subsection{SimWorld Benchmark}

\paragraph{Design Objective.}
SimWorld is designed to isolate the \emph{grounded-topological} component of planning. Complementary to TextWorld, which emphasizes temporally dynamic planning, SimWorld focuses on grounded path planning in structured environments. Its central challenge is that path quality is determined not only by geometric proximity in textual coordinates, but also by topological connectivity in visual scenes. The model must therefore integrate multi-source inputs, align task specifications with scene structure, and generate globally feasible multi-goal plans.

Accordingly, SimWorld primarily evaluates grounded path planning under controlled variations in modality and topology, including spatial information extraction, alignment between symbolic tasks and grounded scene structure, cross-modal robustness, and cross-scene generalization.

\paragraph{Task Setup.}
SimWorld adopts a static task mechanism in which all tasks are available at $t=0$ and share a unified reward. It retains the same pickup--delivery action space and core feasibility rules as TextWorld: each task still consists of a pickup location and a delivery location, and valid execution must still satisfy precedence and capacity constraints.
We design five scene groups with distinct topological characteristics: \textit{open}, \textit{T1}, \textit{T2}, \textit{indoor}, and \textit{campus}. Among them, \textit{open} contains no obstacles and serves as a simulated counterpart to the simplified setting of TextWorld. \textit{T1} and \textit{T2} are deliberately constructed procedural environments for parameterizing topological complexity. \textit{T1} features relatively regular corridor-style layouts, where locally reasonable choices often align with globally short paths. \textit{T2}, by contrast, increases difficulty through denser branching structures and more frequent detours, making Euclidean proximity an unreliable proxy for actual navigation cost. The \textit{indoor} group emphasizes complex enclosed topologies, whereas \textit{campus} emphasizes semi-open environments that require long-range traversal.

Together, these settings define a controlled topology gradient: \textit{T1}/\textit{T2} provide an interpretable and procedural axis of increasing spatial difficulty, while \textit{indoor} and \textit{campus} test transfer beyond synthetic layouts. This design enables the benchmark to distinguish failure modes arising from local grounding, topological reasoning, and cross-scene generalization.

\paragraph{Planning and Evaluation Interface.}
Each SimWorld instance is evaluated under three input modalities. \textbf{Text} provides only symbolic task descriptions and coordinates. \textbf{Image} provides only simplified task descriptions together with a BEV map annotated by visual markers. \textbf{Multimodal} provides both the visual scene and structured task descriptions, enabling explicit alignment between symbolic task specifications and scene content. This design tests not only whether a model can plan from a given modality, but also whether multimodal evidence improves grounded planning.
All modalities are prompted to output the same JSON node sequence, which the evaluator maps to task coordinates and expands into executable \texttt{move}, \texttt{pickup}, and \texttt{deliver} actions. Each mapped coordinate is matched to the nearest valid navigation anchor, and path cost is computed from a preconstructed navigation graph~\citep{localplanner,farplanner} together with local access offsets.

\subsection{DualWorld-VQA}

Beyond end-to-end planning metrics, DualWorldBench includes a \emph{hierarchical diagnostic VQA suite} for analyzing failure modes at intermediate stages of the planning pipeline. Its role is diagnostic: it complements the primary planning experiments with controlled probes that help explain whether errors arise from deficient visual grounding, incomplete spatial understanding, or flawed decision making. Consistent with the benchmark design, VQA items are procedurally generated from simulator state, scene annotations, and task metadata, then verified through a lightweight human validation pass before release. Standard answers are expected to be derived directly from simulator ground truth or scene annotations, and scoring can be instantiated with task-appropriate criteria such as exact match, option accuracy, or constrained judgment correctness.

The suite is organized into three layers. \textbf{Perception-level VQA} targets basic visual geometric perception from the map image, including absolute axis ordering, Euclidean distance comparison to an anchor node, and pairwise Euclidean distance extremes. These questions deliberately ignore walls, connectivity, route cost, and label numbering patterns, so as to isolate whether models can reliably perceive spatial relations in the rendered image. \textbf{Topology-level VQA} introduces wall constraints and evaluates whether models can reason over valid walking routes rather than relying on misleading straight-line proximity. \textbf{Planning-level VQA} further requires models to solve constrained pickup-and-delivery decisions under capacity and pickup-before-delivery constraints, where success depends on jointly using grounded spatial evidence and symbolic task constraints. Together, these probes provide a compact diagnostic decomposition of agent failures into visual perception, spatial topology understanding, and grounded logical planning.




\section{Experiments}
\label{sec:ex}

We evaluate MLLMs on DualWorldBench: \textbf{TextWorld} targets symbolic-temporal planning from structured text, \textbf{SimWorld} targets grounded-topological planning under multimodal observations, and \textbf{DualWorld-VQA} decomposes visual perception, topology reasoning, and constrained planning.

\subsection{Experimental Setup}

We benchmark recent LLMs from both open-weight and proprietary families, including Qwen, Llama, DeepSeek, Claude, Gemini, and GPT variants. TextWorld contains $14$ LLMs and $3$ solver baselines; SimWorld and DualWorld-VQA use the same $6$ VLMs to support paired comparisons.

\paragraph{TextWorld.}
TextWorld measures symbolic-temporal planning from structured task descriptions alone. We report \textbf{Cumulative Reward ($R$)} and \textbf{Completion Time ($T$)}, and compare against three combinatorial baselines: DFS, Online DP, and Offline DP (Appendix~\ref{app:solvers}). We evaluate across task count and the four benchmark regimes induced by the factorial design $\{A,M\}\,\times\,\{L1,L2\}$. Task count ranges from $5$ to $15$. A run that delivers an A/B task while carrying any Type-C task violates the priority constraint and is scored as $R{=}0$, $T{=}9999$.

\paragraph{SimWorld.}
We evaluate three modalities: \textbf{Text}, \textbf{Image}, and \textbf{Multimodal}. All modalities emit the same node-sequence format.
We report three metrics: \textbf{Feasible} (F), \textbf{Completion} (C), and \textbf{Normalized Cost} (NC). Feasible is the fraction of runs with a successfully evaluated plan and no execution constraint violations. Completion is the average fraction of tasks validly delivered during evaluation. Invalid pickup or delivery actions do not update the task state, but evaluation continues, and valid deliveries are credited even if other actions violate constraints. Runs that cannot be planned or evaluated receive zero feasibility and completion. NC divides a plan's travelled distance by the length of the optimal pickup--delivery route over the subset of tasks that plan delivered. We evaluate stability under increasing planning horizon by varying task cardinality with $n \in \{5,8,10\}$ across five scene groups: \textit{open}, \textit{T1}, \textit{T2}, \textit{indoor}, and \textit{campus}.

\subsection{Benchmark Results}

DualWorldBench evaluates grounded planning beyond a single aggregate score. It decomposes model capability into three aspects: feasible multi-goal planning under increasing combinatorial difficulty (\textbf{Q1, Q2, Q4}), extraction and alignment of symbolic/visual evidence (\textbf{Q3}), and robustness across scene topology and environment complexity (\textbf{Q5}).

\begin{table*}[t!]
    \caption{TextWorld results under varying task counts. Gray shading indicates the best performance.}
    \resizebox{\textwidth}{!}{
    \begin{tabular}{cccccccccccccc}
    \toprule
    \multicolumn{1}{c|}{\multirow{3}{*}{Method}} & \multicolumn{1}{c|}{\multirow{3}{*}{Rank}} & \multicolumn{2}{c|}{\multirow{2}{*}{Average}} & \multicolumn{10}{c}{\cellcolor[HTML]{E0F2F1} Task Count}                                                                                         \\
    \multicolumn{1}{c|}{}                         & \multicolumn{1}{c|}{}                      & \multicolumn{2}{c|}{}                                     & \multicolumn{2}{c}{5} & \multicolumn{2}{c}{8} & \multicolumn{2}{c}{10} & \multicolumn{2}{c}{12} & \multicolumn{2}{c}{15} \\  \cmidrule(lr){3-4} \cmidrule(lr){5-6} \cmidrule(lr){7-8} \cmidrule(lr){9-10} \cmidrule(lr){11-12} \cmidrule(lr){13-14}
    \multicolumn{1}{c|}{}                         & \multicolumn{1}{c|}{}                      & R $\uparrow$               & \multicolumn{1}{c|}{T$\downarrow$}                 &  R $\uparrow$        & T $\downarrow$         & R $\uparrow$        & T $\downarrow$        & R $\uparrow$        & T $\downarrow$         & R $\uparrow$        & T $\downarrow$         & R $\uparrow$        & T $\downarrow$         \\ \hline
    \multicolumn{14}{c}{\cellcolor[HTML]{E6F2FF} Open-source LLMs}                                                        \\
    \multicolumn{1}{c|}{Qwen3.6-27B} & \multicolumn{1}{c|}{9} & 251.41 & \multicolumn{1}{c|}{5061.46} & 99.33 & 1969.57 & 213.17 & 5552.39 & 249.12 & 5377.14 & 312.94 & 5158.18 & 382.49 & 7250.02 \\
    \multicolumn{1}{c|}{Llama-2-13B} & \multicolumn{1}{c|}{12} & 212.33 & \multicolumn{1}{c|}{5515.20} & 100.28 & 2396.66 & 209.73 & 5471.45 & 202.32 & 5403.88 & 298.18 & 5912.93 & 251.15 & 8391.05 \\
    \multicolumn{1}{c|}{Qwen3-VL-8B} & \multicolumn{1}{c|}{13} & 68.81 & \multicolumn{1}{c|}{3305.46} & 19.23 & 1449.30 & 121.67 & 4481.61 & 35.69 & 2902.33 & 135.69 & 4653.41 & 31.79 & 3040.67 \\
    \multicolumn{1}{c|}{Llama-3-8B} & \multicolumn{1}{c|}{14} & 19.33 & \multicolumn{1}{c|}{3223.40} & 26.14 & 1640.78 & 13.94 & 3526.92 & 50.24 & 3609.83 & 6.33 & 3237.74 & 0.00 & 4101.73 \\ \hline
    \multicolumn{14}{c}{\cellcolor[HTML]{E6F2FF} Proprietary LLMs}          \\
    \multicolumn{1}{c|}{Gemini-2.5-Flash} & \multicolumn{1}{c|}{1} & 265.73 & \multicolumn{1}{c|}{4568.84} & 111.10 & 1875.90 & 213.24 & 5231.71 & 279.18 & 4175.72 & 333.12 & 4934.41 & 392.01 & 6626.44 \\
    \multicolumn{1}{c|}{Gemini-3-Pro} & \multicolumn{1}{c|}{2} & 264.36 & \multicolumn{1}{c|}{4310.73} & 114.77 & 1761.01 & 213.24 & 4843.20 & 278.53 & 3991.18 & 323.13 & 4716.06 & 392.14 & 6242.17 \\
    \multicolumn{1}{c|}{GPT-5.6-Sol} & \multicolumn{1}{c|}{3} & 264.33 & \multicolumn{1}{c|}{5013.85} & 114.77 & 2011.85 & 213.24 & 5384.76 & 278.53 & 4785.25 & 323.13 & 5665.16 & 392.01 & 7222.23 \\
    \multicolumn{1}{c|}{Claude-Opus-5} & \multicolumn{1}{c|}{4} & 264.10 & \multicolumn{1}{c|}{4347.24} & 114.77 & 1745.07 & 213.24 & 5058.77 & 277.35 & 4009.72 & 323.13 & 4695.04 & 392.01 & 6227.59 \\
    \multicolumn{1}{c|}{DeepSeek-v4-Flash} & \multicolumn{1}{c|}{5} & 263.60 & \multicolumn{1}{c|}{4711.47} & 111.10 & 1789.44 & 213.24 & 5437.33 & 278.53 & 4051.14 & 323.13 & 5406.74 & 392.01 & 6872.71 \\
    \multicolumn{1}{c|}{Qwen3.7-Plus} & \multicolumn{1}{c|}{6} & 263.24 & \multicolumn{1}{c|}{4560.71} & 114.77 & 1863.30 & 213.24 & 5094.50 & 273.90 & 4165.28 & 322.27 & 5185.82 & 392.01 & 6494.65 \\
    \multicolumn{1}{c|}{Claude-Sonnet-5} & \multicolumn{1}{c|}{7} & 260.16 & \multicolumn{1}{c|}{4780.29} & 114.54 & 1782.60 & 196.24 & 5408.39 & 278.43 & 4275.89 & 319.52 & 5634.78 & 392.05 & 6799.79 \\
    \multicolumn{1}{c|}{Claude-4.5-Haiku} & \multicolumn{1}{c|}{8} & 253.01 & \multicolumn{1}{c|}{4900.80} & 111.10 & 1927.15 & 204.95 & 5333.92 & 266.26 & 4818.51 & 304.06 & 5532.84 & 378.70 & 6891.58 \\
    \multicolumn{1}{c|}{GPT-4o} & \multicolumn{1}{c|}{10} & 251.05 & \multicolumn{1}{c|}{5245.35} & 114.77 & 2226.95 & 189.94 & 5808.08 & 272.77 & 4805.22 & 303.95 & 5662.43 & 373.82 & 7724.05 \\
    \multicolumn{1}{c|}{GPT-5.4-Mini} & \multicolumn{1}{c|}{11} & 246.81 & \multicolumn{1}{c|}{5207.10} & 111.10 & 2219.29 & 198.84 & 5532.03 & 248.54 & 4925.06 & 321.70 & 5662.36 & 353.88 & 7696.76 \\ \hline
    \multicolumn{14}{c}{\cellcolor[HTML]{E6F2FF} Combinatorial Optimization Solver}                                                                                                                                                                                                                            \\
    \multicolumn{1}{c|}{DFS}                      & \multicolumn{1}{c|}{3}                     & 264.33           & \multicolumn{1}{c|}{4702.45}           & 114.77    & 2028.71   & 213.24    & 5349.95   & 278.53    & 4344.86    & 323.13    & 5186.65    & 392.01    & 6602.05    \\
    \multicolumn{1}{c|}{Online DP}                & \multicolumn{1}{c|}{2}                     & 264.36           & \multicolumn{1}{c|}{4083.25}           & 114.77    & 1758.71   & 213.24    & 4742.61   & 278.53    & 3806.96    & 323.13    & 4479.11    & 392.14    & 5628.85    \\
    \multicolumn{1}{c|}{Offline DP}               & \multicolumn{1}{c|}{1}                     & \best{278.17}           & \multicolumn{1}{c|}{\best{4015.28}}           & \best{126.20}    & \best{1718.67}   & \best{230.27}    & \best{4602.62}   & \best{290.00}    & \best{3794.25}    & \best{334.65}    & \best{4406.57}    & \best{409.71}    & \best{5554.29}   
    \\ \bottomrule
    \end{tabular}}
    \label{table:tasknum}
\end{table*}

\begin{figure}[t]
    \centering
    \begin{subfigure}[t]{0.49\textwidth}
        \centering
        \includegraphics[width=\linewidth]
            {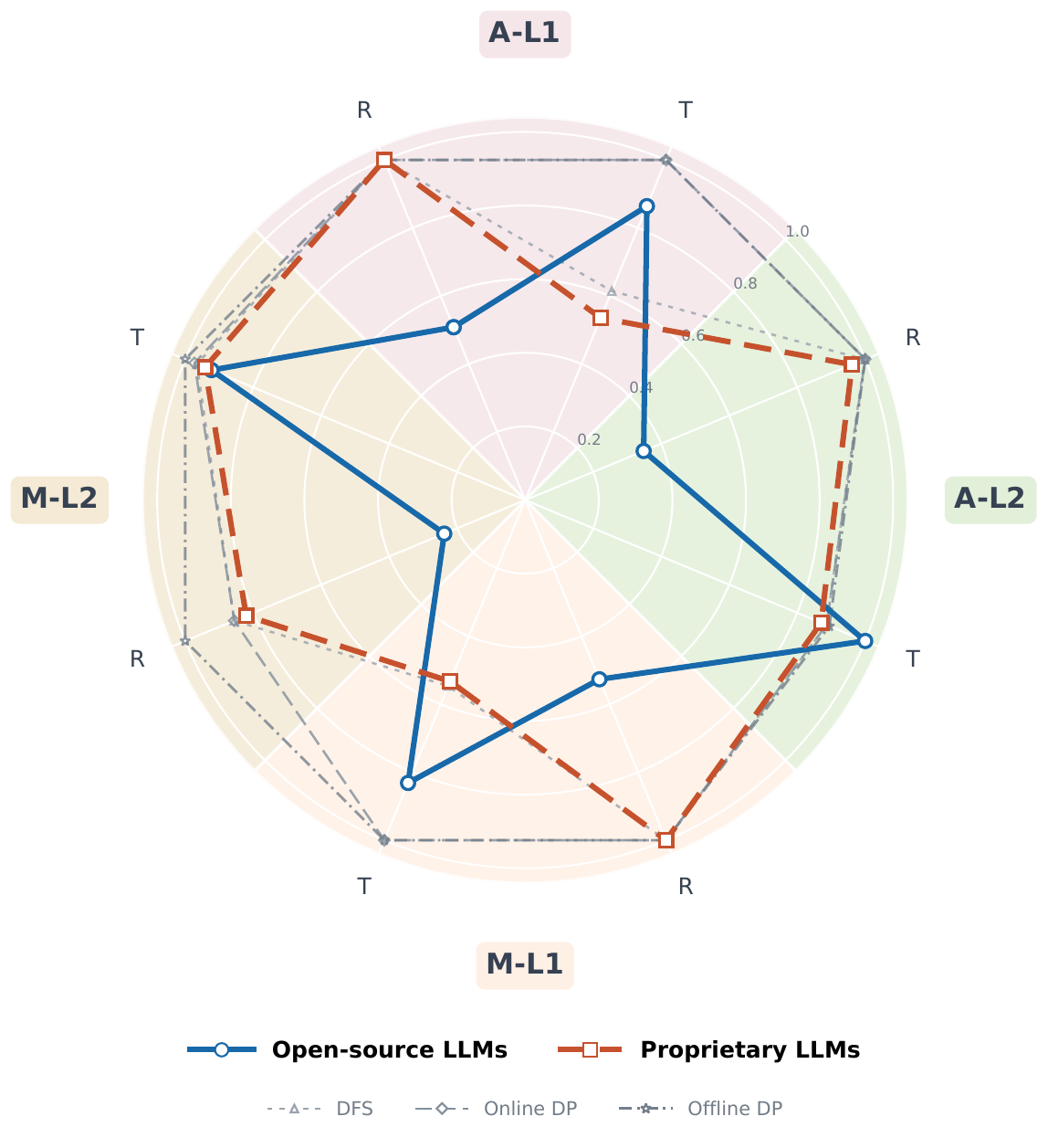}
        \caption{TextWorld (task count 15).}
        \label{fig:radar_textworld}
    \end{subfigure}
    \hfill
    \begin{subfigure}[t]{0.49\textwidth}
        \centering
        \includegraphics[width=\linewidth]
            {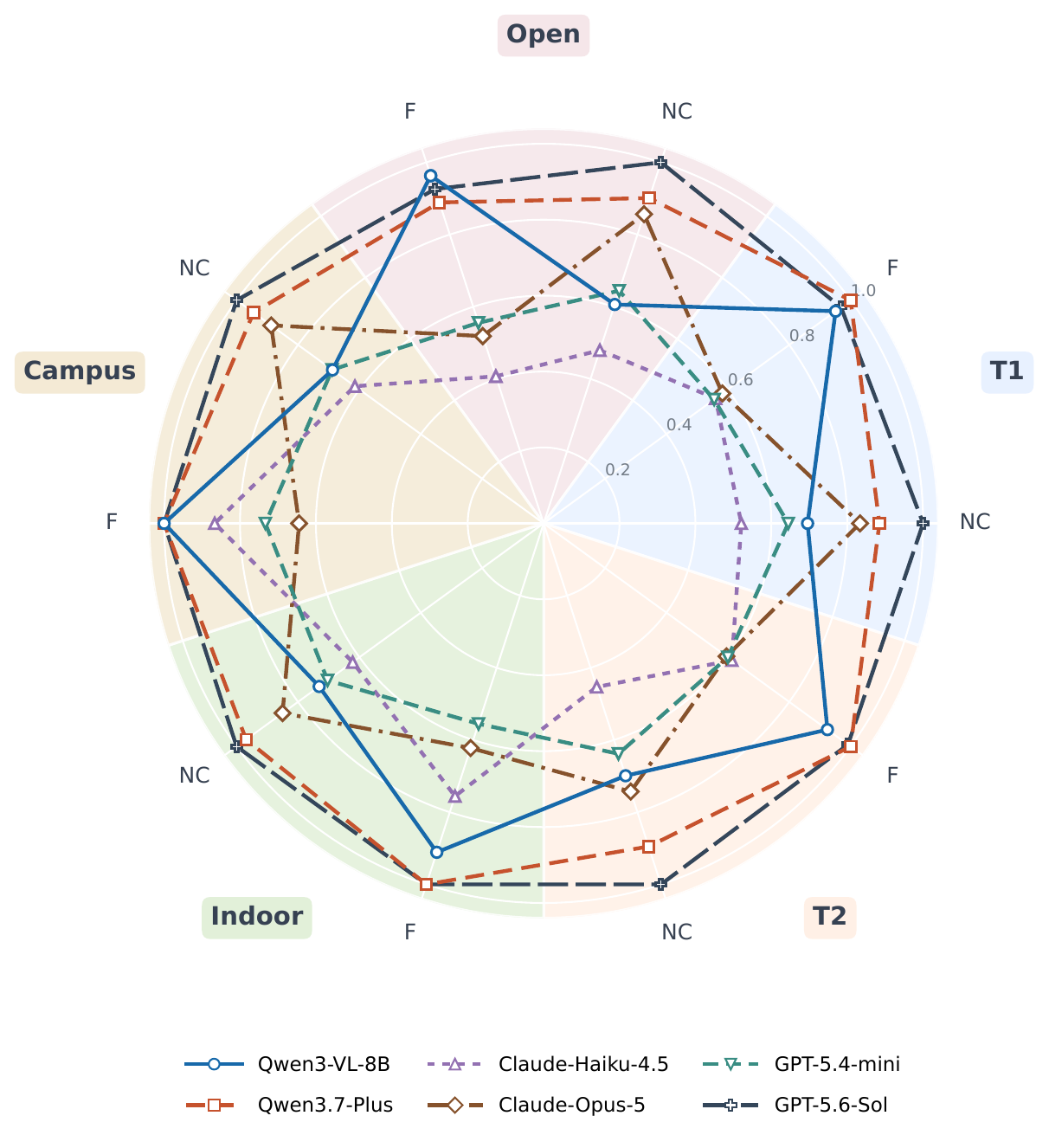}
        \caption{SimWorld (averaged across modalities).}
        \label{fig:radar_simworld}
    \end{subfigure}

    \caption{
        Performance comparison across TextWorld and SimWorld. In TextWorld, reward and completion time are normalized as
        $R/\max R$ and $\min T/T$. In SimWorld, scores are averaged across modalities, with $NC$ inversely normalized within each scene. 
    }
    \label{fig:radar_comparison}
\end{figure}

\subsubsection{TextWorld}

Table~\ref{table:tasknum} and Fig.~\ref{fig:radar_textworld} show TextWorld results across task counts and symbolic-temporal conditions.

\paragraph{Q1: How well do planners scale as the symbolic--temporal search space grows?}
Table~\ref{table:tasknum} shows that frontier models can already generate complete task routes and consequently obtain high task rewards. \texttt{Gemini-2.5-Flash} achieves the best average performance among the MLLMs ($R{=}265.7$, $T{=}4568.8$). Although its reward slightly exceeds that of Online DP ($R{=}264.3$), its simulated completion time remains longer than those of the solvers, indicating that even the strongest models cannot identify the optimal routes. The performance of open-source models is considerably more dispersed: at $n{=}15$, \texttt{Qwen3.6-27B} achieves $R{=}382.4$, whereas \texttt{Qwen3-VL-8B} drops to $R{=}31.7$ and \texttt{Llama-3-8B} to $R{=}0.0$. Thus, as the number of tasks and the search space increase, models exhibit clear performance divergence. Nevertheless, even the best-performing models remain unable to produce optimal task plans that achieve the highest reward within the shortest completion time.

\paragraph{Q2: How robust are planners under richer symbolic constraints and online task release?}
Fig.~\ref{fig:radar_textworld} reveals a substantial capability gap between the two model groups; Appendix Table~\ref{tab:app_textworld_task15_full} provides the complete model-wise results. Proprietary LLMs achieve an average reward of $R{=}385.0$ across the four settings, compared with only $166.3$ for open-source LLMs, and reach the reward ceiling in both A-L1 and M-L1. This gap persists under online task release. In A-L2, proprietary models retain $96.1\%$ of the Offline-DP reward ($278.5$ vs.\ $289.7$), whereas open-source models retain only $34.8\%$ ($100.9$). M-L2 is the most challenging setting: proprietary models achieve an average reward of $R{=}399.8$, corresponding to $82.1\%$ of the Offline-DP reward, whereas open-source models achieve only $116.0$ ($23.8\%$). The primary bottleneck therefore lies not merely in interpreting individual constraints, but in jointly reasoning over heterogeneous priorities and deadlines while dynamically replanning as new tasks become available.

\begin{table*}[t!]
    \centering
    \caption{SimWorld scaling results under varying task counts.}
    \label{tab:simworld_task_scaling}
    \resizebox{\textwidth}{!}{
    \begin{tabular}{c|c|ccc|ccccccccc}
    \toprule
    \multirow{3}{*}{Method}    & \multirow{3}{*}{Modality} & \multicolumn{3}{c|}{\multirow{2}{*}{Average}} & \multicolumn{9}{c}{\cellcolor[HTML]{E0F2F1} Task Count}   \\
     &     & \multicolumn{3}{c|}{}                                                                   & \multicolumn{3}{c}{5} & \multicolumn{3}{c}{8} & \multicolumn{3}{c}{10} \\  \cmidrule(lr){3-5} \cmidrule(lr){6-8} \cmidrule(lr){9-11} \cmidrule(lr){12-14} &  & F$\uparrow$                 & C$\uparrow$                 & NC$\downarrow$              & F$\uparrow$   & C$\uparrow$   & \multicolumn{1}{c|}{NC$\downarrow$}   & F$\uparrow$   & C$\uparrow$   & \multicolumn{1}{c|}{NC$\downarrow$}   & F$\uparrow$           & C$\uparrow$          & NC$\downarrow$          \\ \hline
    \multirow{3}{*}{Qwen3-VL-8B}
    & Multi.
    & 0.989 & 0.974 & 1.90
    & 0.984 & 0.946 & \multicolumn{1}{c|}{1.56}
    & 0.984 & 0.976 & \multicolumn{1}{c|}{1.95}
    & 1.000 & 1.000 & 2.20 \\
    & Image
    & 0.836 & 0.904 & 1.90
    & 0.778 & 0.895 & \multicolumn{1}{c|}{1.58}
    & 0.857 & 0.907 & \multicolumn{1}{c|}{1.95}
    & 0.873 & 0.911 & 2.17 \\
    & Text
    & 1.000 & 0.544 & 1.46
    & 1.000 & 0.527 & \multicolumn{1}{c|}{1.28}
    & 1.000 & 0.619 & \multicolumn{1}{c|}{1.56}
    & 1.000 & 0.486 & 1.53 \\
    \hline
    \multirow{3}{*}{Qwen3.7-Plus}
    & Multi.
    & \best{1.000} & \best{1.000} & 1.27
    & 1.000 & 1.000 & \multicolumn{1}{c|}{1.13}
    & 1.000 & 1.000 & \multicolumn{1}{c|}{1.30}
    & \best{1.000} & \best{1.000} & 1.38 \\
    & Image
    & 0.984 & 0.995 & 1.54
    & 0.984 & 0.997 & \multicolumn{1}{c|}{1.32}
    & 0.984 & 0.996 & \multicolumn{1}{c|}{1.56}
    & 0.984 & 0.992 & 1.74 \\
    & Text
    & 1.000 & 0.972 & 1.26
    & \best{1.000} & \best{1.000} & \multicolumn{1}{c|}{1.12}
    & \best{1.000} & \best{1.000} & \multicolumn{1}{c|}{1.23}
    & 1.000 & 0.916 & 1.42 \\
    \hline
    \multirow{3}{*}{Claude-Haiku-4.5}
    & Multi.
    & 0.762 & 0.961 & 2.16
    & 0.921 & 0.987 & \multicolumn{1}{c|}{1.57}
    & 0.698 & 0.964 & \multicolumn{1}{c|}{2.49}
    & 0.667 & 0.930 & 2.43 \\
    & Image
    & 0.175 & 0.707 & 3.19
    & 0.270 & 0.752 & \multicolumn{1}{c|}{2.69}
    & 0.159 & 0.708 & \multicolumn{1}{c|}{3.48}
    & 0.095 & 0.660 & 3.40 \\
    & Text
    & 0.905 & 0.981 & 1.93
    & 1.000 & 1.000 & \multicolumn{1}{c|}{1.49}
    & 0.810 & 0.972 & \multicolumn{1}{c|}{1.95}
    & 0.905 & 0.971 & 2.35 \\
    \hline
    \multirow{3}{*}{Claude-Opus-5}
    & Multi.
    & 0.519 & 0.945 & 1.74
    & 0.730 & 0.968 & \multicolumn{1}{c|}{1.36}
    & 0.429 & 0.933 & \multicolumn{1}{c|}{1.78}
    & 0.397 & 0.935 & 2.08 \\
    & Image
    & 0.561 & 0.865 & 1.45
    & 0.635 & 0.902 & \multicolumn{1}{c|}{1.30}
    & 0.571 & 0.859 & \multicolumn{1}{c|}{1.46}
    & 0.476 & 0.835 & 1.59 \\
    & Text
    & 0.698 & 0.971 & 1.38
    & 0.857 & 0.994 & \multicolumn{1}{c|}{1.23}
    & 0.683 & 0.972 & \multicolumn{1}{c|}{1.38}
    & 0.556 & 0.948 & 1.52 \\
    \hline
    \multirow{3}{*}{GPT-5.4-mini}
    & Multi.
    & 0.640 & 0.934 & 1.81
    & 0.778 & 0.952 & \multicolumn{1}{c|}{1.52}
    & 0.698 & 0.948 & \multicolumn{1}{c|}{1.88}
    & 0.444 & 0.900 & 2.04 \\
    & Image
    & 0.127 & 0.695 & 2.03
    & 0.143 & 0.698 & \multicolumn{1}{c|}{1.95}
    & 0.175 & 0.742 & \multicolumn{1}{c|}{1.98}
    & 0.063 & 0.644 & 2.17 \\
    & Text
    & 0.995 & 0.966 & 1.78
    & 1.000 & 1.000 & \multicolumn{1}{c|}{1.55}
    & 0.984 & 0.998 & \multicolumn{1}{c|}{1.85}
    & 1.000 & 0.900 & 1.95 \\
    \hline
    \multirow{3}{*}{GPT-5.6-Sol}
    & Multi.
    & 0.984 & 0.998 & 1.13
    & 0.984 & 0.997 & \multicolumn{1}{c|}{1.08}
    & 0.984 & 1.000 & \multicolumn{1}{c|}{1.16}
    & 0.984 & 0.997 & \best{1.17} \\
    & Image
    & 0.989 & 0.996 & 1.39
    & 0.984 & 0.994 & \multicolumn{1}{c|}{1.25}
    & 1.000 & 1.000 & \multicolumn{1}{c|}{1.39}
    & 0.984 & 0.995 & 1.53 \\
    & Text
    & 0.968 & 0.995 & \best{1.13}
    & 0.968 & 0.994 & \multicolumn{1}{c|}{\best{1.07}}
    & 0.937 & 0.992 & \multicolumn{1}{c|}{\best{1.13}}
    & 1.000 & 0.998 & 1.20 \\
    \bottomrule
    \end{tabular}}
\end{table*}

\subsubsection{SimWorld}

Tables~\ref{tab:simworld_task_scaling} and Fig.~\ref{fig:radar_simworld} report SimWorld results across task scales, input modalities, and scene groups.

\paragraph{Q3: Can MLLMs align textual and visual information into feasible grounded plans?}
Grounded planning remains strongly dependent on both the model and input modality. Text input generally produces more feasible plans, whereas image-only planning is substantially more challenging for weaker models. However, this ordering is not universal: stronger models remain reliable with visual observations, suggesting that the main limitation is not the visual representation itself, but the model's ability to convert visual evidence into a constraint-consistent plan. Moreover, feasibility does not necessarily imply adequate task coverage. For example, \texttt{Qwen3-VL-8B} produces consistently feasible text-based plans but omits many tasks, while its multimodal plans achieve a better balance between legality and completion. Overall, multimodal input is most useful when the model can jointly preserve grounding accuracy, task coverage, and global plan consistency; visual input alone does not provide a systematic routing advantage.

\paragraph{Q4: How does grounded planning scale with task count across modalities?}
Increasing the task count reveals a clear separation between task completion and plan validity. Several models maintain high completion while losing feasibility under multimodal input, indicating that they continue to include the requested goals but increasingly violate connectivity or ordering constraints. In contrast, \texttt{GPT-5.6-Sol} and the Qwen models remain comparatively stable, showing that this degradation is not unavoidable and depends strongly on planning capability. Nevertheless, NC generally increases with task count across all modalities, including for models whose feasibility remains nearly unchanged. Thus, larger task sets challenge route efficiency before they necessarily cause complete planning failure. These results identify global goal ordering and constraint management, rather than individual target recognition, as the central difficulty in scaling grounded planning.

\paragraph{Q5: How robust is grounded planning across scene topology?}
Scene topology primarily affects whether a valid plan can be constructed, while its effect on routing efficiency is comparatively moderate. Open scenes are generally more difficult in terms of feasibility, whereas campus scenes tend to be easier, but this trend is dominated by models with weaker visual grounding. In particular, image-only performance varies substantially across scene groups for these models, while stronger models remain reliable across most topologies. Text-based planning is considerably more stable, further indicating that the main challenge lies in recovering planning-relevant structure from visual observations. Therefore, robustness to scene variation depends on the interaction between topology and visual grounding capability, rather than on scene type alone.

\begin{figure}[t]
  \centering
  \begin{subfigure}[c]{0.66\textwidth}
    \centering
    \includegraphics[width=\linewidth]{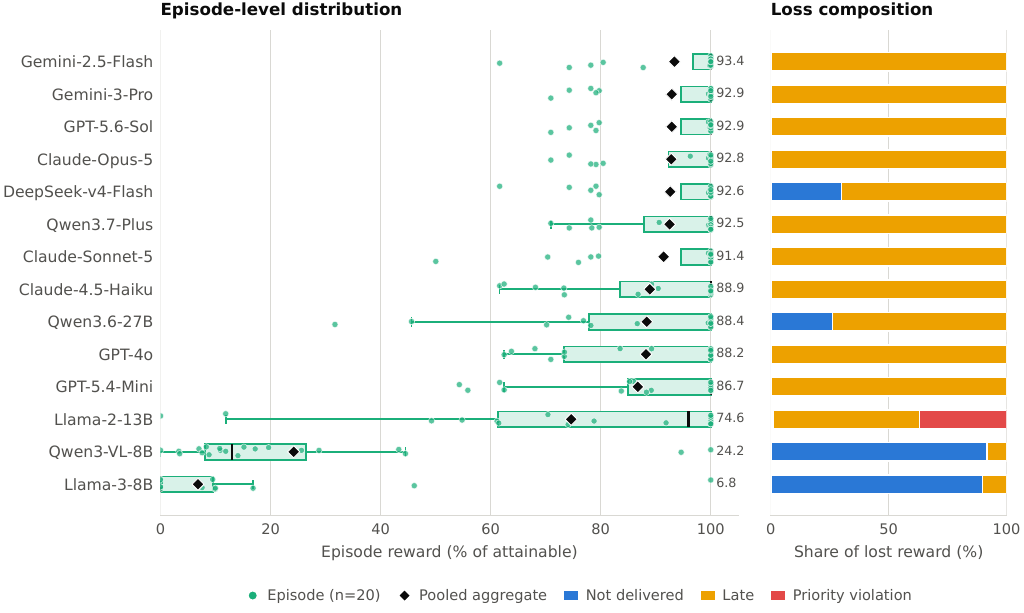}
    \caption{TextWorld reward-gap attribution.}
    \label{fig:err_textworld}
  \end{subfigure}\hfill
  \begin{subfigure}[c]{0.31\textwidth}
    \centering
    \includegraphics[width=0.96\linewidth]{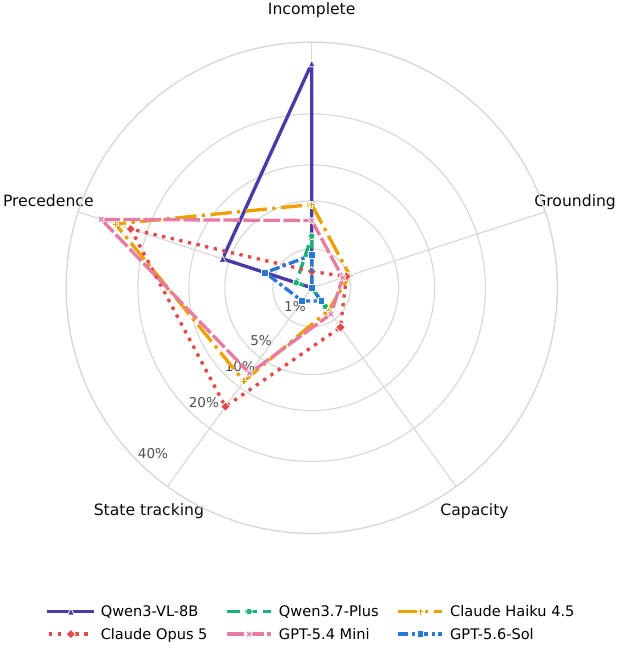}
    \caption{SimWorld root-cause profile.}
    \label{fig:err_simworld}
  \end{subfigure}
  \caption{\textbf{Failure attribution across the two worlds.} \textbf{(a)} TextWorld episode reward and loss composition. \textbf{(b)} SimWorld root-cause shares pooled over the three input modalities for each model.}
  \label{fig:error_attribution}
\end{figure}

\subsection{Error Analysis}
\label{ex:error_analysis}

We use exhaustive, mutually exclusive error taxonomies to explain performance losses: TextWorld decomposes the reward gap, while SimWorld assigns each run a single root cause.

\paragraph{TextWorld: from coverage to scheduling.}
We attribute lost reward to \emph{not delivered} (forfeited reward), \emph{late} (deadline penalties), or \emph{priority violation} (run invalidation), and aggregate reward-weighted shares across scenarios and task counts.
Figure~\ref{fig:err_textworld} shows that frontier models obtain over $92\%$ of attainable reward, with lateness accounting for most residual loss. Lateness dominates under temporal constraints and vanishes without deadlines, while priority violations occur in only one of $280$ runs. Weaker models instead lose reward mainly through undelivered tasks, indicating a shift from task coverage to deadline-aware scheduling as model capability improves.

\paragraph{SimWorld: action sequencing dominates.}
We assign root causes in the order \emph{grounding} $>$ \emph{capacity} $>$ \emph{state tracking} $>$ \emph{precedence}, labeling violation-free runs as \emph{success} or \emph{incomplete}. Derived symptoms, such as \texttt{deliver without carry}, are subsumed under their underlying sequencing errors.
As shown in Figure~\ref{fig:err_simworld}, precedence and state-tracking violations account for most invalid plans; grounding and capacity errors are uncommon. These results identify action sequencing and deadline-aware scheduling as key remaining challenges.

\begin{wrapfigure}{r}{0.5\textwidth}
  \centering
  \includegraphics[width=\linewidth]{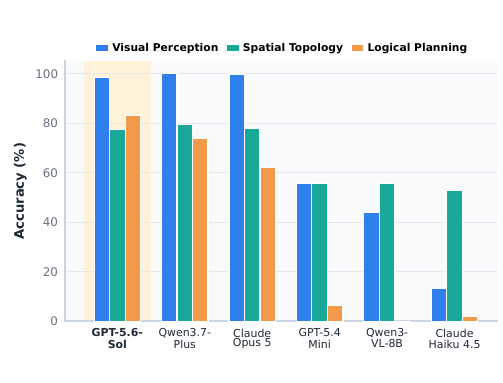}
  \caption{Evaluation in DualWorld-VQA from three levels: Visual Perception, Spatial Topology, and Logical Planning.}
  \label{fig:vqa}
\end{wrapfigure}

\subsection{DualWorld-VQA Results}
\label{ex:qa_results}

Figure~\ref{fig:vqa} shows that the three-level capability decomposition clearly distinguishes models across capability tiers. Weaker models exhibit a sharp gap between intermediate understanding and end-to-end planning. For example, \texttt{GPT-5.4-mini} achieves $55.5\%$ and $55.9\%$ accuracy in visual perception and spatial topology, respectively, but only $6.3\%$ in logical planning. This result indicates that partial competence in perception and topology does not necessarily enable a model to integrate visual and symbolic information into constraint-consistent plans. This diagnosis is consistent with the SimWorld results, where \texttt{GPT-5.4-mini} maintains relatively high task completion but exhibits low feasibility, particularly with image-only input, and degrades further as the task count increases. In contrast, stronger flagship models approach saturation in visual perception and avoid the pronounced collapse in logical planning observed for weaker models. Their SimWorld performance is similarly robust across modalities, task scales, and scene topologies, suggesting an emerging ability to jointly reason over visual evidence, spatial structure, and symbolic constraints. Together, the VQA and SimWorld results consistently show that the central bottleneck is not perception or topology understanding alone, but their composition into globally feasible grounded plans.

\section{Conclusion}

We introduced \textbf{DualWorldBench}, a controlled benchmark for agent planning in symbolic--temporal and grounded--topological settings. Through a shared pickup--delivery interface, \textbf{TextWorld}, \textbf{SimWorld}, and \textbf{DualWorld-VQA} combine end-to-end evaluation with fine-grained diagnosis. Stronger models improve task coverage and grounded feasibility, but scheduling and consistent action sequencing remain bottlenecks. DualWorld-VQA further suggests that translating perception and topology understanding into executable plans remains challenging.
The benchmark currently evaluates temporal and topological complexity separately, motivating future work on their integration within shared instances.

\section*{Acknowledgments and Disclosure of Funding}

We appreciate the anonymous reviewers for their valuable insights and helpful comments. This work is supported by the National Natural Science Foundation of China (Grant No. 62306133), the Key Program of Jiangsu Science Foundation (BK20243012), and the ``111 Center'' (No. B26023).

\bibliography{references}

@article{webshop,
  title={WebShop: Towards scalable real-world web interaction with grounded language agents},
  author={Yao, Shunyu and Chen, Howard and Yang, John and Narasimhan, Karthik},
  journal={Advances in Neural Information Processing Systems},
  pages={20744--20757},
  year={2022}
}

@article{mind2web,
  title={Mind2Web: Towards a generalist agent for the web},
  author={Deng, Xiang and Gu, Yu and Zheng, Boyuan and Chen, Shijie and Stevens, Sam and Wang, Boshi and Sun, Huan and Su, Yu},
  journal={Advances in Neural Information Processing Systems},
  pages={28091--28114},
  year={2023}
}

@article{webarena,
  title={WebArena: A realistic web environment for building autonomous agents},
  author={Zhou, Shuyan and Xu, Frank F and Zhu, Hao and Zhou, Xuhui and Lo, Robert and Sridhar, Abishek and Cheng, Xianyi and Ou, Tianyue and Bisk, Yonatan and Fried, Daniel and others},
  journal={arXiv preprint arXiv:2307.13854},
  year={2023}
}

@article{webembodied,
  title={Embodied Web Agents: Bridging physical-digital realms for integrated agent intelligence},
  author={Hong, Yining and Sun, Rui and Li, Bingxuan and Yao, Xingcheng and Wu, Maxine and Chien, Alexander and Yin, Da and Wu, Ying Nian and Wang, Zhecan James and Chang, Kai-Wei},
  journal={arXiv preprint arXiv:2506.15677},
  year={2025}
}

@inproceedings{visualwebarena,
  title={VisualWebArena: Evaluating multimodal agents on realistic visual web tasks},
  author={Koh, Jing Yu and Lo, Robert and Jang, Lawrence and Duvvur, Vikram and Lim, Ming and Huang, Po-Yu and Neubig, Graham and Zhou, Shuyan and Salakhutdinov, Russ and Fried, Daniel},
  booktitle={Proceedings of the 62nd Annual Meeting of the Association for Computational Linguistics},
  pages={881--905},
  year={2024}
}

@article{deliverybench,
  title={DeliveryBench: Can Agents Earn Profit in Real World?},
  author={Mao, Lingjun and Ren, Jiawei and Zhou, Kun and Chen, Jixuan and Ma, Ziqiao and Qin, Lianhui},
  journal={arXiv preprint arXiv:2512.19234},
  year={2025}
}

@article{cvbench,
  title={CVBench: Benchmarking Cross-Video Synergies for Complex Multimodal Reasoning},
  author={Zhu, Nannan and Dong, Yonghao and Wang, Teng and Li, Xueqian and Deng, Shengjun and Wang, Yijia and Hong, Zheng and Geng, Tiantian and Niu, Guo and Huang, Hanyan and others},
  journal={arXiv preprint arXiv:2508.19542},
  year={2025}
}

@inproceedings{embspatial,
  title={EmbSpatial-Bench: Benchmarking spatial understanding for embodied tasks with large vision-language models},
  author={Du, Mengfei and Wu, Binhao and Li, Zejun and Huang, Xuan-Jing and Wei, Zhongyu},
  booktitle={Proceedings of the 62nd Annual Meeting of the Association for Computational Linguistics},
  pages={346--355},
  year={2024}
}

@inproceedings{alfred,
  title={ALFRED: A benchmark for interpreting grounded instructions for everyday tasks},
  author={Shridhar, Mohit and Thomason, Jesse and Gordon, Daniel and Bisk, Yonatan and Han, Winson and Mottaghi, Roozbeh and Zettlemoyer, Luke and Fox, Dieter},
  booktitle={Proceedings of the IEEE/CVF Conference on Computer Vision and Pattern Recognition},
  pages={10740--10749},
  year={2020}
}

@inproceedings{embodiedbench,
  title={EmbodiedBench: Comprehensive Benchmarking Multi-modal Large Language Models for Vision-Driven Embodied Agents},
  author={Yang, Rui and Chen, Hanyang and Zhang, Junyu and Zhao, Mark and Qian, Cheng and Wang, Kangrui and Wang, Qineng and Koripella, Teja Venkat and Movahedi, Marziyeh and Li, Manling and others},
  booktitle={International Conference on Machine Learning},
  pages={70576--70631},
  year={2025},
}

@article{eai,
  title={Embodied Agent Interface: Benchmarking llms for embodied decision making},
  author={Li, Manling and Zhao, Shiyu and Wang, Qineng and Wang, Kangrui and Zhou, Yu and Srivastava, Sanjana and Gokmen, Cem and Lee, Tony and Li, Li E and Zhang, Ruohan and others},
  journal={Advances in Neural Information Processing Systems},
  pages={100428--100534},
  year={2024}
}

@article{embodiedarena,
  title={Embodied Arena: A comprehensive, unified, and evolving evaluation platform for embodied ai},
  author={Ni, Fei and Zhang, Min and Li, Pengyi and Yuan, Yifu and Zhang, Lingfeng and Liu, Yuecheng and Han, Peilong and Kou, Longxin and Ma, Shaojin and Qiao, Jinbin and others},
  journal={arXiv preprint arXiv:2509.15273},
  year={2025}
}

@inproceedings{localplanner,
  title={Autonomous exploration development environment and the planning algorithms},
  author={Cao, Chao and Zhu, Hongbiao and Yang, Fan and Xia, Yukun and Choset, Howie and Oh, Jean and Zhang, Ji},
  booktitle={2022 International Conference on Robotics and Automation},
  pages={8921--8928},
  year={2022},
}

@inproceedings{farplanner,
  title={FAR Planner: Fast, attemptable route planner using dynamic visibility update},
  author={Yang, Fan and Cao, Chao and Zhu, Hongbiao and Oh, Jean and Zhang, Ji},
  booktitle={2022 IEEE/RSJ International Conference on Intelligent Robots and Systems},
  pages={9--16},
  year={2022},
}

@inproceedings{teach,
  title={Teach: Task-driven embodied agents that chat},
  author={Padmakumar, Aishwarya and Thomason, Jesse and Shrivastava, Ayush and Lange, Patrick and Narayan-Chen, Anjali and Gella, Spandana and Piramuthu, Robinson and Tur, Gokhan and Hakkani-Tur, Dilek},
  booktitle={Proceedings of the AAAI Conference on Artificial Intelligence},
  pages={2017--2025},
  year={2022}
}

@inproceedings{behavior,
  title={Behavior-1k: A benchmark for embodied ai with 1,000 everyday activities and realistic simulation},
  author={Li, Chengshu and Zhang, Ruohan and Wong, Josiah and Gokmen, Cem and Srivastava, Sanjana and Mart{\'\i}n-Mart{\'\i}n, Roberto and Wang, Chen and Levine, Gabrael and Lingelbach, Michael and Sun, Jiankai and others},
  booktitle={Conference on Robot Learning},
  pages={80--93},
  year={2023},
}

@article{vsr,
  title={Visual spatial reasoning},
  author={Liu, Fangyu and Emerson, Guy and Collier, Nigel},
  journal={Transactions of the Association for Computational Linguistics},
  pages={635--651},
  year={2023},
}

@inproceedings{Spatialmm,
  title={An empirical analysis on spatial reasoning capabilities of large multimodal models},
  author={Shiri, Fatemeh and Guo, Xiao-Yu and Far, Mona Golestan and Yu, Xin and Haf, Reza and Li, Yuan-Fang},
  booktitle={Proceedings of the 2024 Conference on Empirical Methods in Natural Language Processing},
  pages={21440--21455},
  year={2024}
}

@article{eb1,
  title={LoTa-Bench: Benchmarking language-oriented task planners for embodied agents},
  author={Choi, Jae-Woo and Yoon, Youngwoo and Ong, Hyobin and Kim, Jaehong and Jang, Minsu},
  journal={arXiv preprint arXiv:2402.08178},
  year={2024}
}

@article{web1,
  title={AgentBench: Evaluating llms as agents},
  author={Liu, Xiao and Yu, Hao and Zhang, Hanchen and Xu, Yifan and Lei, Xuanyu and Lai, Hanyu and Gu, Yu and Ding, Hangliang and Men, Kaiwen and Yang, Kejuan and others},
  journal={arXiv preprint arXiv:2308.03688},
  year={2023}
}

@article{spatial1,
  title={VisuLogic: A benchmark for evaluating visual reasoning in multi-modal large language models},
  author={Xu, Weiye and Wang, Jiahao and Wang, Weiyun and Chen, Zhe and Zhou, Wengang and Yang, Aijun and Lu, Lewei and Li, Houqiang and Wang, Xiaohua and Zhu, Xizhou and others},
  journal={arXiv preprint arXiv:2504.15279},
  year={2025}
}

@article{spatial2,
  title={BLINK-Twice: You see, but do you observe? A Reasoning Benchmark on Visual Perception},
  author={Ye, Junyan and Jiang, Dongzhi and He, Jun and Zhou, Baichuan and Huang, Zilong and Yan, Zhiyuan and Li, Hongsheng and He, Conghui and Li, Weijia},
  journal={arXiv preprint arXiv:2510.09361},
  year={2025}
}

@article{delivery2,
  title={Human-centered In-building Embodied Delivery Benchmark},
  author={Xu, Zhuoqun and Liu, Yang and Li, Xiaoqi and Zhang, Jiyao and Dong, Hao},
  journal={arXiv preprint arXiv:2406.17898},
  year={2024}
}

@inproceedings{delivery3,
  title={Multi-goal multi-agent pickup and delivery},
  author={Xu, Qinghong and Li, Jiaoyang and Koenig, Sven and Ma, Hang},
  booktitle={2022 IEEE/RSJ International Conference on Intelligent Robots and Systems},
  pages={9964--9971},
  year={2022}
}

@article{travel1,
  title={TravelPlanner: A benchmark for real-world planning with language agents},
  author={Xie, Jian and Zhang, Kai and Chen, Jiangjie and Zhu, Tinghui and Lou, Renze and Tian, Yuandong and Xiao, Yanghua and Su, Yu},
  journal={arXiv preprint arXiv:2402.01622},
  year={2024}
}

@article{travel2,
  title={ChinaTravel: An open-ended benchmark for language agents in chinese travel planning},
  author={Shao, Jie-Jing and Zhang, Bo-Wen and Yang, Xiao-Wen and Chen, Baizhi and Han, Si-Yu and Wei, Wen-Da and Cai, Guohao and Dong, Zhenhua and Guo, Lan-Zhe and Li, Yu-Feng},
  journal={arXiv preprint arXiv:2412.13682},
  year={2024}
}

@article{planbench,
  title={PlanBench: An extensible benchmark for evaluating large language models on planning and reasoning about change},
  author={Valmeekam, Karthik and Marquez, Matthew and Olmo, Alberto and Sreedharan, Sarath and Kambhampati, Subbarao},
  journal={Advances in Neural Information Processing Systems},
  pages={38975--38987},
  year={2023}
}

@article{harnessvln,
  title={HarnessVLN: Unifying Training-Free Embodied Navigation through an Agent Harness},
  author={Chen, Yang and Che, Lirong and Huang, Zhenyu and Fu, Wenbo and Wang, Chuang and Cao, Xu and Liu, Daqi and Yang, Yuzhe and Su, Jian and Guo, Lan-Zhe},
  journal={arXiv preprint arXiv:2609.15195},
  year={2026}
}

@article{navharness,
  author = {Zhao, Xunyi and Zhou, Jian and Lin, Sihao and Zhou, Gengze and Li, Zerui and Yan, Xinyu and Liu, Jiajun and van den Hengel, Anton and Wu, Qi},
  title = {{NavHarness}: Towards Lifelong Embodied Navigation},
  journal = {arXiv preprint arXiv:2609.34276},
  year = {2026}
}

@article{nesy1,
  title={Self-Evolving Neuro-Symbolic Skills for Tool-Augmented Spatial Reasoning},
  author={Tian, Shi-Yu and Wang, Zhuo-Xia and Zhu, Xuan-Yi and Zhou, Zhi and Yang, Xinwei and Yu, Kun-Yang and Yang, Ming and Chen, Yang and Li, Yu-Feng},
  journal={arXiv preprint arXiv:2608.07955},
  year={2026}
}

@article{maptab,
  title={MapTab: Are MLLMs Ready for Multi-Criteria Route Planning in Heterogeneous Graphs?},
  author={Shang, Ziqiao and Ge, Lingyue and Cheng, Zi-Jian and Tian, Shi-Yu and Huang, Zhenyu and Fu, Wenbo and Wu, Weiming and Chen, Yang and Zhang, Xiangwen and Hu, Yulan and others},
  journal={arXiv preprint arXiv:2602.18600},
  year={2026}
}

@article{nesy2,
  title={Aligning progress and feasibility: A neuro-symbolic dual memory framework for long-horizon llm agents},
  author={Wen, Bin and Zhang, Ruoxuan and Chen, Yang and Xie, Hongxia and Guo, Lan-Zhe},
  journal={arXiv preprint arXiv:2604.02734},
  year={2026}
}

@article{nesy3,
  title={Lifting traces to logic: Programmatic skill induction with neuro-symbolic learning for long-horizon agentic tasks},
  author={Shao, Jie-Jing and Yin, Haiyan and Lyu, Yueming and Yu, Xingrui and Guo, Lan-Zhe and Tsang, Ivor and Kwok, James and Li, Yu-Feng},
  journal={arXiv preprint arXiv:2605.01293},
  year={2026}
}

@inproceedings{nesy4,
  title={{$\mathrm{Re}^{2}$ Agent}: Reflection and Re-execution Agent for Embodied Decision Making},
  author={Chen, Yang and You, Hong-Jie and Shao, Jie-Jing and Yang, Xiao-Wen and Yang, Ming and Li, Yu-Feng and Guo, Lan-Zhe},
  booktitle={NeurIPS 2025 Challenge on Foundation Models for Embodied Agents},
  year={2025}
}

@inproceedings{pathplan1,
  title={Evaluating vision-language models as evaluators in path planning},
  author={Aghzal, Mohamed and Yue, Xiang and Plaku, Erion and Yao, Ziyu},
  booktitle={2025 IEEE/CVF Conference on Computer Vision and Pattern Recognition (CVPR)},
  pages={6886--6897},
  year={2025},
  organization={IEEE}
}

@inproceedings{spatial3,
  title={Theory of Space: Can Foundation Models Construct Spatial Beliefs through Active Exploration?},
  author={Zhang, Pingyue and Huang, Zihan and Wang, Yue and Zhang, Jieyu and Xue, Letian and Wang, Zihan and Wang, Qineng and Chandrasegaran, Keshigeyan and Zhang, Ruohan and Choi, Yejin and others},
  booktitle={The Fourteenth International Conference on Learning Representations},
  year={2026}
}

@article{survey,
  title={A survey on pickup and delivery problems: Part II: Transportation between pickup and delivery locations},
  author={Parragh, Sophie N and Doerner, Karl F and Hartl, Richard F},
  journal={Journal f{\"u}r Betriebswirtschaft},
  volume={58},
  number={2},
  pages={81--117},
  year={2008},
  publisher={Springer}
}

@article{survey2,
  title={A review of dynamic vehicle routing problems},
  author={Pillac, Victor and Gendreau, Michel and Gu{\'e}ret, Christelle and Medaglia, Andr{\'e}s L},
  journal={European Journal of Operational Research},
  volume={225},
  number={1},
  pages={1--11},
  year={2013},
  publisher={Elsevier}
}

@article{pathplan2,
  title={Vsp: Assessing the dual challenges of perception and reasoning in spatial planning tasks for vlms},
  author={Wu, Qiucheng and Zhao, Handong and Saxon, Michael and Bui, Trung and Wang, William Yang and Zhang, Yang and Chang, Shiyu},
  journal={arXiv preprint arXiv:2407.01863},
  year={2024}
}
\bibliographystyle{plainnat}

\clearpage
\appendix
\phantomsection
\section*{\Large Table of Contents in Appendix}
\label{app:contents}
\addcontentsline{toc}{section}{Table of Contents in Appendix}
\vspace{0.5cm}

{
\large

\newcommand{\appTOCsection}[1]{%
    \hyperref[#1]{%
        \textbf{\ref*{#1}\quad\nameref*{#1}}%
    }\dotfill & \pageref*{#1}\\
}

\newcommand{\appTOCsubsection}[1]{%
    \hspace{0.5cm}%
    \hyperref[#1]{%
        \ref*{#1}\quad\nameref*{#1}%
    }\dotfill & \pageref*{#1}\\
}

\begin{center}
\begin{tabular}{@{}p{0.86\textwidth}@{\hspace{0.1cm}}r@{}}

    \appTOCsection{app:enviro}
    \appTOCsubsection{app:textworld}
    \appTOCsubsection{app:simworld}
    \\[0.2cm]

    \appTOCsection{app:simworld_exp}
    \appTOCsubsection{app:textworld_task15}
    \appTOCsubsection{app:simworld_full}
    \appTOCsubsection{app:simworld_aggregate}
    \appTOCsubsection{app:failure_attribution}
    \appTOCsubsection{app:qa_examples}
    \appTOCsubsection{app:question_families}
    \appTOCsubsection{app:vqa_cases}
    \\[0.2cm]

    \appTOCsection{app:solvers}
    \appTOCsubsection{bpp:dfs}
    \appTOCsubsection{bpp:onlinedp}
    \appTOCsubsection{bpp:offlinedp}
    \\[0.2cm]

    \appTOCsection{app:prompt}
    \appTOCsubsection{dpp:protocol}
    \appTOCsubsection{dpp:text}
    \appTOCsubsection{dpp:sim}
    \appTOCsubsection{dpp:scene_specific}

\end{tabular}
\end{center}
\vspace{1cm}
}

\FloatBarrier
\newpage

\section{Benchmark Details}
\label{app:enviro}
DualWorldBench contains two complementary execution environments. \textit{TextWorld} isolates symbolic and temporal decision making under deterministic dynamics, whereas \textit{SimWorld} evaluates grounded planning under map-dependent navigation costs. Both environments share the same pickup--delivery abstraction: each task has a pickup node, a delivery node, a reward, and hard precedence and capacity constraints.

\subsection{TextWorld: Symbolic Event-Driven Planning}
\label{app:textworld}
TextWorld uses a continuous-time event simulator. The agent state consists of current time, two-dimensional position, currently carried tasks, completed tasks, and the active task pool. Unless otherwise specified, the agent starts from $(0,0)$ with speed $v=8.33\,\mathrm{m/s}$ and capacity $K=5$. Motion cost is Euclidean distance divided by speed, and pickup or delivery is accepted only when the agent is within a $5.0\,\mathrm{m}$ tolerance of the target coordinate.

The benchmark varies two factors. First, task composition is either homogeneous (\textbf{A}) or mixed (\textbf{M}). In the homogeneous setting, all tasks are Type-A. In the mixed setting, tasks are sampled from Type-A/B/C with an effective $7{:}2{:}1$ distribution. Their base rewards are sampled from $[10,30]$, $[30,60]$, and $[60,100]$, respectively. Type-C tasks are marked as priority tasks. Second, task release is either static (\textbf{L1}) or online (\textbf{L2}). In L1, all trigger times are zero and deadlines are non-binding. In L2, tasks are released over time and deadlines are determined from the pickup--delivery distance and a type-dependent time factor: $5.0$ for Type-A, $3.0$ for Type-B, and $2.0$ for Type-C. Delivery revenue is
\begin{equation}
r_i(t)=\max\{0, R_i-\lambda_{\tau_i}\max(0,t-e_i)\},
\end{equation}
with penalty rates $\lambda_A=0.1$, $\lambda_B=0.2$, and $\lambda_C=0.3$.

When a new task is released while the agent is moving, execution is interrupted at the interpolated physical position, the current plan is discarded, and the planner is invoked again from the updated state. This mechanism makes L2 a genuine online replanning setting rather than a static scheduling task with delayed visibility.

\subsection{SimWorld: Grounded Topological Planning}
\label{app:simworld}
SimWorld uses the same pickup--delivery interface but replaces Euclidean symbolic motion with scene-dependent navigation costs. The benchmark contains 13 scenes organized into five reported groups: \textit{typical} as the open-layout group; five T1 procedural indoor scenes (\textit{spine rooms}, \textit{cross suite}, \textit{offset wing}, \textit{dual room}, and \textit{corner hall}); five T2 procedural scenes (\textit{loop rooms}, \textit{dual spine}, \textit{ring hub}, \textit{offset blocks}, and \textit{cluster corridor}); and the \textit{indoor} and \textit{campus} transfer groups. Each scene provides task instances with $N\in\{5,8,10\}$ pickup--delivery pairs. The procedural scenes contain five instances per task count, while the transfer scenes use manually curated or simulator-derived instances.

All SimWorld tasks are available at $t=0$ and use a uniform pickup--delivery objective, so the intended difficulty is grounded spatial planning rather than temporal release. Each task instance is evaluated under three input modalities: text-only task coordinates, image-only BEV map with labeled nodes, and multimodal input that combines the BEV map with the structured task table. All modalities output node sequences, which the evaluator maps to task coordinates and expands into executable actions.

Evaluation uses a precomputed navigation-distance matrix for each scene. Each mapped task coordinate is matched to the nearest valid navigation anchor within a bounded tolerance; traveled distance is computed from graph distance plus anchor-access offsets. The evaluator records feasibility, completion rate, success rate, total distance, total reward, travel time, and constraint violations. Constraint violations include invalid graph targets, pickup or delivery without arrival, duplicate actions, delivery before pickup, and capacity overflow.

\begin{figure}[htbp]
    \centering
    \begin{subfigure}[b]{0.32\textwidth}
        \centering
        \includegraphics[width=\textwidth]{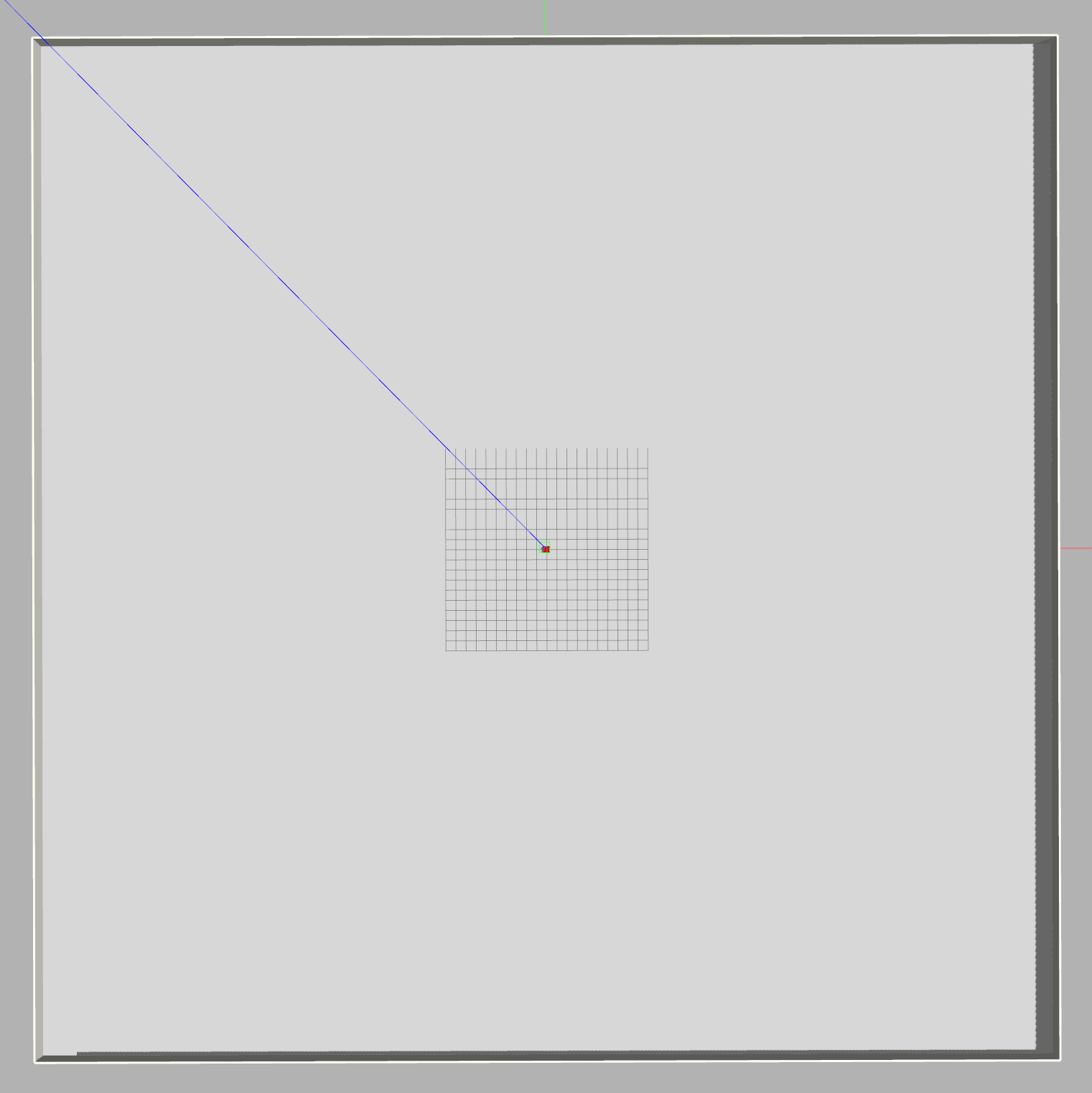}
        \caption{Open-layout baseline.}
        \label{fig:app_open}
    \end{subfigure}
    \hfill
    \begin{subfigure}[b]{0.32\textwidth}
        \centering
        \includegraphics[width=\textwidth]{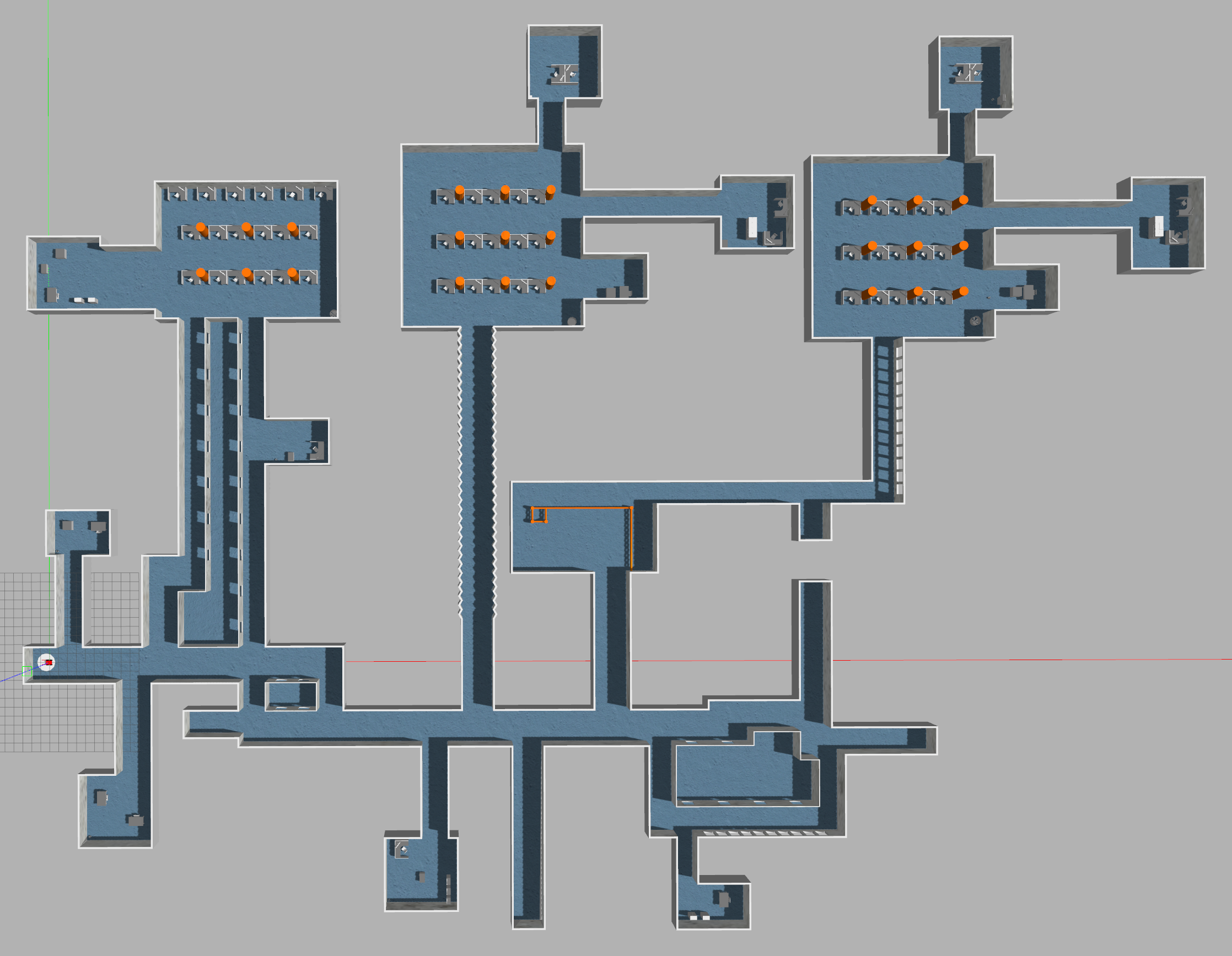}
        \caption{Indoor transfer scene.}
        \label{fig:app_indoor}
    \end{subfigure}
    \hfill
    \begin{subfigure}[b]{0.32\textwidth}
        \centering
        \includegraphics[width=\textwidth]{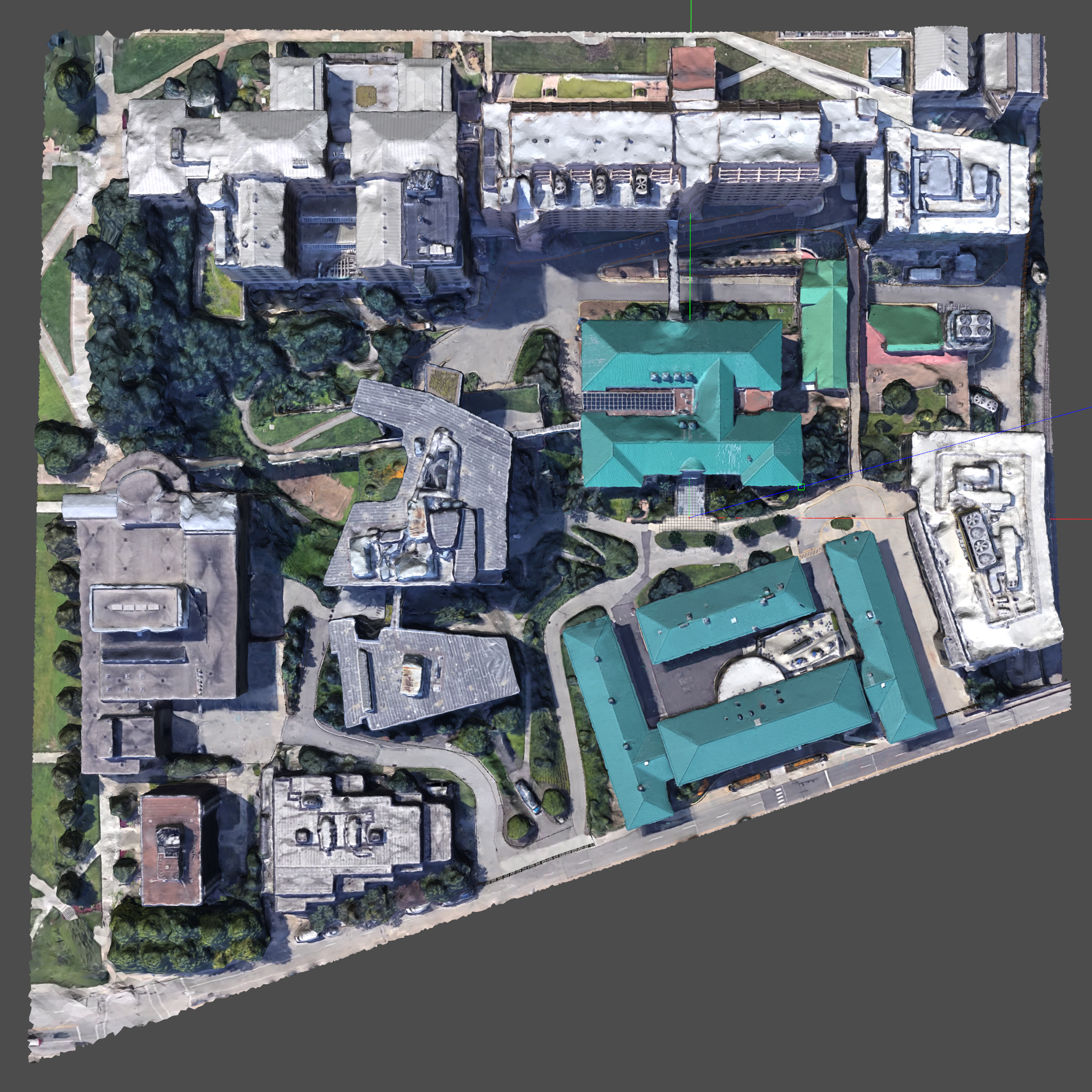}
        \caption{Campus transfer scene.}
        \label{fig:app_campus}
    \end{subfigure}
    \caption{Representative Gazebo BEV previews used in SimWorld.}
    \label{fig:app_simworld_representative}
\end{figure}

\begin{figure*}[htbp]
    \centering
    \begin{subfigure}[b]{0.19\textwidth}
        \centering
        \includegraphics[width=\textwidth]{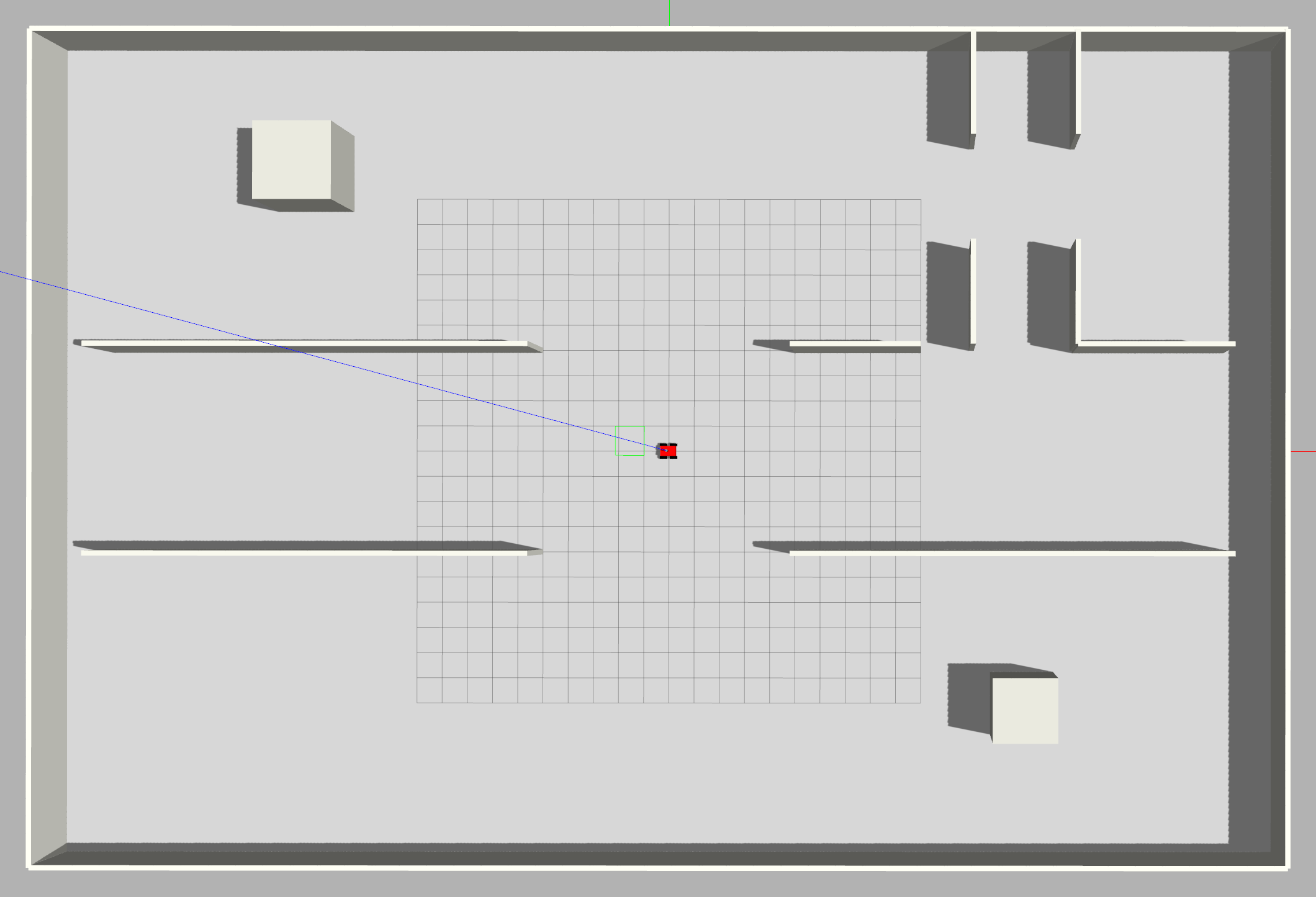}
        \caption{Spine rooms.}
    \end{subfigure}
    \hfill
    \begin{subfigure}[b]{0.19\textwidth}
        \centering
        \includegraphics[width=\textwidth]{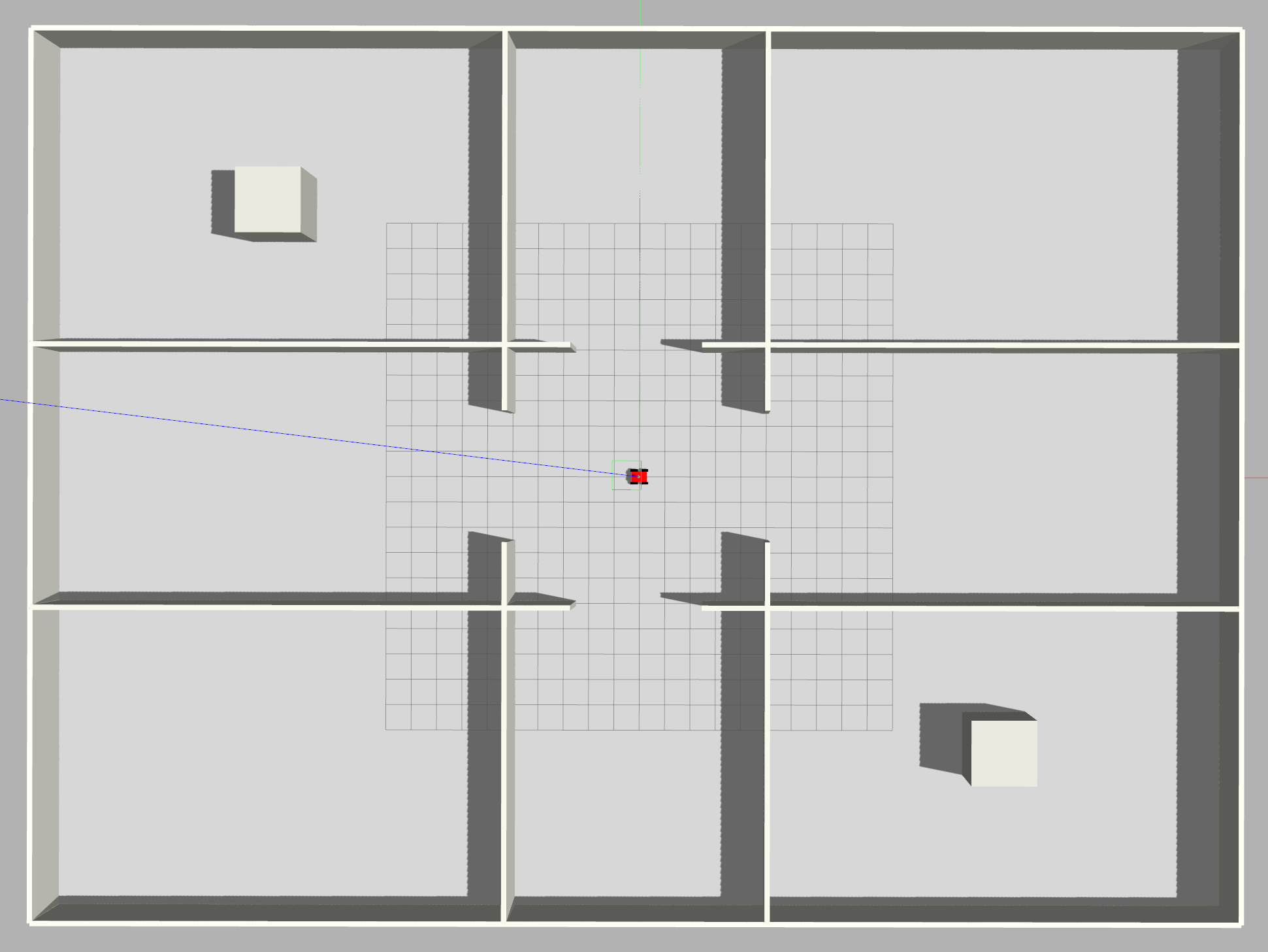}
        \caption{Cross suite.}
    \end{subfigure}
    \hfill
    \begin{subfigure}[b]{0.19\textwidth}
        \centering
        \includegraphics[width=\textwidth]{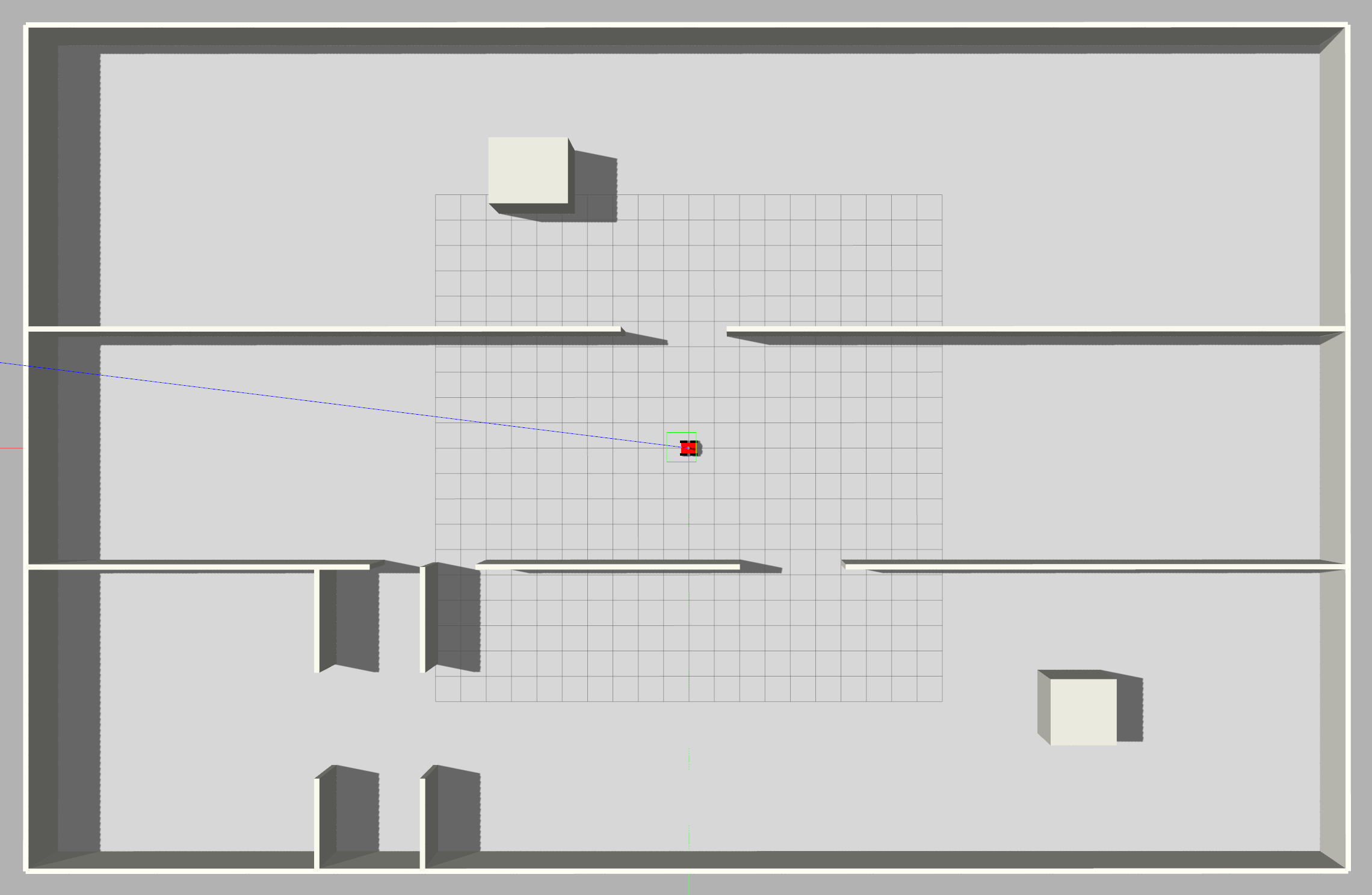}
        \caption{Offset wing.}
    \end{subfigure}
    \hfill
    \begin{subfigure}[b]{0.19\textwidth}
        \centering
        \includegraphics[width=\textwidth]{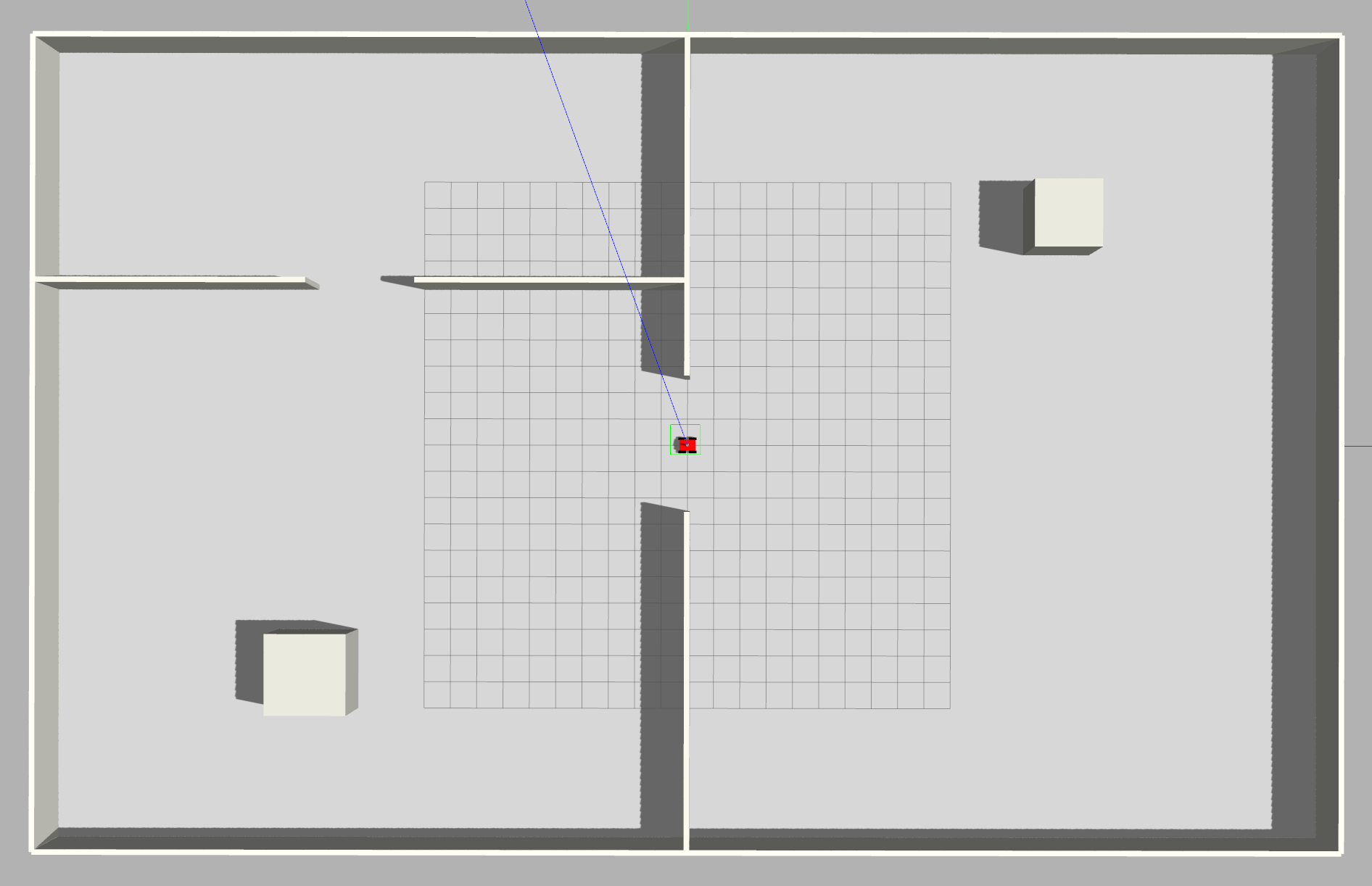}
        \caption{Dual room.}
    \end{subfigure}
    \hfill
    \begin{subfigure}[b]{0.19\textwidth}
        \centering
        \includegraphics[width=\textwidth]{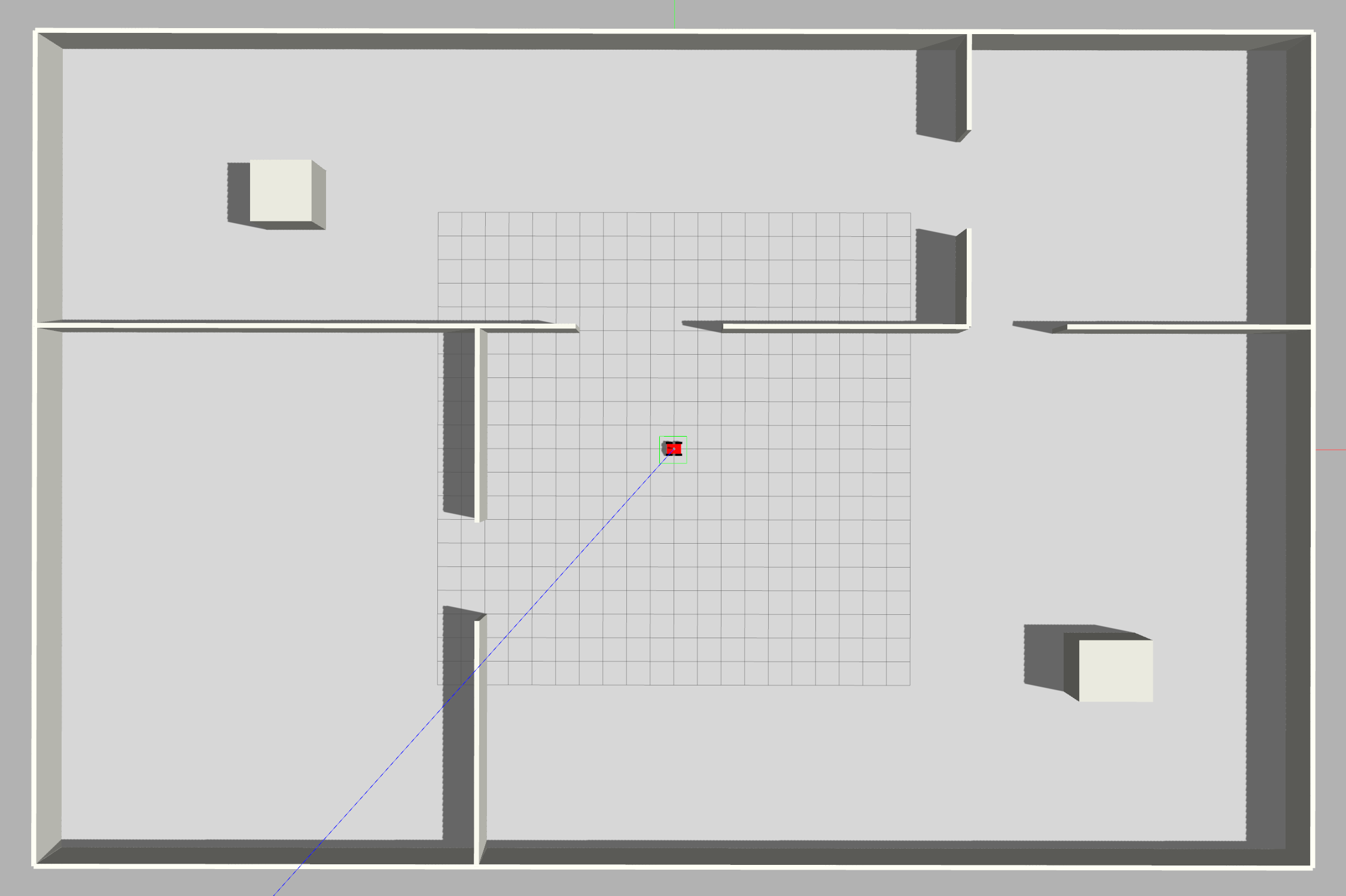}
        \caption{Corner hall.}
    \end{subfigure}
    \caption{T1 procedural SimWorld layouts. These scenes contain structured corridor systems where local visual proximity often remains a useful but imperfect heuristic.}
    \label{fig:app_t1_layouts}
\end{figure*}

\begin{figure*}[htbp]
    \centering
    \begin{subfigure}[b]{0.19\textwidth}
        \centering
        \includegraphics[width=\textwidth]{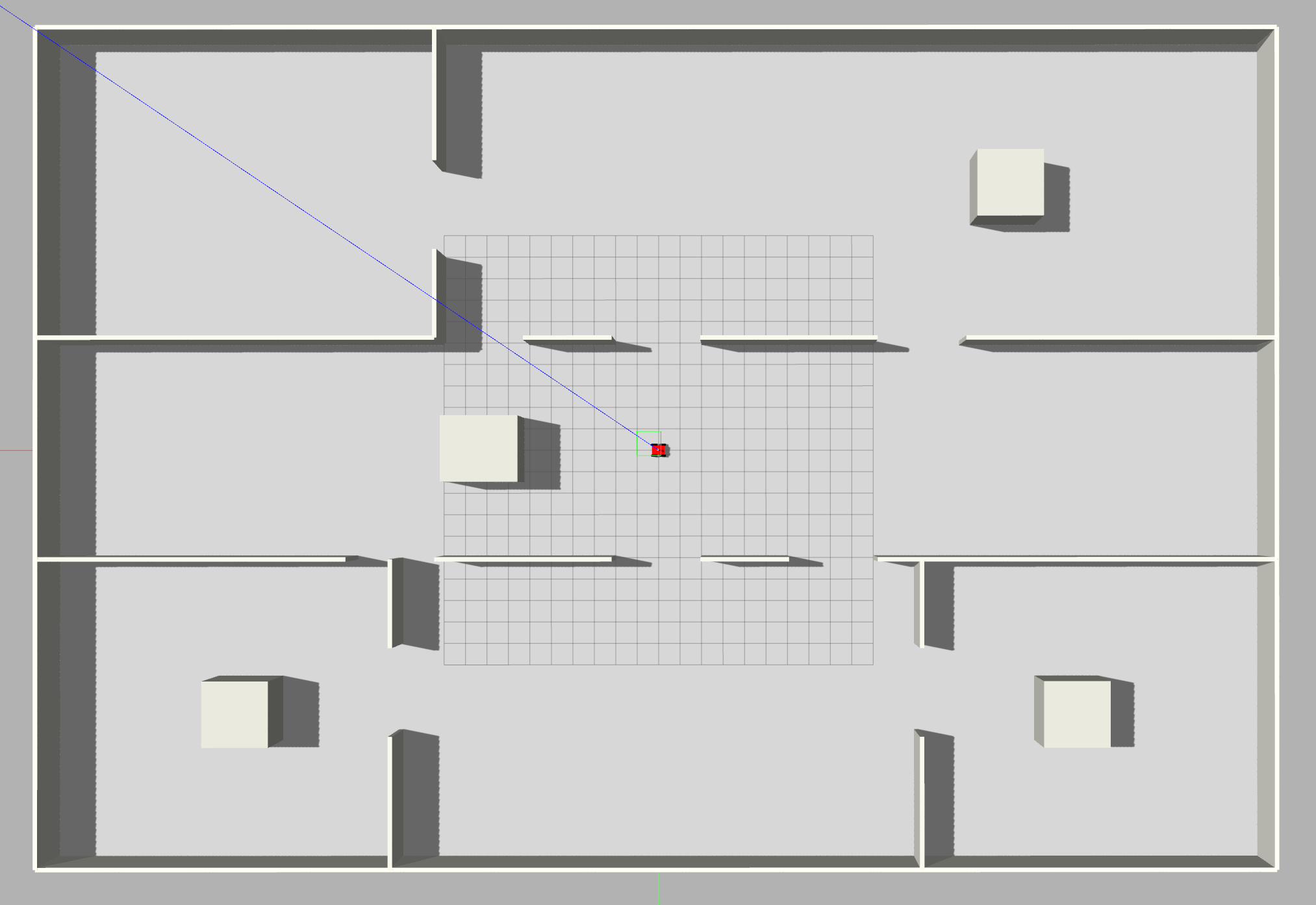}
        \caption{Loop rooms.}
    \end{subfigure}
    \hfill
    \begin{subfigure}[b]{0.19\textwidth}
        \centering
        \includegraphics[width=\textwidth]{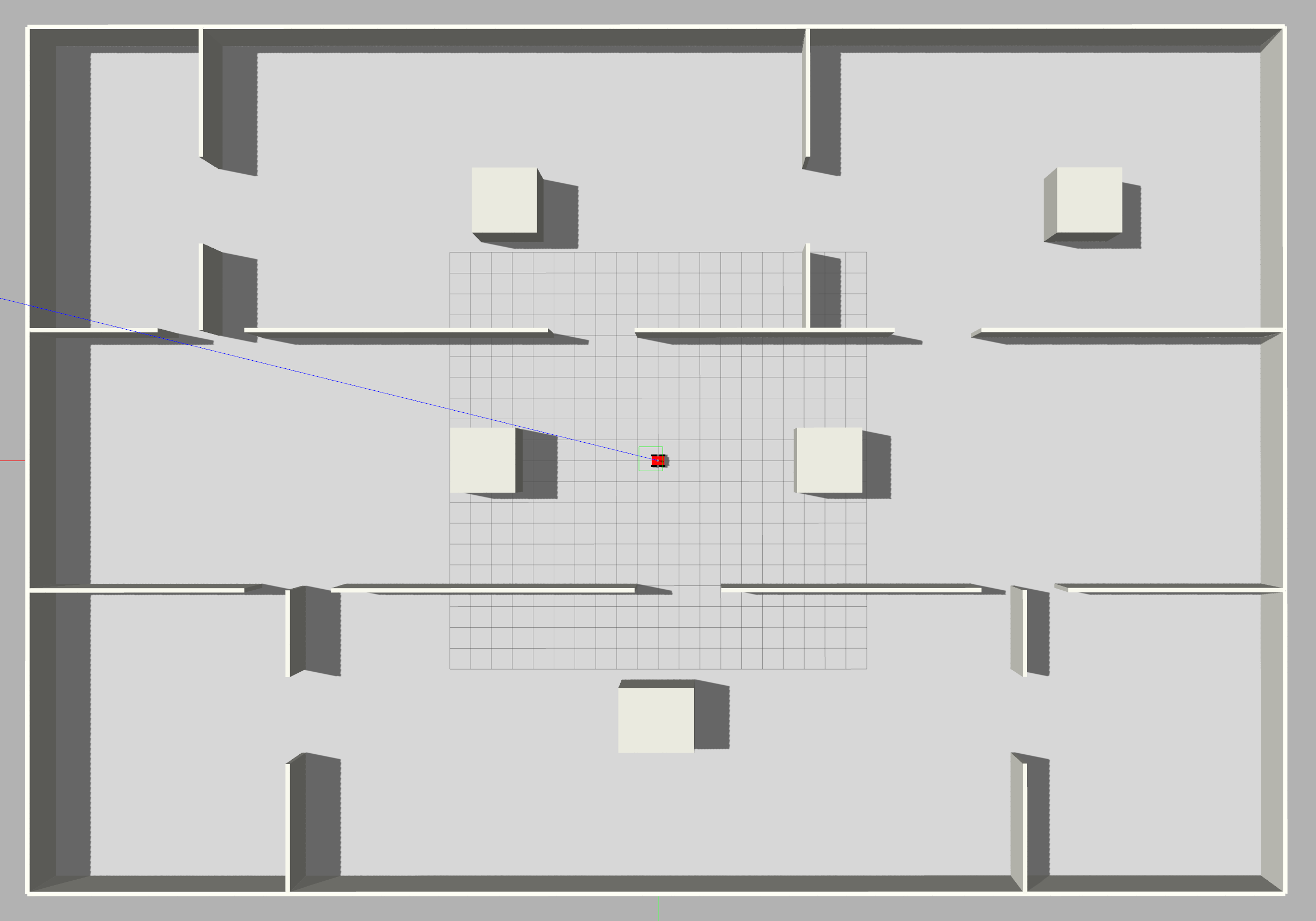}
        \caption{Dual spine.}
    \end{subfigure}
    \hfill
    \begin{subfigure}[b]{0.19\textwidth}
        \centering
        \includegraphics[width=\textwidth]{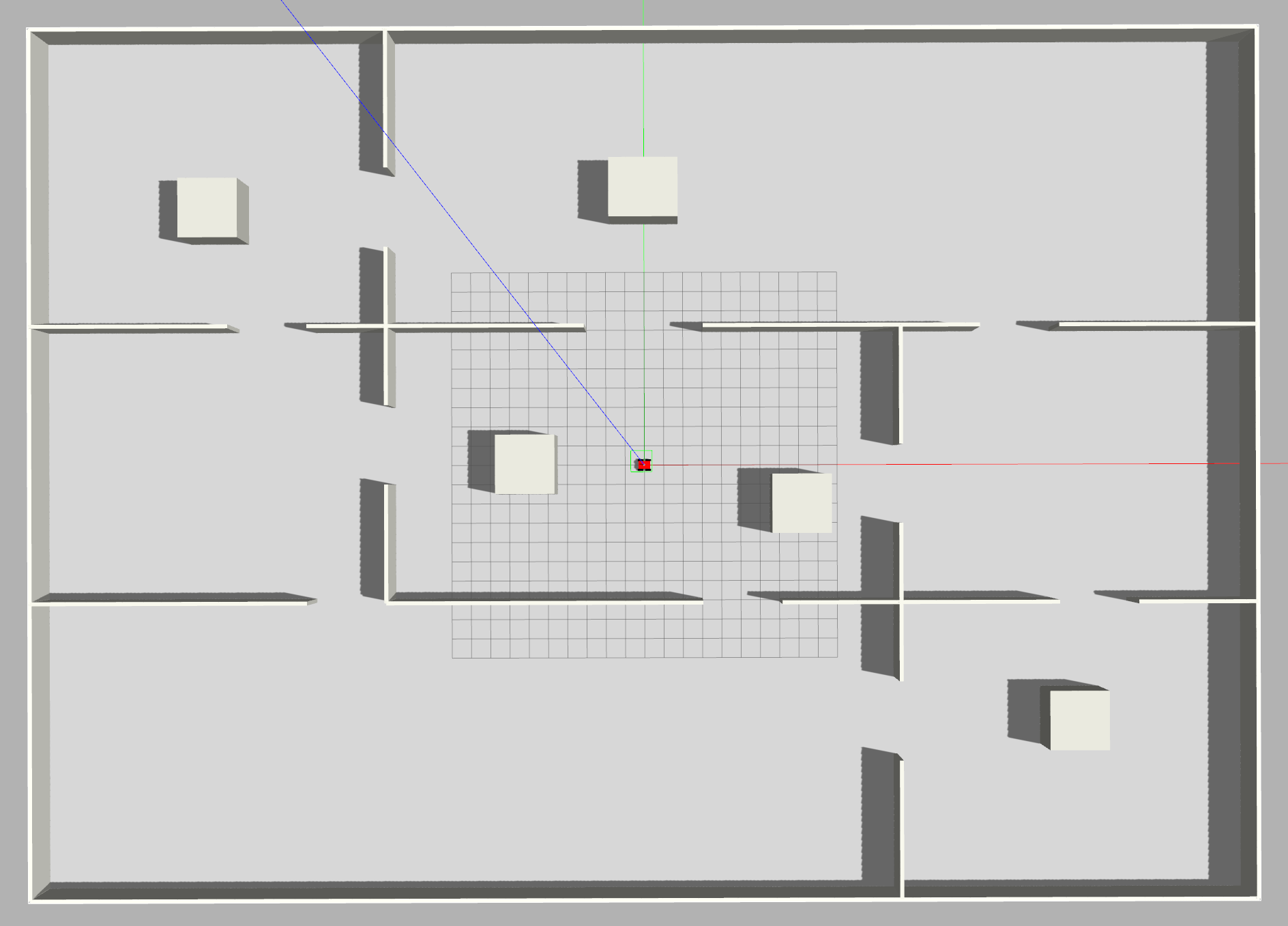}
        \caption{Ring hub.}
    \end{subfigure}
    \hfill
    \begin{subfigure}[b]{0.19\textwidth}
        \centering
        \includegraphics[width=\textwidth]{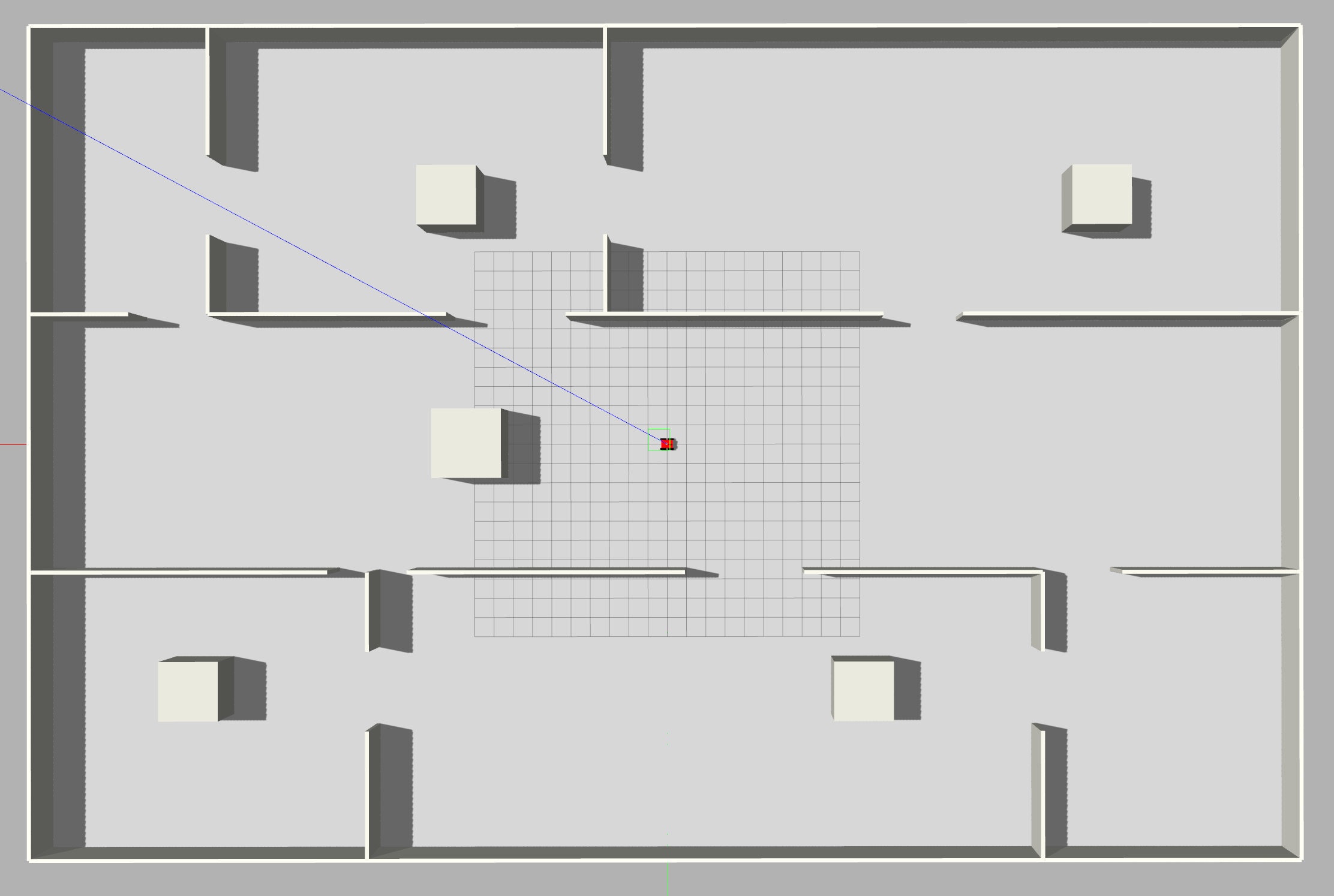}
        \caption{Offset blocks.}
    \end{subfigure}
    \hfill
    \begin{subfigure}[b]{0.19\textwidth}
        \centering
        \includegraphics[width=\textwidth]{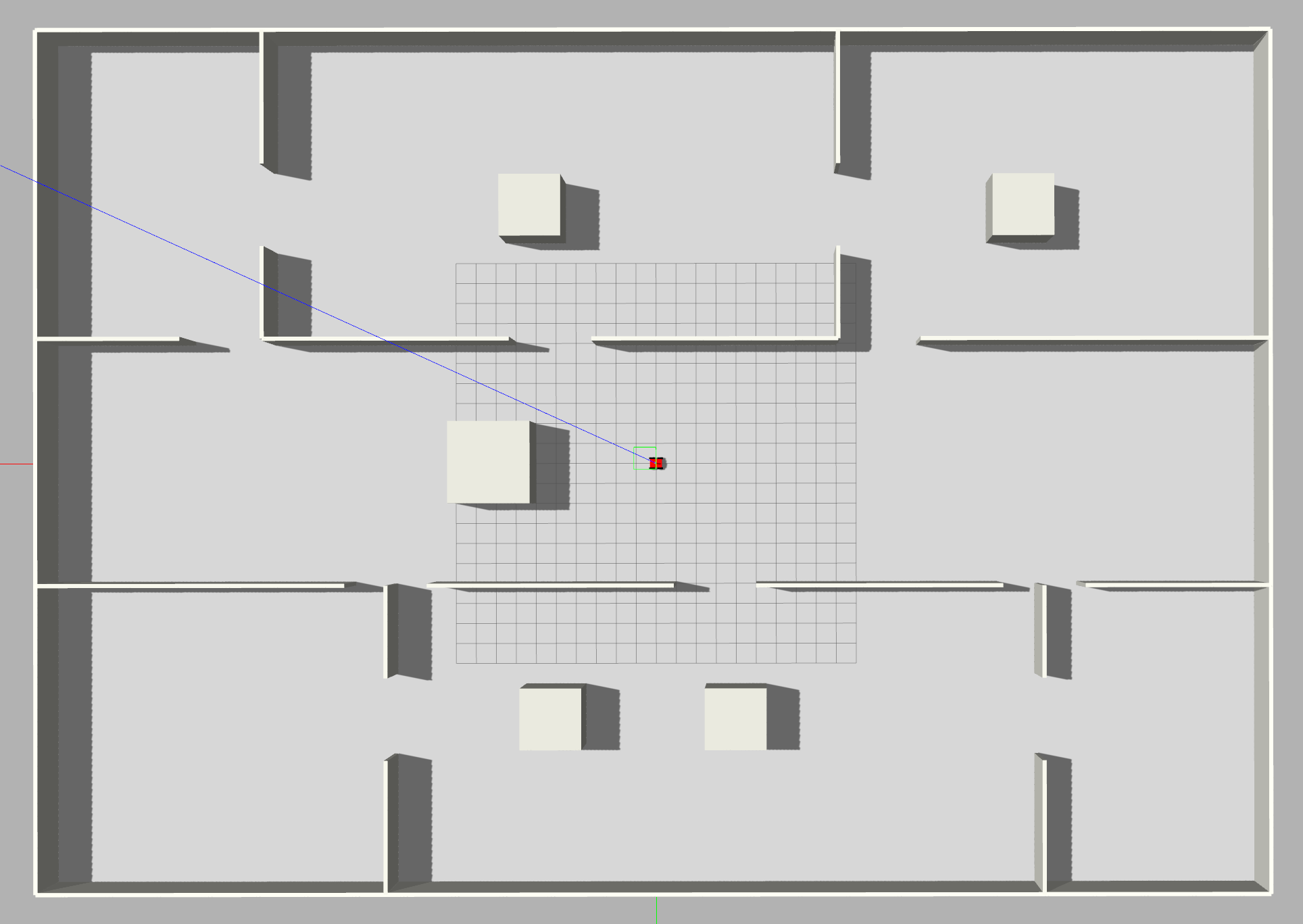}
        \caption{Cluster corridor.}
    \end{subfigure}
    \caption{T2 procedural SimWorld layouts. These scenes increase topological ambiguity through loops, disconnected-looking regions, and longer detours, making straight-line proximity less reliable.}
    \label{fig:app_t2_layouts}
\end{figure*}

Figures~\ref{fig:app_simworld_representative}--\ref{fig:app_t2_layouts} show BEV previews of the Gazebo scenes used to instantiate SimWorld. These previews are provided to illustrate the simulated physical layouts; task-node annotations are introduced later in the VQA examples.

\FloatBarrier

\section{Supplementary Experiments}
\label{app:simworld_exp}
This section supplements the main results with the complete TextWorld task-15 breakdown, violation-aware SimWorld aggregates, TextWorld reward-gap attribution, and a question-family decomposition of DualWorld-VQA. For SimWorld, we report \textit{success rate} as the fraction of samples for which all tasks are completed without violations, \textit{completion rate} as the average fraction of completed tasks, and \textit{violations} as the average number of executable-interface violations per sample.

\subsection{Full TextWorld Results at \texorpdfstring{$n=15$}{n=15}}
\label{app:textworld_task15}
Table~\ref{tab:app_textworld_task15_full} restores the model-wise results underlying the grouped comparison in Fig~\ref{fig:radar_textworld}. A priority violation is applied before any averaging: a model--setting cell that delivers an A/B task while carrying a Type-C task is scored as $R{=}0$, $T{=}9999$.


\begin{table*}[htbp]
\centering
\small
\setlength{\tabcolsep}{3.2pt}
\renewcommand{\arraystretch}{1.08}
\caption{Complete TextWorld results at task count $15$.
Average is the unweighted mean of A-L1, M-L1, A-L2, and M-L2 at seed $1$.
Priority-violating cells are scored as $R{=}0$, $T{=}9999$.
Ranks are computed separately within the $14$ LLMs and the three solver baselines.
Bold identifies reward ceilings and the Offline-DP reference values.
Gray shading highlights the Offline-DP reference.}
\label{tab:app_textworld_task15_full}

\begin{adjustbox}{max width=\textwidth}
\begin{tabular}{lccccccccccc}
\toprule
\multirow{2}{*}{Method}
& \multirow{2}{*}{Rank}
& \multicolumn{2}{c}{\cellcolor[HTML]{E0F2F1} Average}
& \multicolumn{2}{c}{\cellcolor[HTML]{F5E6E9} A-L1}
& \multicolumn{2}{c}{\cellcolor[HTML]{FFF0E6} M-L1}
& \multicolumn{2}{c}{\cellcolor[HTML]{E2F0D9} A-L2}
& \multicolumn{2}{c}{\cellcolor[HTML]{F4EAD5} M-L2} \\
\cmidrule(lr){3-4}
\cmidrule(lr){5-6}
\cmidrule(lr){7-8}
\cmidrule(lr){9-10}
\cmidrule(lr){11-12}
& & $R\uparrow$ & $T\downarrow$
& $R\uparrow$ & $T\downarrow$
& $R\uparrow$ & $T\downarrow$
& $R\uparrow$ & $T\downarrow$
& $R\uparrow$ & $T\downarrow$ \\
\midrule

\multicolumn{12}{c}{
    \cellcolor[HTML]{E6F2FF}\textit{Open-source LLMs}
} \\
Qwen3.6-27B
& 8 & 382.49 & 7250.02
& \textbf{302.97} & 5291.29
& \textbf{558.94} & 5612.23
& 251.02 & 10166.85
& 417.01 & 7929.71 \\
Llama-2-13B
& 12 & 251.15 & 8391.05
& \textbf{302.97} & 6165.02
& \textbf{558.94} & 6820.17
& 142.68 & 10580.02
& 0.00 & 9999.00 \\
Qwen3-VL-8B
& 13 & 31.79 & 3040.67
& 10.04 & 453.02
& 60.08 & 455.02
& 10.04 & 4102.59
& 47.00 & 7152.04 \\
Llama-3-8B
& 14 & 0.00 & 4101.73
& 0.00 & 4.00
& 0.00 & 4.00
& 0.00 & 9246.89
& 0.00 & 7152.04 \\
\midrule

\multicolumn{12}{c}{
    \cellcolor[HTML]{E6F2FF}\textit{Proprietary LLMs}
} \\
Gemini-3-Pro
& 1 & 392.14 & 6242.17
& \textbf{302.97} & 3614.01
& \textbf{558.94} & 3806.19
& \textbf{289.72} & 9619.76
& 416.92 & 7928.71 \\
Claude-Sonnet-5
& 2 & 392.05 & 6799.79
& \textbf{302.97} & 5500.38
& \textbf{558.94} & 4149.32
& 289.36 & 9619.76
& 416.92 & 7929.71 \\
Claude-Opus-5
& 3 & 392.01 & 6227.59
& \textbf{302.97} & 4366.61
& \textbf{558.94} & 3107.06
& 289.21 & 9619.76
& 416.92 & 7816.91 \\
Qwen3.7-Plus
& 4 & 392.01 & 6494.65
& \textbf{302.97} & 3684.40
& \textbf{558.94} & 4745.75
& 289.21 & 9619.76
& 416.92 & 7928.71 \\
Gemini-2.5-Flash
& 5 & 392.01 & 6626.44
& \textbf{302.97} & 4307.42
& \textbf{558.94} & 4649.88
& 289.21 & 9619.76
& 416.92 & 7928.71 \\
DeepSeek-v4-Flash
& 6 & 392.01 & 6872.71
& \textbf{302.97} & 4973.04
& \textbf{558.94} & 5256.33
& 289.21 & 9619.76
& 416.92 & 7641.72 \\
GPT-5.6-Sol
& 7 & 392.01 & 7222.23
& \textbf{302.97} & 5093.83
& \textbf{558.94} & 6359.43
& 289.21 & 9619.76
& 416.92 & 7815.91 \\
Claude-4.5-Haiku
& 9 & 378.70 & 6891.58
& \textbf{302.97} & 4877.98
& \textbf{558.94} & 5535.86
& 262.07 & 9619.76
& 390.80 & 7532.73 \\
GPT-4o
& 10 & 373.82 & 7724.05
& \textbf{302.97} & 6020.76
& \textbf{558.94} & 6346.16
& 241.98 & 10273.35
& 391.38 & 8255.92 \\
GPT-5.4-Mini
& 11 & 353.88 & 7696.76
& \textbf{302.97} & 5615.12
& \textbf{558.94} & 6324.61
& 255.85 & 10685.64
& 297.76 & 8161.68 \\
\midrule

\multicolumn{12}{c}{
    \cellcolor[HTML]{E6F2FF}
    \textit{Combinatorial optimization solvers}
} \\
DFS
& 3 & 392.01 & 6602.05
& \textbf{302.97} & 4187.31
& \textbf{558.94} & 4903.47
& 289.21 & 9619.76
& 416.92 & 7697.67 \\
Online DP
& 2 & 392.14 & 5628.85
& \textbf{302.97} & 2573.66
& \textbf{558.94} & 2680.27
& \textbf{289.72} & 9619.76
& 416.92 & 7641.72 \\
\rowcolor[HTML]{F0F0F0}
Offline DP
& 1 & \textbf{409.71} & \textbf{5554.29}
& \textbf{302.97} & \textbf{2573.66}
& \textbf{558.94} & \textbf{2680.27}
& \textbf{289.72} & \textbf{9529.04}
& \textbf{487.22} & \textbf{7434.19} \\
\bottomrule
\end{tabular}
\end{adjustbox}
\end{table*}

\subsection{Full SimWorld Results across five scene groups}
\label{app:simworld_full}

Table~\ref{tab:simworld_scene_groups} reports $F$, $C$, and $NC$ for all six models under multimodal, image-only, and text-only inputs across Open, T1, T2, Indoor, and Campus.
Qwen3.7-Plus achieves $F=C=1$ across all scenes with multimodal input, while GPT-5.6-Sol attains the lowest $NC$ in every scene--modality combination.
Image-only performance varies substantially across models, particularly in $F$, highlighting differences in planning robustness across input modalities.

\begin{table*}[t!]
    \centering
    \caption{SimWorld scene-wise results across five scene groups: open, T1, T2, indoor, and campus.}
    \label{tab:simworld_scene_groups}
    \resizebox{\textwidth}{!}{
    \begin{tabular}{c|c|ccc|ccc|ccc|ccc|ccc}
    \toprule
    \multirow{2}{*}{Method} & \multirow{2}{*}{Modality} & \multicolumn{3}{c|}{\cellcolor[HTML]{F5E6E9} Open} & \multicolumn{3}{c|}{\cellcolor[HTML]{E8F1FF} T1} & \multicolumn{3}{c|}{\cellcolor[HTML]{FFF0E6} T2} & \multicolumn{3}{c|}{\cellcolor[HTML]{E2F0D9} Indoor} & \multicolumn{3}{c}{\cellcolor[HTML]{F4EAD5} Campus} \\
     &  & F$\uparrow$ & C$\uparrow$ & NC$\downarrow$ & F$\uparrow$ & C$\uparrow$ & NC$\downarrow$ & F$\uparrow$ & C$\uparrow$ & NC$\downarrow$ & F$\uparrow$ & C$\uparrow$ & NC$\downarrow$ & F$\uparrow$ & C$\uparrow$ & NC$\downarrow$ \\
    \midrule
    \multirow{3}{*}{Qwen3-VL-8B} & Multi. & 1.000 & 0.783 & 1.66 & 0.973 & 0.979 & 1.91 & 1.000 & 0.984 & 1.91 & 1.000 & 1.000 & 1.73 & 1.000 & 0.987 & 2.16 \\
     & Image & 0.889 & 0.950 & 1.92 & 0.880 & 0.915 & 1.91 & 0.773 & 0.885 & 1.91 & 0.733 & 0.905 & 1.76 & 1.000 & 0.920 & 1.97 \\
     & Text & 1.000 & 0.853 & 1.83 & 1.000 & 0.462 & 1.37 & 1.000 & 0.533 & 1.43 & 1.000 & 0.808 & 1.64 & 1.000 & 0.557 & 1.60 \\
    \midrule
    \multirow{3}{*}{Qwen3.7-Plus} & Multi. & 1.000 & 1.000 & 1.24 & \best{1.000} & \best{1.000} & 1.23 & \best{1.000} & \best{1.000} & 1.30 & 1.000 & 1.000 & 1.29 & 1.000 & 1.000 & 1.29 \\
     & Image & 0.667 & 0.894 & 1.20 & 1.000 & 1.000 & 1.57 & 1.000 & 1.000 & 1.57 & 1.000 & 1.000 & 1.35 & 1.000 & 1.000 & 1.60 \\
     & Text & 1.000 & 1.000 & 1.20 & 1.000 & 0.968 & 1.28 & 1.000 & 0.977 & 1.23 & 1.000 & 0.920 & 1.23 & 1.000 & 1.000 & 1.28 \\
    \midrule
    \multirow{3}{*}{Claude-Haiku-4.5} & Multi. & 0.333 & 0.908 & 2.68 & 0.640 & 0.947 & 2.36 & 0.880 & 0.971 & 1.97 & 0.867 & 0.977 & 2.06 & 0.933 & 0.993 & 1.92 \\
     & Image & 0.000 & 0.653 & 2.12 & 0.133 & 0.679 & 2.85 & 0.027 & 0.693 & 4.10 & 0.533 & 0.760 & 2.05 & 0.867 & 0.900 & 2.08 \\
     & Text & 0.889 & 0.931 & 2.04 & 0.907 & 0.985 & 1.72 & 0.933 & 0.984 & 2.03 & 0.867 & 0.977 & 1.92 & 0.800 & 0.983 & 2.41 \\
    \midrule
    \multirow{3}{*}{Claude-Opus-5} & Multi. & 0.778 & 0.956 & 1.20 & 0.520 & 0.926 & 1.49 & 0.440 & 0.966 & 2.18 & 0.600 & 0.930 & 1.47 & 0.667 & 0.950 & 1.34 \\
     & Image & 0.111 & 0.678 & 1.33 & 0.560 & 0.878 & 1.45 & 0.640 & 0.893 & 1.44 & 0.467 & 0.785 & 1.57 & 0.533 & 0.857 & 1.46 \\
     & Text & 0.667 & 0.944 & 1.30 & 0.667 & 0.972 & 1.39 & 0.707 & 0.964 & 1.32 & 0.800 & 0.993 & 1.37 & 0.733 & 0.993 & 1.64 \\
    \midrule
    \multirow{3}{*}{GPT-5.4-mini} & Multi. & 0.556 & 0.897 & 1.59 & 0.573 & 0.917 & 1.71 & 0.680 & 0.945 & 1.94 & 0.533 & 0.925 & 1.71 & 0.933 & 0.993 & 1.90 \\
     & Image & 0.111 & 0.747 & 1.73 & 0.093 & 0.684 & 2.10 & 0.120 & 0.699 & 2.03 & 0.200 & 0.687 & 1.96 & 0.267 & 0.705 & 1.93 \\
     & Text & 1.000 & 0.975 & 1.77 & 1.000 & 0.969 & 1.79 & 1.000 & 0.964 & 1.78 & 0.933 & 0.953 & 1.66 & 1.000 & 0.967 & 1.89 \\
    \midrule
    \multirow{3}{*}{GPT-5.6-Sol} & Multi. & \best{1.000} & \best{1.000} & \best{1.08} & 0.973 & 0.996 & \best{1.12} & 0.987 & 0.999 & 1.13 & \best{1.000} & \best{1.000} & \best{1.19} & \best{1.000} & \best{1.000} & \best{1.19} \\
     & Image & 0.778 & 0.956 & 1.11 & 1.000 & 1.000 & 1.36 & 1.000 & 1.000 & 1.43 & 1.000 & 0.987 & 1.36 & 1.000 & 0.993 & 1.54 \\
     & Text & 1.000 & 1.000 & 1.09 & 0.933 & 0.988 & 1.13 & 0.987 & 0.998 & \best{1.11} & 1.000 & 1.000 & 1.20 & 1.000 & 1.000 & 1.21 \\
    \bottomrule
    \end{tabular}}
\end{table*}

\subsection{SimWorld Aggregate Diagnostics}
\label{app:simworld_aggregate}
Table~\ref{tab:app_simworld_overall} reports aggregate SimWorld performance across all evaluated scene groups and task counts for the Claude, GPT, and Qwen model families. This view emphasizes executable consistency: a plan can complete many tasks while still failing the strict success criterion if it contains invalid actions, wrong node references, capacity errors, or pickup--delivery precedence violations.

\begin{table*}[htbp]
\centering
\small
\caption{Aggregate SimWorld performance by model and modality. SR: success rate; CR: average completion rate; Viol.: average violations per sample.}
\label{tab:app_simworld_overall}
\begin{tabular}{llrrrr}
\toprule
Model & Modality & Samples & SR $\uparrow$ & CR $\uparrow$ & Viol. $\downarrow$ \\
\midrule
Claude Haiku 4.5 & Image & 189 & 0.122 & 0.707 & 25.090 \\
Claude Haiku 4.5 & Multimodal & 189 & 0.751 & 0.961 & 6.492 \\
Claude Haiku 4.5 & Text & 189 & 0.831 & 0.981 & 3.476 \\
Claude Opus 5 & Image & 189 & 0.561 & 0.865 & 3.889 \\
Claude Opus 5 & Multimodal & 189 & 0.513 & 0.945 & 8.788 \\
Claude Opus 5 & Text & 189 & 0.698 & 0.971 & 3.344 \\
GPT-5.4 Mini & Image & 189 & 0.116 & 0.695 & 5.132 \\
GPT-5.4 Mini & Multimodal & 189 & 0.614 & 0.934 & 1.386 \\
GPT-5.4 Mini & Text & 189 & 0.942 & 0.966 & 0.005 \\
GPT-5.6-Sol & Image & 189 & 0.979 & 0.996 & 0.042 \\
GPT-5.6-Sol & Multimodal & 189 & 0.979 & 0.998 & 0.026 \\
GPT-5.6-Sol & Text & 189 & 0.963 & 0.995 & 0.063 \\
Qwen3-VL-8B & Image & 189 & 0.561 & 0.904 & 0.656 \\
Qwen3-VL-8B & Multimodal & 189 & 0.942 & 0.974 & 0.021 \\
Qwen3-VL-8B & Text & 189 & 0.323 & 0.544 & \textbf{0.000} \\
Qwen3.7-Plus & Image & 189 & 0.984 & 0.995 & 0.122 \\
Qwen3.7-Plus & Multimodal & 189 & \textbf{1.000} & \textbf{1.000} & \textbf{0.000} \\
Qwen3.7-Plus & Text & 189 & 0.947 & 0.972 & \textbf{0.000} \\
\bottomrule
\end{tabular}
\end{table*}

The aggregate results separate task coverage from executable consistency. Violations increase non-decreasingly from text to multimodal to image for four of six models (both Qwens, Haiku, and mini), but Claude-Opus-5 and GPT-5.6-Sol break this pattern: Opus-5 has fewer violations in image mode than multimodal ($3.889$ vs.\ $8.788$), and GPT-5.6-Sol has its fewest violations in multimodal mode ($0.026$). Success rate and completion rate split the models differently: three favor multimodal input (Qwen3-VL-8B, Qwen3.7-Plus, and GPT-5.6-Sol), whereas Haiku, Opus, and mini favor text. Qwen3.7-Plus attains perfect multimodal performance ($\text{SR}{=}1.000$, $\text{CR}{=}1.000$, zero violations), the highest in the table. Qwen3-VL-8B is the clearest coverage--consistency split: its multimodal SR is $0.942$, while text mode emits zero violations but completes only $0.544$ of tasks. High completion therefore does not by itself imply executable consistency.

\begin{table}[htbp]
\centering
\small
\caption{Sample-weighted SimWorld summary by input modality.}
\label{tab:app_simworld_modality_summary}
\begin{tabular}{lrrrr}
\toprule
Modality & Samples & SR $\uparrow$ & CR $\uparrow$ & Viol. $\downarrow$ \\
\midrule
Image & 1134 & 0.554 & 0.860 & 5.822 \\
Multimodal & 1134 & \textbf{0.800} & \textbf{0.969} & 2.786 \\
Text & 1134 & 0.784 & 0.905 & \textbf{1.148} \\
\bottomrule
\end{tabular}
\end{table}

Table~\ref{tab:app_simworld_modality_summary} pools these cells by modality. Multimodal prompting gives the highest complete-and-legal success rate ($0.800$) and completion rate ($0.969$), while text-only prompting produces the fewest violations ($1.148$ per sample). Relative to text, multimodal input gains $1.6$ success points and $6.4$ completion points but incurs $2.4\times$ as many violations, showing that improved coverage can coexist with weaker executable consistency. Image-only prompting is worst on all three aggregate measures ($\text{SR}{=}0.554$, $\text{CR}{=}0.860$, $5.822$ violations), although the per-model results show that this penalty is concentrated in the smaller models rather than shared uniformly.

\FloatBarrier

\subsection{Failure Attribution by Model}
\label{app:failure_attribution}
Section~\ref{ex:error_analysis} pools modalities within each model to summarize the overall error profile. Table~\ref{tab:app_failure_attrib} restores the model--modality breakdown using the same root-cause taxonomy and $189$ retained runs per cell ($13$ scenes $\times$ $5$ samples $\times$ $3$ task counts, minus the held-out open-layout cells). Each row sums to $100\%$.

The modality effect is not an artefact of pooling, but it is no longer universal. Because precedence and state-tracking errors are adjacent in the severity order and co-occur in a minority of runs, we read them together as \emph{sequencing} errors; their sum increases monotonically from text to multimodal to image-only for four of the six models, and precedence alone is monotone for five of six. The two exceptions are \texttt{GPT-5.6-Sol}, which inverts the ordering entirely ($3.2\%$ text, $1.6\%$ multimodal, $0.5\%$ image-only), and \texttt{Claude-Opus-5}, which peaks in multimodal mode ($44.4\%$) rather than image-only ($40.7\%$). The magnitude is strongly model-dependent and spans two orders of magnitude. \texttt{GPT-5.4-mini} and \texttt{Claude-Haiku-4.5} are the most fragile under image-only input ($84.1\%$ and $79.4\%$ sequencing errors, leaving success rates of $11.6\%$ and $12.2\%$), whereas \texttt{GPT-5.6-Sol} and \texttt{Qwen3.7-Plus} keep sequencing errors at $0.5\%$ and still succeed on $97.9\%$ and $98.4\%$ of image-only runs. \texttt{Qwen3-VL-8B} shows the opposite failure: its text-mode success rate ($32.3\%$) is the lowest of any model--modality cell despite it never emitting a root violation; $67.7\%$ of its text-mode runs are legal but incomplete, so it under-plans rather than mis-orders. \texttt{Claude-Opus-5} is the inverse: it leaves nothing incomplete in text mode yet reaches only $69.8\%$ success because $29.6\%$ of its runs carry a sequencing violation.

\begin{table}[htbp]
\centering
\small
\caption{SimWorld failure attribution by model and input modality (\% of runs, $N=189$ per cell). Categories follow Section~\ref{ex:error_analysis}: Prec.\ (delivery before pickup), State (duplicate pickup/delivery), Cap.\ (capacity exceeded), Gnd.\ (node outside graph or no parseable plan), Incomp.\ (legal plan, tasks remaining), Succ.\ (legal and complete). Each run is assigned a single root cause under the severity order Gnd.\ $>$ Cap.\ $>$ State $>$ Prec., so rows sum to $100\%$; columns are ordered for readability, not by severity.}
\label{tab:app_failure_attrib}
\begin{tabular}{llcccccc}
\toprule
Modality & Model & Prec.\ $\downarrow$ & State $\downarrow$ & Cap.\ $\downarrow$ & Gnd.\ $\downarrow$ & Incomp.\ $\downarrow$ & Succ.\ $\uparrow$ \\
\midrule
\multirow{6}{*}{Text-only}
    & \texttt{GPT-5.6-Sol} & 3.2 & \best{0.0} & \best{0.0} & \best{0.0} & 0.5 & \best{96.3} \\
    & \texttt{GPT-5.4-mini} & \best{0.0} & 0.5 & \best{0.0} & \best{0.0} & 5.3 & 94.2 \\
    & \texttt{Claude-Opus-5} & 16.4 & 13.2 & 0.5 & \best{0.0} & \best{0.0} & 69.8 \\
    & \texttt{Claude-Haiku-4.5} & 5.8 & 3.2 & \best{0.0} & 0.5 & 7.4 & 83.1 \\
    & \texttt{Qwen3.7-Plus} & \best{0.0} & \best{0.0} & \best{0.0} & \best{0.0} & 5.3 & 94.7 \\
    & \texttt{Qwen3-VL-8B} & \best{0.0} & \best{0.0} & \best{0.0} & \best{0.0} & 67.7 & 32.3 \\
\midrule
\multirow{6}{*}{Multimodal}
    & \texttt{GPT-5.6-Sol} & 1.1 & 0.5 & \best{0.0} & \best{0.0} & 0.5 & 97.9 \\
    & \texttt{GPT-5.4-mini} & 24.3 & 10.6 & \best{0.0} & 1.1 & 2.6 & 61.4 \\
    & \texttt{Claude-Opus-5} & 20.6 & 23.8 & 1.6 & 2.1 & 0.5 & 51.3 \\
    & \texttt{Claude-Haiku-4.5} & 15.3 & 7.4 & 0.5 & 0.5 & 1.1 & 75.1 \\
    & \texttt{Qwen3.7-Plus} & \best{0.0} & \best{0.0} & \best{0.0} & \best{0.0} & \best{0.0} & \best{100.0} \\
    & \texttt{Qwen3-VL-8B} & 1.1 & \best{0.0} & \best{0.0} & \best{0.0} & 4.8 & 94.2 \\
\midrule
\multirow{6}{*}{Image-only}
    & \texttt{GPT-5.6-Sol} & \best{0.5} & \best{0.0} & 0.5 & \best{0.0} & 1.1 & 97.9 \\
    & \texttt{GPT-5.4-mini} & 73.0 & 11.1 & 2.1 & 1.1 & 1.1 & 11.6 \\
    & \texttt{Claude-Opus-5} & 34.9 & 5.8 & 2.6 & 0.5 & \best{0.0} & 56.1 \\
    & \texttt{Claude-Haiku-4.5} & 63.5 & 15.9 & 1.1 & 2.1 & 5.3 & 12.2 \\
    & \texttt{Qwen3.7-Plus} & \best{0.5} & \best{0.0} & 1.1 & \best{0.0} & \best{0.0} & \best{98.4} \\
    & \texttt{Qwen3-VL-8B} & 16.4 & \best{0.0} & \best{0.0} & \best{0.0} & 27.5 & 56.1 \\
\bottomrule
\end{tabular}
\end{table}

\paragraph{TextWorld reward-gap attribution.}
Table~\ref{tab:app_tw_attrib} lists the numeric values plotted in Figure~\ref{fig:err_textworld}, together with the L1/L2 split of the lateness term. Lateness is identically zero for all $14$ models in L1, where tasks carry no deadlines, and accounts for $10.5$--$32.3\%$ of attainable reward within L2. The split shows that the deadline mechanism, not task volume, produces the temporal penalty. \texttt{Gemini-2.5-Flash} has the highest earned share ($93.4\%$), with its entire gap attributable to lateness. Only one of the $280$ runs violates the Type-C priority rule: \texttt{Llama-2-13B} delivers a non-C task while carrying a C task in M-L2 at task count $15$. At the lower end, \texttt{Qwen3-VL-8B} and \texttt{Llama-3-8B} instead lose $69.5\%$ and $83.6\%$ to undelivered tasks.

\begin{table}[htbp]
\centering
\small
\caption{TextWorld reward-gap attribution (\% of attainable reward $\sum_i \rho_i$), pooled over $4$ scenarios $\times$ $5$ task counts at seed $1$. Earned $+$ Not delivered $+$ Late $+$ Priority $= 100\%$ by construction. Late (L1) / Late (L2) decompose the lateness term by deadline regime; L1 scenarios have no deadlines. Runs voided by a priority violation are counted in the last column.}
\label{tab:app_tw_attrib}
\begin{tabular}{lcccccc}
\toprule
Model & Avg.\ $R$ $\uparrow$ & Earned $\uparrow$ & Not deliv.\ $\downarrow$ & Late $\downarrow$ & Late L2 $\downarrow$ & Priority $\downarrow$ \\
\midrule
\texttt{Gemini-2.5-Flash} & \best{265.73} & \best{93.4} & \best{0.0} & 6.6 & 13.5 & \best{0.0} \\
\texttt{Gemini-3-Pro} & 264.36 & 92.9 & \best{0.0} & 7.1 & 14.5 & \best{0.0} \\
\texttt{GPT-5.6-Sol} & 264.33 & 92.9 & \best{0.0} & 7.1 & 14.5 & \best{0.0} \\
\texttt{Claude-Opus-5} & 264.10 & 92.8 & \best{0.0} & 7.2 & 14.7 & \best{0.0} \\
\texttt{DeepSeek-v4-Flash} & 263.60 & 92.6 & 2.2 & \best{5.1} & \best{10.5} & \best{0.0} \\
\texttt{Qwen3.7-Plus} & 263.24 & 92.5 & \best{0.0} & 7.5 & 15.3 & \best{0.0} \\
\texttt{Claude-Sonnet-5} & 260.16 & 91.4 & \best{0.0} & 8.6 & 17.5 & \best{0.0} \\
\texttt{Claude-4.5-Haiku} & 253.01 & 88.9 & \best{0.0} & 11.1 & 22.7 & \best{0.0} \\
\texttt{Qwen3.6-27B} & 251.41 & 88.4 & 3.1 & 8.6 & 17.6 & \best{0.0} \\
\texttt{GPT-4o} & 251.05 & 88.2 & \best{0.0} & 11.8 & 24.1 & \best{0.0} \\
\texttt{GPT-5.4-Mini} & 246.81 & 86.7 & \best{0.0} & 13.3 & 27.1 & \best{0.0} \\
\texttt{Llama-2-13B} & 212.33 & 74.6 & 0.2 & 15.8 & 32.3 & 9.4 \\
\texttt{Qwen3-VL-8B} & 68.81 & 24.2 & 69.5 & 6.3 & 13.0 & \best{0.0} \\
\texttt{Llama-3-8B} & 19.33 & 6.8 & 83.6 & 9.6 & 19.6 & \best{0.0} \\
\bottomrule
\end{tabular}
\end{table}

\paragraph{Derived versus root violations.}
The evaluator reports a \texttt{deliver\_without\_carry} flag whenever a delivery targets an item not currently held. This condition is entailed by, rather than independent of, a precedence or duplicate-delivery error: every occurrence in the retained evaluation grid co-occurs with at least one of those root causes. Counting it separately would therefore inflate the apparent number of distinct error types without identifying an additional failure. Under the stated severity order, $731$ of $3{,}402$ runs are assigned to one of the four root-cause categories; all other runs are violation-free and are labeled complete or incomplete.

\FloatBarrier

\subsection{DualWorld-VQA Diagnostic Experiments}
\label{app:qa_examples}
DualWorld-VQA is a diagnostic suite for decomposing grounded planning errors into perceptual, topological, and decision-level failures. The suite contains 1,665 questions across 13 scenes. Questions are generated from scene annotations, node coordinates, navigation-distance matrices, and task metadata, and each item contains an unambiguous ground-truth answer. We avoid open-ended coordinate prediction and use exact match for categorical answers; sequence questions additionally report LCS-based auxiliary metrics, and planning questions are scored by whether the predicted feasible sequence stays within a relative distance tolerance of the ground truth.

\subsection{Question Families}
\label{app:question_families}
The suite contains four question families:
\begin{itemize}
    \item \textbf{Absolute axis sorting} asks models to order nodes along an image axis. This tests basic visual localization and label grounding while explicitly ignoring walls and route connectivity.
    \item \textbf{Euclidean distance comparison} includes anchor-based and pairwise variants. These questions test straight-line visual geometry, again without topological constraints.
    \item \textbf{Wall-illusion topology} contrasts Euclidean proximity with valid walking distance. These questions are designed so that the visually closer option can be topologically longer.
    \item \textbf{Capacity interleaving} requires a complete pickup--delivery sequence under capacity and precedence constraints, using true navigable distance as the optimization criterion.
\end{itemize}

\begin{table}[htbp]
\centering
\small
\caption{DualWorld-VQA dataset composition.}
\label{tab:app_vqa_dataset}
\begin{tabular}{lrrl}
\toprule
Question family & Level & Count & Primary metric \\
\midrule
Absolute axis sorting & 1 & 185 & Exact match, LCS \\
Euclidean anchor choice & 1 & 185 & Exact match \\
Euclidean pair choice & 1 & 185 & Exact match \\
Wall-illusion topology & 2 & 555 & Exact match \\
Capacity interleaving & 3 & 555 & Distance tolerance \\
\midrule
Total & -- & 1665 & -- \\
\bottomrule
\end{tabular}
\end{table}

\begin{table*}[htbp]
\centering
\small
\caption{DualWorld-VQA accuracy by question family.}
\label{tab:app_vqa_results}
\begin{tabular}{lrrrrrr}
\toprule
Model & Overall $\uparrow$ & Axis $\uparrow$ & Anchor $\uparrow$ & Pair $\uparrow$ & Topology $\uparrow$ & Planning $\uparrow$ \\
\midrule
Qwen3-VL-8B & 0.335 & 0.162 & 0.622 & 0.535 & 0.566 & 0.000 \\
Qwen3.7-Plus & 0.843 & \textbf{1.000} & \textbf{1.000} & \textbf{1.000} & \textbf{0.793} & 0.737 \\
Claude Haiku 4.5 & 0.227 & 0.005 & 0.232 & 0.162 & 0.530 & 0.018 \\
Claude Opus 5 & 0.799 & \textbf{1.000} & \textbf{1.000} & 0.995 & 0.778 & 0.622 \\
GPT-5.4 Mini & 0.392 & 0.422 & 0.589 & 0.654 & 0.559 & 0.063 \\
GPT-5.6-Sol & \textbf{0.864} & 0.968 & 0.989 & 0.995 & 0.775 & \textbf{0.832} \\
\bottomrule
\end{tabular}
\end{table*}

The results reveal a generation-dependent hierarchy rather than a uniform perception-to-planning decline. \texttt{GPT-5.6-Sol}, \texttt{Qwen3.7-Plus}, and \texttt{Claude-Opus-5} nearly saturate the three Euclidean perception families, but topology remains at $77.5$--$79.3\%$ and planning spans $62.2$--$83.2\%$. In contrast, \texttt{GPT-5.4-Mini}, \texttt{Qwen3-VL-8B}, and \texttt{Claude-Haiku-4.5} fall to $0.0$--$6.3\%$ on planning. Failure inspection further separates the causes: \texttt{Qwen3-VL-8B} usually outputs the wrong task-node set, the two smaller proprietary models mainly produce feasible but overly long routes, and the newer models' residual errors are dominated by route suboptimality. DualWorld-VQA therefore distinguishes instruction grounding, constraint satisfaction, and route optimization rather than attributing every failure to one integration bottleneck.

\begin{figure}[htbp]
    \centering
    \begin{subfigure}[b]{0.32\textwidth}
        \centering
        \includegraphics[width=\textwidth]{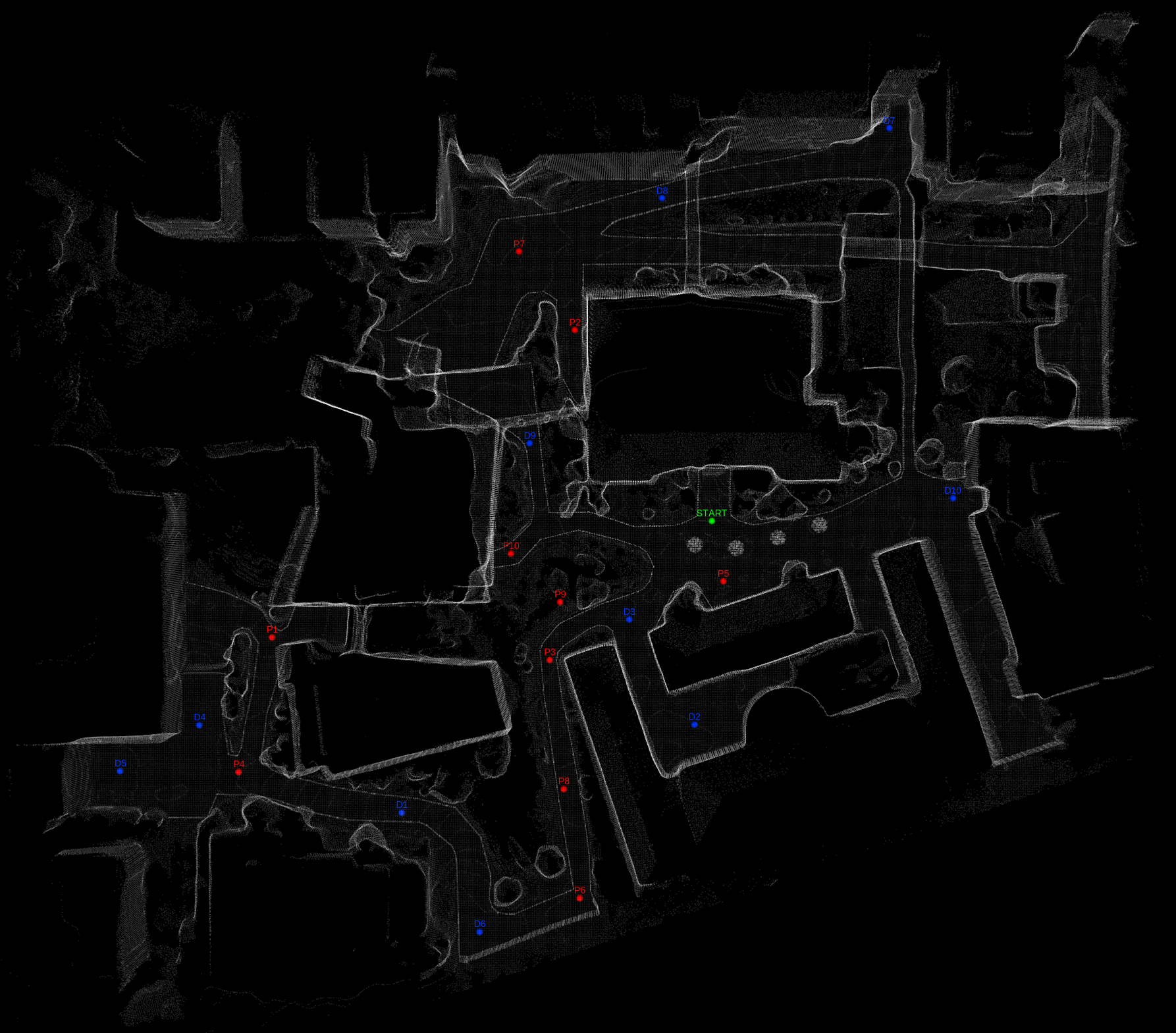}
        \caption{Campus.}
        \label{fig:app_vqa_campus}
    \end{subfigure}
    \hfill
    \begin{subfigure}[b]{0.32\textwidth}
        \centering
        \includegraphics[width=\textwidth]{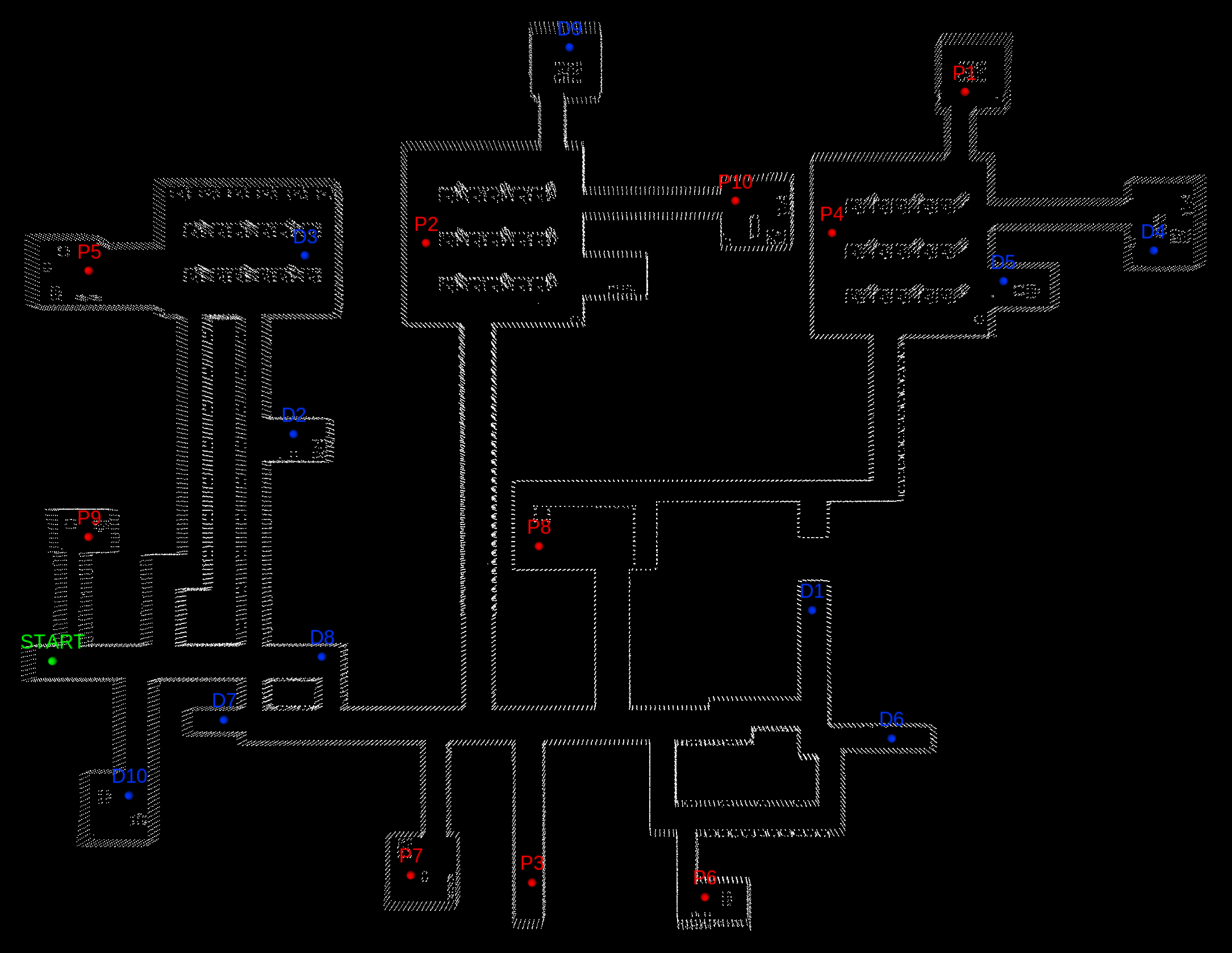}
        \caption{Indoor.}
        \label{fig:app_vqa_indoor}
    \end{subfigure}
    \hfill
    \begin{subfigure}[b]{0.32\textwidth}
        \centering
        \includegraphics[width=\textwidth]{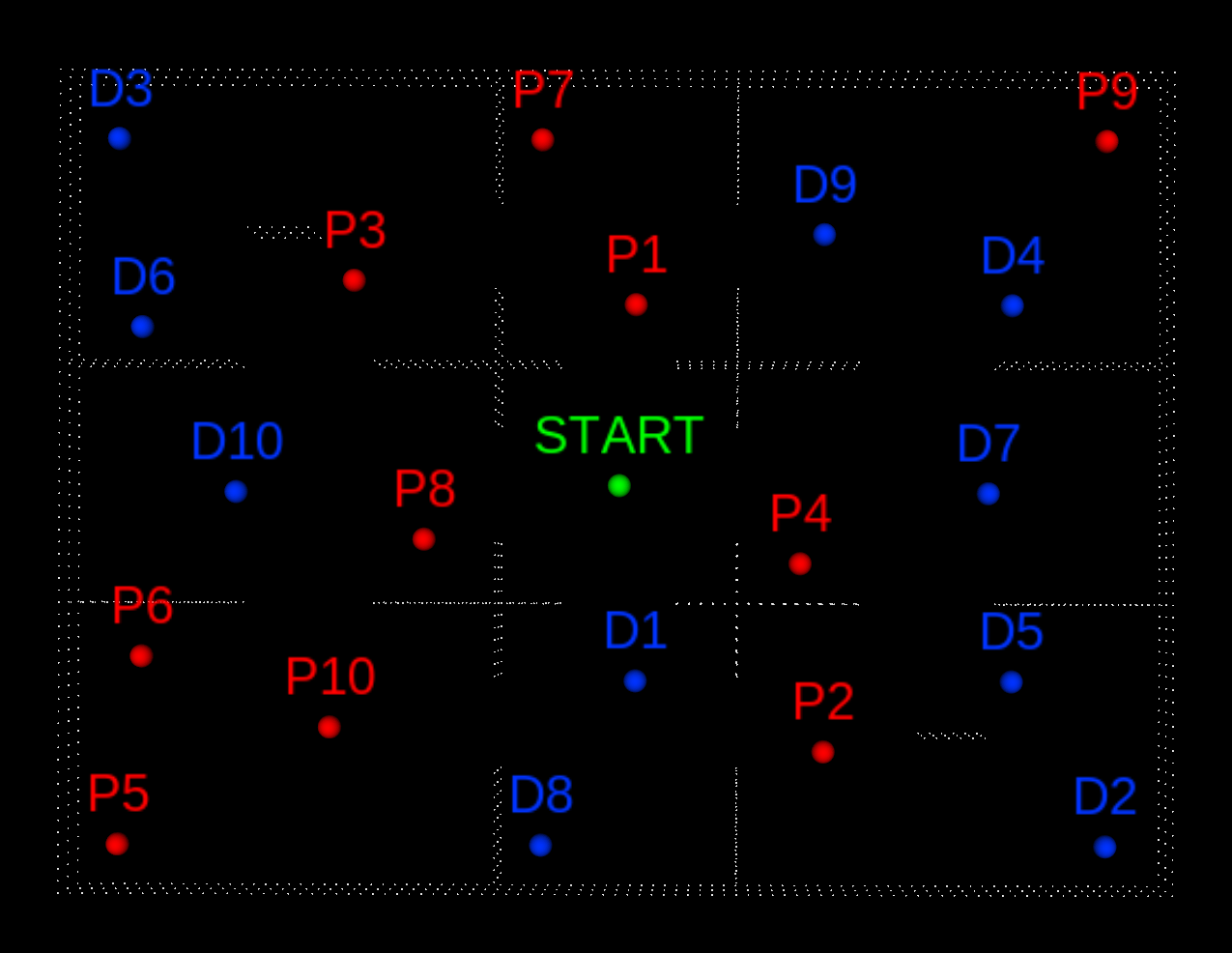}
        \caption{T1 cross suite.}
        \label{fig:app_vqa_t1}
    \end{subfigure}
    \caption{Annotated BEV observations used in DualWorld-VQA. Red markers indicate pickup nodes, blue markers indicate delivery nodes, and the green marker indicates the start position.}
    \label{fig:app_vqa_cases}
\end{figure}

\subsection{Representative Cases}
\label{app:vqa_cases}
\begin{tcolorbox}[colback=gray!5, colframe=black!50, title=\textbf{Level 1: Axis Sorting}]
\small
\textbf{Question:} Use only the map image and compare the nodes by their absolute vertical position. Order START, P9, D9, P4, P3 from top to bottom. Ignore walls, corridors, connectivity, and any numbering pattern in the labels. \\
\textbf{Expected format:} node labels joined by \texttt{->}. \\
\textbf{Ground truth:} D9 $\to$ P4 $\to$ START $\to$ P3 $\to$ P9.
\end{tcolorbox}

\begin{tcolorbox}[colback=gray!5, colframe=black!50, title=\textbf{Level 1: Euclidean Distance}]
\small
\textbf{Question:} Use only the map image and compare straight-line Euclidean distances. Among D5, P1, D4, P3, and P4, which node is closest to P2? Ignore walls, corridors, connectivity, and numbering patterns. \\
\textbf{Expected format:} one node label. \\
\textbf{Ground truth:} D4.
\end{tcolorbox}

\begin{tcolorbox}[colback=gray!5, colframe=black!50, title=\textbf{Level 2: Wall-Illusion Topology}]
\small
\textbf{Question:} Considering only valid walking routes that do not pass through walls, which path is longer: P6 $\to$ P1 or P6 $\to$ D7? \\
\textbf{Expected format:} ``P6 $\to$ X is longer''. \\
\textbf{Ground truth:} P6 $\to$ D7 is longer. The corresponding navigable distances are $209.399\,\mathrm{m}$ and $518.870\,\mathrm{m}$, while the Euclidean distances are $141.481\,\mathrm{m}$ and $129.861\,\mathrm{m}$, respectively. This case deliberately reverses the Euclidean and topological orderings.
\end{tcolorbox}

\begin{tcolorbox}[colback=gray!5, colframe=black!50, title=\textbf{Level 3: Capacity-Constrained Interleaving}]
\small
\textbf{Question:} Start at START with empty load and capacity $K=2$. Complete three orders: Task A is P4 $\to$ D4, Task B is P6 $\to$ D6, and Task C is P7 $\to$ D7. Never exceed capacity and never deliver before pickup. Find the visiting sequence that minimizes walking distance. \\
\textbf{Expected format:} node labels joined by \texttt{->}. \\
\textbf{Ground truth:} START $\to$ P6 $\to$ P4 $\to$ D6 $\to$ P7 $\to$ D4 $\to$ D7. The optimal distance is $942.430\,\mathrm{m}$, compared with $1718.843\,\mathrm{m}$ for the best non-batched simple sequence.
\end{tcolorbox}

\section{Solver Details}
\label{app:solvers}

This section specifies the deterministic solvers used to contextualize model performance.
All three solvers operate on the same pickup--delivery abstraction but differ in their
information assumptions and computational budget. DFS provides an anytime real-time
planner, Online DP provides an exact non-learning policy under currently visible tasks,
and Offline DP provides an oracle upper bound with full task knowledge.

\paragraph{Unified task abstraction.}
For a task set $\mathcal{T}=\{1,\dots,N\}$, each task $i$ has pickup location $p_i$,
delivery location $d_i$, release time $t^{tr}_i$, deadline $e_i$, reward $R_i$, and
type $\tau_i\in\{A,B,C\}$. Given start location $s$, the solvers build a shared point
set
\begin{equation}
\mathcal{P}=\{s\}\cup\{p_i\}_{i=1}^N\cup\{d_i\}_{i=1}^N .
\end{equation}
An action consists of moving to one point in $\mathcal{P}$ and executing either
\texttt{pickup} or \texttt{deliver}. Travel time is distance divided by speed; TextWorld
uses Euclidean distance, while SimWorld replaces it with the corresponding navigation
distance when a grounded distance matrix is available.

\paragraph{Shared constraints and objective.}
All solvers enforce the same feasibility rules: a task cannot be picked up twice, a
delivery is valid only after pickup, delivered tasks cannot be revisited, and the number
of carried tasks cannot exceed capacity $K$. We also enforce the Type-C priority rule:
while any Type-C task is carried, an A/B task cannot be delivered; intervening pickups
and the delivery of any carried Type-C task remain valid. At pickup time, the agent may wait until the task release
time; delivery receives no waiting benefit. The realized revenue for delivering task
$i$ at time $t$ is
\begin{equation}
r_i(t)=\max\{0, R_i-\lambda_{\tau_i}\max(0,t-e_i)\},
\end{equation}
with $\lambda_A=0.1$, $\lambda_B=0.2$, and $\lambda_C=0.3$.

\begin{table}[htbp]
\centering
\small
\caption{Roles of deterministic solvers used in the benchmark.}
\label{tab:app_solver_roles}
\begin{tabular}{llll}
\toprule
Solver & Task visibility & Optimality & Primary role \\
\midrule
DFS & Currently visible & Anytime heuristic & Real-time feasible planning \\
Online DP & Currently visible & Exact over visible tasks & Strong online baseline \\
Offline DP & Full task set & Exact with hindsight & Upper bound and label generation \\
\bottomrule
\end{tabular}
\end{table}

\begin{figure*}[h]
  \centering
  \begin{minipage}{0.49\linewidth}
    \centering
    \includegraphics[width=0.94\linewidth]{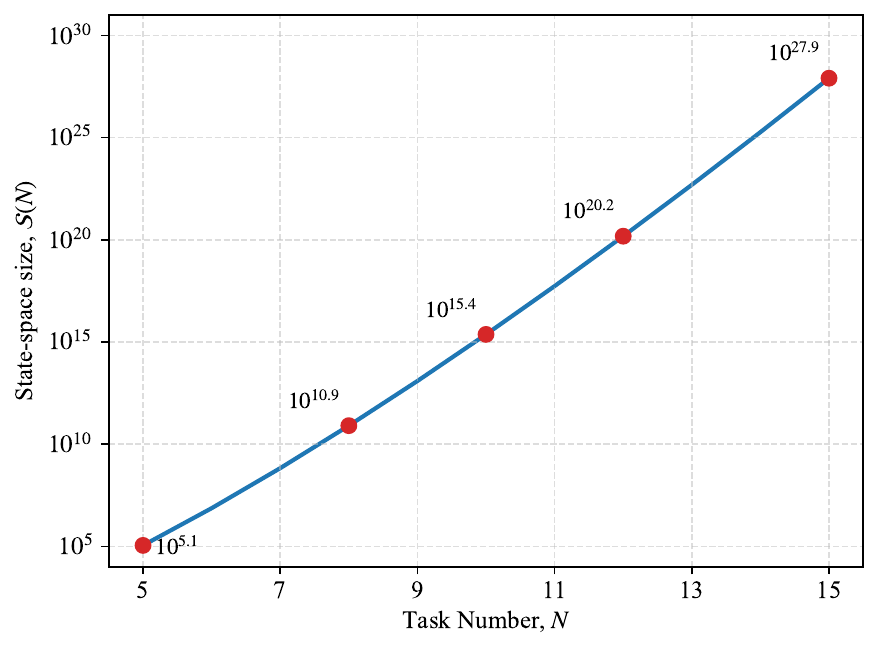}
    \caption{\textbf{State-space complexity.} The size of the valid permutation space grows super-exponentially with the number of tasks $N$.}
    \label{fig:state_space}
  \end{minipage}\hfill
  \begin{minipage}{0.49\linewidth}
    \centering
    \includegraphics[width=\linewidth]{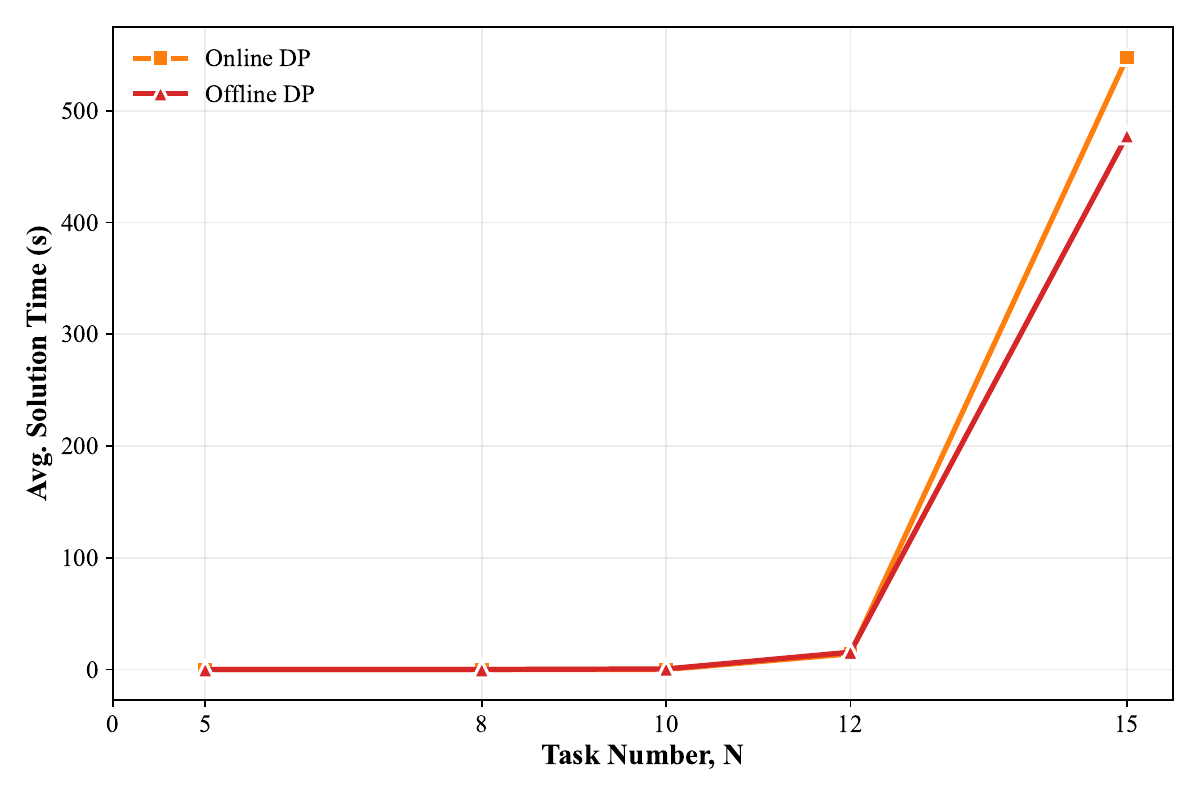}
    \caption{Average solver runtime versus task size $N$. Runtime increases sharply with $N$ and follows an
approximately exponential trend, reflecting the combinatorial growth of the DP space.}
    \label{fig:solver_time}
  \end{minipage}
\end{figure*}

\paragraph{State-space growth.}
The pickup--delivery problem has a natural permutation-based structure. If geometric
details are ignored and only pickup-before-delivery precedence is considered, the number
of feasible visit orders for $N$ tasks scales as
\begin{equation}
|\mathcal{S}_N| \;=\; \frac{(2N)!}{2^N},
\end{equation}
because each task contributes one pickup and one delivery and half of the relative
orders for each pair violate precedence. Fig.~\ref{fig:state_space} shows this rapid
combinatorial growth. Fig.~\ref{fig:solver_time} further confirms that runtime increases
sharply with $N$, which motivates separating real-time heuristic planning from exact
oracle-style evaluation.

\subsection{DFS}
\label{bpp:dfs}
DFS is the real-time planner. It is designed to return a feasible plan under a fixed
time budget rather than to prove global optimality.

\paragraph{Warm start.}
The planner first constructs a greedy solution to initialize the incumbent revenue and
path. The greedy policy scores candidate actions by expected finish time, immediate
delivery revenue, and priority-task preference, giving the subsequent search a strong
lower bound.

\paragraph{Branch-and-bound search.}
After warm start, DFS explores feasible action sequences in heuristic order. For each
partial plan, it estimates an optimistic upper bound on remaining revenue using a
geometry-based lower bound on travel time and the same deadline penalty model as the
evaluator. A node is pruned when
$\text{current\_revenue}+\text{upper\_bound}\le\text{best\_revenue}$. The action
generator enforces capacity, precedence, and Type-C priority constraints before any
recursive expansion.

\begin{breakablealgorithm}
\caption{Greedy Warm-up + DFS Branch-and-Bound algorithm}
\label{alg:algo}
\begin{algorithmic}[1]
\Statex \textbf{Input:} Context $(t_0,s,v,K,\mathcal{C}_0,\mathcal{U}_0)$, time limit $T_{lim}$. 
\Statex \textbf{Output:} A plan list (\texttt{move} + \texttt{pickup/deliver}) with high revenue.
\State Merge tasks $\mathcal{T}\gets \mathcal{C}_0\cup\mathcal{U}_0$ and index them $1..N$.
\If{$N=0$} \State \Return $\emptyset$ \EndIf
\State Initialize best revenue $best \gets -\infty$, best path $path^*\gets []$.
\State $carrying \gets$ indices of $\mathcal{C}_0$; $pending \gets$ indices of $\mathcal{U}_0$; $load\gets |carrying|$.
\State \Call{GreedyWarmUp}{$s,t_0,load,carrying,pending$} \Comment{sets baseline $best$ and $path^*$}
\If{time remains}
  \State \Call{DFS}{$s,t_0,load,0,carrying,pending,[]$}
\EndIf
\State \Return \Call{FormatOutput}{$path^*$}  \Comment{insert \texttt{move} before each action}
\\
\Function{GreedyWarmUp}{$pos,t,load,carrying,pending$}
  \State $rev\gets 0$, $path\gets []$
  \While{$carrying\neq\emptyset$ \textbf{or} $pending\neq\emptyset$}
    \State $A \gets$ \Call{ValidActions}{$pos,t,load,carrying,pending$}
    \If{$A=\emptyset$} \State \textbf{break} \EndIf
    \State Pick $a^* \gets \arg\min_{a\in A} score(a)$
    \State Apply $a^*$: update $(pos,t,load)$ and sets $(carrying,pending)$; $rev\gets rev + revenue(a^*)$
    \State Append $a^*$ to $path$
  \EndWhile
  \If{$rev>best$} \State $best\gets rev$; $path^*\gets path$ \EndIf
\EndFunction
\\
\Function{DFS}{$pos,t,load,rev,carrying,pending,path$}
  \If{timeout reached} \State \Return \EndIf
  \If{$carrying=\emptyset$ \textbf{and} $pending=\emptyset$}
    \If{$rev>best$} \State $best\gets rev$; $path^*\gets path$ \EndIf
    \State \Return
  \EndIf
  \State $ub \gets$ \Call{EstimatePotential}{$pos,t,carrying,pending$}
  \If{$rev+ub \le best$} \State \Return \EndIf
  \State $A \gets$ \Call{ValidActions}{$pos,t,load,carrying,pending$}; sort $A$ by increasing $score$
  \For{\textbf{each} $a \in A$}
    \State Recurse on next state after applying $a$
  \EndFor
\EndFunction
\\
\Function{ValidActions}{$pos,t,load,carrying,pending$}
  \State $A\gets []$
  \State $hasC \gets$ \textbf{true} iff $\exists i\in carrying: \tau_i=C$
  \If{$load < K$}
    \For{$i\in pending$}
      \State $t_p \gets \max(t + dist(pos,p_i)/v,\; t^{tr}_i)$
      \State Define pickup action $a$ with $finish=t_p$, $revenue=0$, $score=t_p - \mathbb{I}[\tau_i=C]\cdot \alpha$
      \State Append $a$ to $A$
    \EndFor
  \EndIf
  \For{$i\in carrying$}
    \If{$hasC \land \tau_i\neq C$} \State \textbf{continue} \EndIf
    \State $t_d \gets t + dist(pos,d_i)/v$
    \State $rev_i \gets \max(0, R_i - \max(0,t_d-t^{end}_i)\cdot rate(\tau_i))$
    \State Define deliver action $a$ with $finish=t_d$, $revenue=rev_i$, $score=t_d - \beta\cdot rev_i - \mathbb{I}[\tau_i=C]\cdot \gamma$
    \State Append $a$ to $A$
  \EndFor
  \State \Return $A$
\EndFunction
\\
\Function{EstimatePotential}{$pos,t,carrying,pending$}
  \State $ub\gets 0$
  \For{$i\in carrying \cup pending$}
    \State Compute optimistic minimal travel time to complete $i$ (deliver-only if carrying; pickup$\rightarrow$deliver if pending)
    \State Convert to optimistic arrival $t_{min}$ and optimistic revenue $\widehat{rev}_i$
    \State $ub \gets ub + \widehat{rev}_i$
  \EndFor
  \State \Return $ub$
\EndFunction
\end{algorithmic}
\end{breakablealgorithm}

\subsection{Online DP}
\label{bpp:onlinedp}
Online DP is the exact non-learning baseline under the online information constraint.
At each decision point, it observes the current robot state, carried tasks, and currently
released task pool, solves the induced finite-horizon problem exactly, and executes the
first action of the resulting plan.

\paragraph{State compression.}
The DP state is $(D,C,u)$, where $D$ is a bitmask of delivered tasks, $C$ is a bitmask
of carried tasks, and $u$ is the current point index in $\mathcal{P}$. Each state stores
a pair $(\text{reward},\Delta t)$ and is compared lexicographically by maximum reward
and then minimum elapsed time.

\paragraph{Online interpretation.}
Because unreleased future tasks are not visible, Online DP is optimal only with respect
to the currently observable task set. It therefore serves as a strong online baseline
without hindsight. Its worst-case number of state transitions is on the order of
$O(N^2 3^N)$, making it useful for moderate-size instances but unsuitable as a scalable
real-time controller for large $N$.
\begin{breakablealgorithm}
\caption{Online DP Planner}
\label{alg:online-dp-matched}
\begin{algorithmic}[1]
\Statex \textbf{Input:} Context: current time $t_0$, start position $s$, speed $v$, capacity $K$,
carried tasks $\mathcal{C}_0$, available tasks $\mathcal{U}_0$.
\Statex Each task $i$ has fields $(id_i, p_i, d_i, t^{end}_i, R_i, type_i)$ where $type_i\in\{A,B,C,\dots\}$.
\Statex Penalty rate function $pen(type)$.
\Statex \textbf{Output:} action list: \texttt{move} + \texttt{pickup/deliver}.

\State $\mathcal{T} \gets \mathcal{C}_0 \cup \mathcal{U}_0$; $N \gets |\mathcal{T}|$ (use 0-based task index $i=0..N-1$).
\If{$N=0$} \State \Return $\emptyset$ \EndIf

\State Build points list $P$:
\State \hspace{1em} $P[0]\gets s$
\State \hspace{1em} for each task $i$: append pickup $p_i$ to $P$  \Comment{$P[1+i]=p_i$}
\State \hspace{1em} for each task $i$: append deliver $d_i$ to $P$ \Comment{$P[1+N+i]=d_i$}
\State $|P| = 1+2N$.

\State Build initial carried bitmask $C_0$ from $\mathcal{C}_0$; set delivered bitmask $D_0\gets 0$.
\State Initialize $dp[(D_0,C_0,0)]\gets(rev=0,\Delta t=0)$ and parent map $\pi\gets\emptyset$.

\For{$D=0$ to $2^N-1$}
  \For{$C=0$ to $2^N-1$}
    \If{$(D \,\&\, C)\neq 0$} \State \textbf{continue} \EndIf
    \For{$u=0$ to $|P|-1$}
      \If{$(D,C,u)\notin dp$} \State \textbf{continue} \EndIf
      \State $(rev,\Delta t)\gets dp[(D,C,u)]$

      \State $priority \gets$ true iff $\exists i: ((C\gg i)\&1=1)\land(type_i=\texttt{'C'})$
      \State $F \gets \{\,1+N+i \mid ((C\gg i)\&1=1)\land(type_i=\texttt{'C'})\,\}$
      \State $load \gets popcount(C)$

      \For{$w=1$ to $|P|-1$}
        \If{$priority \land w\notin F$} \State \textbf{continue} \EndIf

        \State Decode $(i,action)$ from $w$:
        \If{$1 \le w < 1+N$} \State $i\gets w-1,\;\; action\gets pickup$
        \Else \State $i\gets w-(1+N),\;\; action\gets deliver$
        \EndIf

        \If{$((D\gg i)\&1)=1$} \State \textbf{continue} \EndIf
        \If{$action=pickup \land ((C\gg i)\&1)=1$} \State \textbf{continue} \EndIf
        \If{$action=deliver \land ((C\gg i)\&1)=0$} \State \textbf{continue} \EndIf
        \If{$action=pickup \land load\ge K$} \State \textbf{continue} \EndIf

        \State $\Delta t' \gets \Delta t + \frac{dist(P_u,P_w)}{v}$

        \If{$action=pickup$}
          \State $D'\gets D,\;\; C'\gets C\cup\{i\},\;\; gain\gets 0$
        \Else
          \State $t_{arr}\gets t_0 + \Delta t'$
          \State $late\gets \max(0,t_{arr}-t^{end}_i)$
          \State $gain\gets \max(0,R_i - late\cdot pen(type_i))$
          \State $D'\gets D\cup\{i\},\;\; C'\gets C\setminus\{i\}$
        \EndIf

        \State $rev'\gets rev + gain$
        \If{$(rev',\Delta t')$ better than $dp[(D',C',w)]$ (max $rev'$, tie min $\Delta t'$)}
          \State Update $dp[(D',C',w)]\gets(rev',\Delta t')$ and $\pi[(D',C',w)]\gets(D,C,u)$
        \EndIf
      \EndFor
    \EndFor
  \EndFor
\EndFor

\State Choose best state in $dp$ by max revenue (tie min time), then backtrack $\pi$ to get (task, action).
\State Convert to executable plan: each step append \texttt{move(to target)} then \texttt{pickup/deliver(task)}.
\State \Return plan list.
\end{algorithmic}
\end{breakablealgorithm}

\subsection{Offline DP}
\label{bpp:offlinedp}
Offline DP is the hindsight oracle used for upper-bound analysis and label generation.
It receives the full task set, including future releases, and optimizes over the complete
sequence space subject to the same feasibility and reward rules.

\paragraph{Bellman formulation.}
The state is $S=(m_D,m_C,\ell)$, where $m_D$ and $m_C$ are delivered and carried
bitmasks, and $\ell$ is the index of the last visited point in $\mathcal{P}$. The value
function satisfies
\begin{equation}
    V(S) = \max_{a \in \mathcal{A}(S)} \left\{ R(S, a) + V(S') \right\}
\end{equation}
where $\mathcal{A}(S)$ is the feasible action set and $S'$ is the successor state. As
in Online DP, each DP entry stores $(\text{reward},\text{time})$ and is optimized
lexicographically by reward first and finishing time second.

\paragraph{Use in evaluation.}
Offline DP is not an executable online policy because it has access to future task
releases. We use it only to construct oracle plans, normalize regret, and separate model
planning errors from the intrinsic difficulty of each task instance.

\begin{breakablealgorithm}
\caption{Offline DP Solver}
\label{alg:godview-dp-matched}
\begin{algorithmic}[1]
\Statex \textbf{Input:} Tasks $\mathcal{T}=\{0,\dots,N-1\}$ with pickup $p_i$, deliver $d_i$, trigger $t^{tr}_i$, deadline $t^{end}_i$,
reward $R_i$, type $\tau_i\in\{\texttt{'A'},\texttt{'B'},\texttt{'C'}\}$; start position $s$; speed $v$; capacity $K$.
\Statex Penalty rate $rate(\tau)$.
\Statex \textbf{Output:} An action sequence (pickup/deliver) maximizing total revenue, tie-break on minimal finishing time.

\State Sort tasks by $t^{tr}_i$. Build points list:
\State \hspace{1em} $P[0]\gets s$
\State \hspace{1em} $P[1+i]\gets p_i$ for $i=0..N-1$
\State \hspace{1em} $P[1+N+i]\gets d_i$ for $i=0..N-1$ \Comment{so deliver indices are $N+1..2N$}

\State Initialize DP: $dp[(0,0,0)]\gets(0,0)$
\Comment{$(D,M,u)\mapsto(rev,time)$, where $D$=delivered mask, $M$=carried mask}
\State Initialize parent pointers $\pi \gets \emptyset$.

\For{$D=0$ to $2^N-1$}
  \For{$M=0$ to $2^N-1$}
    \If{$(D \,\&\, M)\neq 0$} \State \textbf{continue} \EndIf
    \For{$u=0$ to $|P|-1$}
      \If{$(D,M,u)\notin dp$} \State \textbf{continue} \EndIf
      \State $(rev,t)\gets dp[(D,M,u)]$

      \State $priority \gets$ true iff $\exists i: ((M\gg i)\&1=1)\land(\tau_i=\texttt{'C'})$
      \State $F \gets \{\,1+N+i \mid ((M\gg i)\&1=1)\land(\tau_i=\texttt{'C'})\,\}$ \Comment{forced deliver points of C}

      \For{$w=1$ to $|P|-1$}
        \If{$priority \land w\notin F$} \State \textbf{continue} \EndIf

        \State Decode $(i, action)$ from $w$:
        \If{$1 \le w \le N$} \State $i\gets w-1,\;\; action\gets pickup$
        \Else \State $i\gets w-(N+1),\;\; action\gets deliver$ \EndIf

        \If{$((D\gg i)\&1)=1$} \State \textbf{continue} \EndIf
        \If{$action=pickup \land ((M\gg i)\&1)=1$} \State \textbf{continue} \EndIf
        \If{$action=pickup \land popcount(M)\ge K$} \State \textbf{continue} \EndIf
        \If{$action=deliver \land ((M\gg i)\&1)=0$} \State \textbf{continue} \EndIf

        \State $t_{arr} \gets t + \frac{dist(P_u,P_w)}{v}$
        \If{$action=pickup$}
          \State $t' \gets \max(t_{arr}, t^{tr}_i)$ \Comment{wait for trigger}
          \State $D' \gets D,\;\; M' \gets M \cup \{i\},\;\; \Delta \gets 0$
        \Else
          \State $t' \gets t_{arr}$
          \State $late \gets \max(0, t' - t^{end}_i)$
          \State $\Delta \gets \max(0, R_i - late\cdot rate(\tau_i))$
          \State $D' \gets D \cup \{i\},\;\; M' \gets M \setminus \{i\}$
        \EndIf

        \State $rev' \gets rev + \Delta$
        \If{$(rev',t')$ better than $dp[(D',M',w)]$ (max $rev'$, tie min $t'$)}
          \State $dp[(D',M',w)] \gets (rev', t')$
          \State $\pi[(D',M',w)] \gets (D,M,u)$
        \EndIf
      \EndFor
    \EndFor
  \EndFor
\EndFor

\State Select best final state by max revenue (tie min time), then backtrack $\pi$ to reconstruct.
\State \Return reconstructed action sequence.
\end{algorithmic}
\end{breakablealgorithm}

\section{Prompt Design}
\label{app:prompt}

All prompts specify the available task information, hard feasibility constraints, and a machine-readable output schema. The prompts are designed to reduce evaluation ambiguity rather than to provide hidden solver assistance. They enforce pickup--delivery precedence and prohibit duplicate task actions. TextWorld requires explicit movement commands, whereas SimWorld uses node sequences that the evaluator expands into executable actions. Placeholders enclosed in braces are instantiated for each evaluation instance.

\subsection{Prompting Protocol}
\label{dpp:protocol}

\begin{table}[htbp]
\centering
\small
\caption{Prompt inputs and output interfaces.}
\label{tab:app_prompt_interfaces}
\resizebox{\textwidth}{!}{%
\begin{tabular}{llll}
\toprule
Setting & Inputs & Planning signal & Output \\
\midrule
TextWorld & State, task list & Symbolic reward, time, deadline & JSON actions \\
SimWorld-text & State, task list, node coordinates & Coordinate proxy & JSON node sequence \\
SimWorld-image & BEV image & Visual topology & JSON node sequence \\
SimWorld-multimodal & BEV image, task list, node mapping & Visual topology and task grounding & JSON node sequence \\
\bottomrule
\end{tabular}
}
\end{table}

TextWorld outputs executable JSON actions. All SimWorld modalities use a shared JSON node-sequence schema, differing in their input information. The evaluator resolves node labels to task identifiers and coordinates and inserts movement commands as needed before pickup or delivery.

\subsection{TextWorld Template}
\label{dpp:text}

The TextWorld prompt below is the complete template used for symbolic delivery planning, typeset with the same boxed style as the SimWorld templates.

\begin{tcolorbox}[
    enhanced,
    breakable,
    colback=gray!5,
    colframe=black!70,
    title=\textbf{TextWorld Prompt Template},
    fonttitle=\bfseries
]
\small
\textbf{Role.} You are the central planning engine for an intelligent delivery robot. Generate a strictly sequential, physically valid, and economically optimal action plan.

\vspace{0.1cm}
\textbf{1. System State}
\begin{itemize}[leftmargin=*, label=-, noitemsep, topsep=2pt]
    \item Speed: \{speed:.2f\} m/s; time: \{now\_time\} s; position: \{\{``x'': \{cur\_x\}, ``y'': \{cur\_y\}\}\}.
    \item Capacity: \{capacity\}/\{max\_capacity\}; carried tasks: \{carried\_ids\}.
    \item Interruption or replanning context: \{interrupt\_section\}.
\end{itemize}

\textbf{2. Rules and Constraints}
\begin{enumerate}[leftmargin=*, label=\arabic*., noitemsep, topsep=2pt]
    \item Accept and deliver all tasks in the pool, even if late.
    \item Revenue is $\max(0,\text{reward}-\text{lateness}\times\text{penalty\_rate})$.
    \item Penalty rates are Type-A $0.1$, Type-B $0.2$, and Type-C $0.3$ per second.
    \item If carrying a Type-C task, deliver Type-C before Type-A or Type-B tasks.
    \item Before every pickup or delivery, move to the target coordinate first.
    \item Multiple pickups may be batched if the capacity constraint is respected.
    \item Travel time is distance divided by speed; distance is Euclidean distance in meters.
\end{enumerate}

\textbf{3. Available Tasks}
\begin{tcolorbox}[colback=white, colframe=black!20, sharp corners, boxrule=0.5pt, breakable]
\ttfamily\small
\{tasks\}
\end{tcolorbox}

\textbf{4. Output Format}
Return strict JSON only with an \texttt{actions} array:
\begin{tcolorbox}[colback=white, colframe=black!20, sharp corners, boxrule=0.5pt, breakable]
\ttfamily\scriptsize
\{ \\
\hspace*{0.8em}``actions'': [ \\
\hspace*{1.6em}\{ \\
\hspace*{2.4em}``action'': ``move'', \\
\hspace*{2.4em}``to'': \{``x'': float, ``y'': float\}, \\
\hspace*{2.4em}``distance\_m'': float, \\
\hspace*{2.4em}``duration\_s'': float, \\
\hspace*{2.4em}``arrival\_time'': float \\
\hspace*{1.6em}\}, \\
\hspace*{1.6em}\{``action'': ``pickup'', ``task\_id'': ``string'', ``time\_s'': float\}, \\
\hspace*{1.6em}\{``action'': ``deliver'', ``task\_id'': ``string'', ``time\_s'': float, ``revenue'': float\} \\
\hspace*{0.8em}] \\
\}
\end{tcolorbox}
\end{tcolorbox}

\subsection{SimWorld Templates}
\label{dpp:sim}

SimWorld prompts share the same pickup--delivery constraints and node-sequence output schema but differ in available evidence. Text-only planning uses coordinates as a travel-cost proxy, image-only planning uses the annotated BEV map, and multimodal planning combines the map with structured task information. The evaluator supplies executable coordinates and movement commands.

\begin{tcolorbox}[
    enhanced,
    breakable,
    colback=gray!5,
    colframe=black!70,
    title=\textbf{SimWorld Text-Only Prompt Skeleton},
    fonttitle=\bfseries
]
\small
\textbf{Role.} Generate a pickup--delivery node-visiting sequence using structured task text without a map image.

\textbf{Inputs.}
\begin{itemize}[leftmargin=*, label=-, noitemsep, topsep=2pt]
    \item Time: \{now:.2f\} s; position: \{\{``x'': \{cur\_x:.2f\}, ``y'': \{cur\_y:.2f\}\}\}; speed: \{speed:.2f\} m/s.
    \item Load: \{capacity\}/\{max\_capacity\}.
    \item Carried tasks: \{carried\_tasks\_json\}.
    \item Available tasks: \{new\_tasks\_json\}.
    \item Task-to-node mapping: \{id\_mapping\_section\}.
\end{itemize}

\textbf{Planning rule.} Minimize estimated travel cost while completing as many feasible tasks as possible. Because the BEV topology is unavailable, use coordinates cautiously and avoid purely greedy nearest-neighbor plans.

\textbf{Constraints.} Pn must precede the matching Dn unless the task is already carried; each pickup and delivery node may appear at most once; already-carried tasks require only delivery; load must never exceed \{max\_capacity\}; use only nodes in the task-to-node mapping.

\textbf{Output schema.} Return only a JSON object with a \texttt{sequence} array of node labels. Do not output coordinates or movement commands.
\begin{tcolorbox}[colback=white, colframe=black!20, sharp corners, boxrule=0.5pt]
\ttfamily\small
\{``sequence'': [``START'', ``P1'', ``P3'', ``D3'', ``D1'']\}
\end{tcolorbox}
\end{tcolorbox}

\begin{tcolorbox}[
    enhanced,
    breakable,
    colback=gray!5,
    colframe=black!70,
    title=\textbf{SimWorld Image-Only Prompt Skeleton},
    fonttitle=\bfseries
]
\small
\textbf{Role.} Infer a valid node-visiting sequence from the BEV map image.

\textbf{Visual semantics.} START is the green marker. Pickup nodes are red markers labeled P1, P2, \ldots. Delivery nodes are blue markers labeled D1, D2, \ldots. In open scenes, near-straight connections can be reasonable; in indoor, campus, and procedural scenes, route cost should follow visible connectivity rather than Euclidean proximity alone.

\textbf{Task logic.} Pn must appear before Dn; pickup and delivery nodes should not be repeated; the number of active undelivered pickups must respect the load capacity; interleaving is allowed when it reduces route cost.

\textbf{Output schema.} Return only a JSON object with a \texttt{sequence} array of node labels. Do not output coordinates or movement commands.
\begin{tcolorbox}[colback=white, colframe=black!20, sharp corners, boxrule=0.5pt]
\ttfamily\small
\{``sequence'': [``START'', ``P1'', ``P3'', ``D3'', ``D1'']\}
\end{tcolorbox}
\end{tcolorbox}

\begin{tcolorbox}[
    enhanced,
    breakable,
    colback=gray!5,
    colframe=black!70,
    title=\textbf{SimWorld Multimodal Prompt Skeleton},
    fonttitle=\bfseries
]
\small
\textbf{Role.} Generate a pickup--delivery node-visiting sequence using the BEV map and task-to-node mapping.

\textbf{Inputs.}
\begin{itemize}[leftmargin=*, label=-, noitemsep, topsep=2pt]
    \item Time: \{now:.2f\} s; position: \{\{``x'': \{cur\_x:.2f\}, ``y'': \{cur\_y:.2f\}\}\}; speed: \{speed:.2f\} m/s.
    \item Load: \{capacity\}/\{max\_capacity\}.
    \item Carried tasks: \{carried\_tasks\_json\}; available tasks: \{new\_tasks\_json\}.
    \item Task-to-node mapping: \{id\_mapping\_section\}.
\end{itemize}

\textbf{Map grounding.} START is green, pickup nodes are red, and delivery nodes are blue. Task A-0001 maps to P1 and D1, with the same indexing rule for the remaining tasks. Use the mapping table to align tasks with map labels.

\textbf{Hard constraints.}
\begin{enumerate}[leftmargin=*, label=\arabic*., noitemsep, topsep=2pt]
    \item Pn must precede the matching Dn unless the task is already carried.
    \item Each pickup and delivery node may appear at most once; already-carried tasks require only delivery.
    \item Load must never exceed \{max\_capacity\}.
    \item Use only node labels that exist in the task-to-node mapping.
    \item In constrained layouts, use connected free-space topology rather than straight-line distance.
\end{enumerate}

\textbf{Output schema.} Return only a JSON object with a \texttt{sequence} array of node labels. Do not output coordinates or movement commands.
\begin{tcolorbox}[colback=white, colframe=black!20, sharp corners, boxrule=0.5pt]
\ttfamily\small
\{``sequence'': [``START'', ``P1'', ``P3'', ``D3'', ``D1'']\}
\end{tcolorbox}
\end{tcolorbox}

\subsection{Scene-Specific Grounding}
\label{dpp:scene_specific}

The SimWorld templates are instantiated with scene-specific grounding statements while preserving the same output interface. Open layouts emphasize geometric efficiency and smooth sweeps. Campus scenes are treated as semi-open environments where large buildings and blocked regions may invalidate direct shortcuts. Indoor and procedural T1/T2 scenes emphasize connected free space, walls, bottlenecks, and room transitions; prompts explicitly warn that visually close nodes can be topologically far apart. These instructions standardize the physical abstraction presented to the model without changing the task constraints or evaluation schema.

\clearpage
\section*{NeurIPS Paper Checklist}

\begin{enumerate}

\item {\bf Claims}
    \item[] Question: Do the main claims made in the abstract and introduction accurately reflect the paper's contributions and scope?
    \item[] Answer: \answerYes{}
    \item[] Justification: The abstract and introduction outline the key challenges addressed, our contributions, and a summary of the experimental findings.

\item {\bf Limitations}
    \item[] Question: Does the paper discuss the limitations of the work performed by the authors?
    \item[] Answer: \answerYes{}
    \item[] Justification: The paper states the benchmark scope and remaining limitation in the Conclusion, including that temporal and topological complexity are evaluated separately and their integration is left for future work.

\item {\bf Theory assumptions and proofs}
    \item[] Question: For each theoretical result, does the paper provide the full set of assumptions and a complete (and correct) proof?
    \item[] Answer: \answerNA{}
    \item[] Justification: This is an empirical benchmark paper and does not present theorems, lemmas, or formal theoretical proofs.

    \item {\bf Experimental result reproducibility}
    \item[] Question: Does the paper fully disclose all the information needed to reproduce the main experimental results of the paper to the extent that it affects the main claims and/or conclusions of the paper (regardless of whether the code and data are provided or not)?
    \item[] Answer: \answerYes{}
    \item[] Justification: See Appendix \ref{app:enviro}.

\item {\bf Open access to data and code}
    \item[] Question: Does the paper provide open access to the data and code, with sufficient instructions to faithfully reproduce the main experimental results, as described in supplemental material?
    \item[] Answer: \answerYes{}
    \item[] Justification: See our project website \url{https://dualworld.netlify.app/}.

\item {\bf Experimental setting/details}
    \item[] Question: Does the paper specify all the training and test details (e.g., data splits, hyperparameters, how they were chosen, type of optimizer) necessary to understand the results?
    \item[] Answer: \answerYes{}
    \item[] Justification: See Appendix \ref{app:enviro}, \ref{app:solvers}.

\item {\bf Experiment statistical significance}
    \item[] Question: Does the paper report error bars suitably and correctly defined or other appropriate information about the statistical significance of the experiments?
    \item[] Answer: \answerNo{}
    \item[] Justification: The reported tables and figures contain aggregate results across benchmark instances, but do not report error bars, confidence intervals, or statistical significance tests.

\item {\bf Experiments compute resources}
    \item[] Question: For each experiment, does the paper provide sufficient information on the computer resources (type of compute workers, memory, time of execution) needed to reproduce the experiments?
    \item[] Answer: \answerNo{}.
    \item[] Justification: The paper describes the benchmark and evaluation protocol in Section \ref{sec:ex} and Appendix \ref{app:enviro}, but does not provide the hardware, memory, runtime, or total compute required for each experiment.
    
\item {\bf Code of ethics}
    \item[] Question: Does the research conducted in the paper conform, in every respect, with the NeurIPS Code of Ethics \url{https://neurips.cc/public/EthicsGuidelines}?
    \item[] Answer: \answerYes{}.
    \item[] Justification:  We fully adhere to the NeurIPS Code of Ethics.

\item {\bf Broader impacts}
    \item[] Question: Does the paper discuss both potential positive societal impacts and negative societal impacts of the work performed?
    \item[] Answer: \answerNo{}.
    \item[] Justification: The paper does not currently include a dedicated discussion of positive and negative societal impacts; the benchmark is intended for research evaluation rather than direct deployment.
    
\item {\bf Safeguards}
    \item[] Question: Does the paper describe safeguards that have been put in place for responsible release of data or models that have a high risk for misuse (e.g., pre-trained language models, image generators, or scraped datasets)?
    \item[] Answer: \answerNA{}
    \item[] Justification: The paper poses no such risks.

\item {\bf Licenses for existing assets}
    \item[] Question: Are the creators or original owners of assets (e.g., code, data, models), used in the paper, properly credited and are the license and terms of use explicitly mentioned and properly respected?
    \item[] Answer: \answerNo{}
    \item[] Justification: The paper cites the external benchmarks, simulators, models, and related software used, but does not yet enumerate the license and terms of use for every existing asset.

\item {\bf New assets}
    \item[] Question: Are new assets introduced in the paper well documented and is the documentation provided alongside the assets?
    \item[] Answer: \answerYes{}
    \item[] Justification: See our project website \url{https://dualworld.netlify.app/}.

\item {\bf Crowdsourcing and research with human subjects}
    \item[] Question: For crowdsourcing experiments and research with human subjects, does the paper include the full text of instructions given to participants and screenshots, if applicable, as well as details about compensation (if any)? 
    \item[] Answer: \answerNA{}
    \item[] Justification: The paper does not involve crowdsourcing nor research with human subjects.

\item {\bf Institutional review board (IRB) approvals or equivalent for research with human subjects}
    \item[] Question: Does the paper describe potential risks incurred by study participants, whether such risks were disclosed to the subjects, and whether Institutional Review Board (IRB) approvals (or an equivalent approval/review based on the requirements of your country or institution) were obtained?
    \item[] Answer: \answerNA{}
    \item[] Justification: The paper does not involve crowdsourcing nor research with human subjects.

\item {\bf Declaration of LLM usage}
    \item[] Question: Does the paper describe the usage of LLMs if it is an important, original, or non-standard component of the core methods in this research? Note that if the LLM is used only for writing, editing, or formatting purposes and does \emph{not} impact the core methodology, scientific rigor, or originality of the research, declaration is not required.
    \item[] Answer: \answerYes{}
    \item[] Justification: LLMs and VLMs are central evaluation subjects in the benchmark; their model families, task settings, and prompting protocols are described in Section \ref{sec:ex} and Appendix \ref{app:prompt}.

\end{enumerate}

\end{document}